%% file: main.tex
\documentclass[a4paper,11pt]{report}
\usepackage{aliases}

\begin{document}

\title{Deep Learning and Computational
Physics \\ (Lecture Notes)}

\author[1]{Deep Ray, Orazio Pinti and Assad A. Oberai}

\affil[1]{Department of  Aerospace and Mechanical Engineering\\University of Southern California, Los Angeles, California, USA }
\date{}

\maketitle


\tableofcontents

\chapter*{Preface}
\addcontentsline{toc}{chapter}{Preface}
These notes were compiled as lecture notes for a course developed and taught at the University of the Southern California. They should be accessible to a typical engineering  graduate student with a strong background in Applied Mathematics. 

The main objective of these notes is to introduce a student who is familiar with concepts in linear algebra and partial differential equations to select topics in deep learning. These lecture notes exploit the strong connections between deep learning algorithms and the more conventional techniques of computational physics to achieve two goals. First, they use concepts from computational physics to develop an understanding of deep learning algorithms. Not surprisingly, many concepts in deep learning can be connected to similar concepts in computational physics, and one can utilize this connection to better understand these algorithms. Second, several novel deep learning algorithms can be used to solve challenging problems in computational physics. Thus, they offer someone who is interested in modeling a physical phenomena with a complementary set of tools.

\input{Intro}

\input{DLintro}

\input{ResNets}

\input{PDEs}

\input{CNNs}

\input{DeepONets}

\input{probML_GANS}

\bibliographystyle{siam}
\bibliography{ref}

\end{document}

%% file: Intro.tex

\chapter{Introduction}
This course deals with topics that lie at the interface of \textit{computational physics} and \textit{machine learning}. Before we can appreciate the need to combine both these important concepts, we need to understand what each of them mean on their own.

\section{Computational physics}
Computational physics plays a fundamental role in solving many problems in fields of science and engineering. To gain an understanding of this concept, we briefly outline the key steps involved in solving a physical problem:
\begin{enumerate}
    
    \item Consider a physical phenomena and collect measurements of some observable of interest. For example, the measurements of the water height and wave direction obtain from ocean bouys, when studying oceanic waves.   
    \item Based on the observations, postulate a \textit{physical law}. For instance, you observe that the mass of fluid is a closed-system is conserved for all time.
    
    \item Write down a mathematical description of the law. This could make use of ordinary differential equations (ODEs), partial differential equations (PDEs), integral equations, etc.
    
    \item Once the mathematical model is framed, solve for the solution of the system. There are two ways to obatin this:
    
    \begin{enumerate}
    
        \item In certain situations an exact analytical form of the solution can be obtained. For instance one could solve ODEs/PDEs using separation of variables, Laplace transforms, Fourier transforms or integration factors.
        
        \item In most scenarios, exact expressions of the solution cannot be obtained and must be suitable approximated using a numerical algorithm. For instance, one could use forward or backward Euler, mid-point rule, or Runge-Kutta schemes for solving systems of ODEs; or one could use finite difference/volume/element methods methods for solving PDEs.

    \end{enumerate}
    
    \item Once the algorithm to evaluate the solution (exactly or approximately) is designed, use it to validate the mathematical model, i.e., see if the predictions are in tune with the data collected. 
\end{enumerate}

All these steps broadly describe what computational physics entails.

\section{Machine learning}
Unlike computational physics, machine learning (ML) does not require the postulation of a physical law. The general steps involved are:
\begin{enumerate}
    
    \item Collect data by observing the physical phenomena, by real-time measurements of some observable or by using an numerical solver approximating the phenomenon.
    
    \item Train a suitable algorithm using the collected data, with the aim of discovering a pattern or relation between the various samples. See Section \ref{sec:MLeg} for some concrete examples.
    
    \item Once trained, use the ML algorithm to make future predictions, and validate it with additional collected data.
\end{enumerate}

\subsection{Examples of ML}\label{sec:MLeg}
\begin{figure}[htbp]
\begin{center}
\subfigure[Linear regression]{\includegraphics[width=0.3\textwidth]{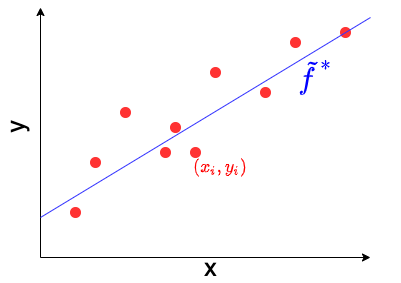}}
\subfigure[Decision tree]{\includegraphics[width=0.3\textwidth]{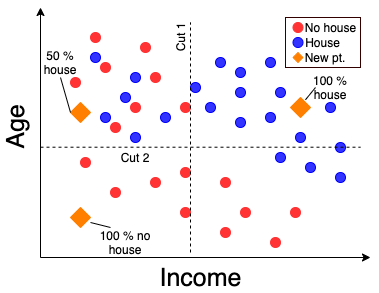}}
\subfigure[Clustering]{\includegraphics[width=0.3\textwidth]{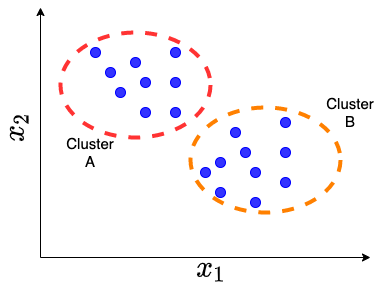}}
\caption{Examples of ML}
\label{fig:MLeg}
\end{center}
\end{figure}

\begin{enumerate}
    
    \item \textbf{Regression algorithms:} Given the set of pairwise data $\{(x_i,y_i): 1 \leq i \leq N\}$ which corresponds to some unknown function $y = f(x)$, fit a polynomial (or any other basis) to this data set in order to approximate $f$. For instance, find the coefficients $a,b$ of the linear fit $\tilde{f}(x; a,b) = ax + b$ to minimize the error 
    \[
    \mathcal{E}(a,b) = \sum_{i=1}^N | y_i - \tilde{f}(x_i) |^2.
    \]
  If $(a^*,b^*) = \arg\min_{a,b} \mathcal{E}(a,b)$, then we can consider $\tilde{f}^*(x) := \tilde{f}(x;a^*,b^*)$ to be the approximation of $f(x)$ (see Figure \ref{fig:MLeg}(a)).  
    
    \item \textbf{Decision trees:} We are given a dataset from a sample population, containing the features: age, income. Furthermore, the data is divided into two groups; an individual in Group A owns a house while an individual in Group B does not. Then, given a features of a new data point, we would like to predict the probability of this new individual owning a house. Decision trees can be used to solve this \text{classification} problem. The way a typical decision tree works is by making cuts the maximize the group-based separation for the samples in the dataset (see Figure \ref{fig:MLeg}(b)). Then, based on these cuts, the algorithm determines the probability of belonging to a particular class/group for a new point. 
    
    \item \textbf{Clustering algorithms:} Given a set of data with a number of features per sample, find cluster/patterns in the data (see Figure \ref{fig:MLeg}(c)).        
\end{enumerate}

\subsection{Types of ML algorithms based leaning task}
Broadly speaking, there are four types of ML algorithms:
\begin{enumerate}
    
    \item \textbf{Supervised learning:} Given the data $\mathcal{S} = \{(\x_i,\y_i): 1 \leq i \leq N\}$, predict $\hat{\y}$ for some new $\hat{\x}$, such that $(\hat{\x}, \hat{\y}) \notin \mathcal{S}$.
    For instance, given a set of images and image labels (e.g. dog, cat, cow, etc), train a classification ML algorithm to learn the relation between images and labels, and use it to predict the label of a new image. 
    
    \item \textbf{Unsupervised learning:} Given the data $\mathcal{S} = \{\x_i \in \Omega_x: 1 \leq i \leq N\}$, find a relation among different regions of $\Omega_x$. For instance, find clusters in the dataset, or find an expression for the probability distribution $p_x(\x)$ governing the spread of this data and generate new samples from it. 
    
    \item \textbf{Semi-supervised learning:} This family of methods falls between the supervised and unsupervised learning families. They typically make use of a combination of labelled and unlabelled data for training. For example let's say we are given 10,000 images that are unlabeled and only 50 images that are labeled. Can we use this dataset to develop an image  classification algorithm? 
    
    \item \textbf{Re-inforcement learning:} The methods belonging to this family learn driven by rewards or penalties for decisions taken. Thus, a suitable path/policy is learned to maximize the reward. These kinds of methods are used to train algorithms to play chess or Go.
    \end{enumerate}
In this course, we will primarily focus on the first two types of ML algorithms.

\section{Artificial Intelligence, Machine Learning and Deep Learning}
\begin{figure}[htbp]
\begin{center}
\includegraphics[width=0.4\textwidth]{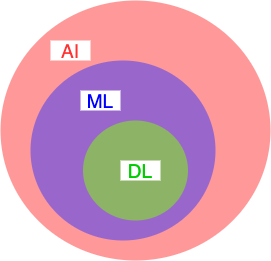}
\caption{The relation between AI, ML and DL}
\label{fig:venn}
\end{center}
\end{figure}

At times, the terms Artificial Intelligence (AI), ML and Deep Learning (DL) are used interchangeably. In reality, these are three related but different concepts. This can be understood by looking at the Venn diagram in Figure \ref{fig:venn}. 

\begin{figure}[htbp]
\begin{center}
\includegraphics[width=0.4\textwidth]{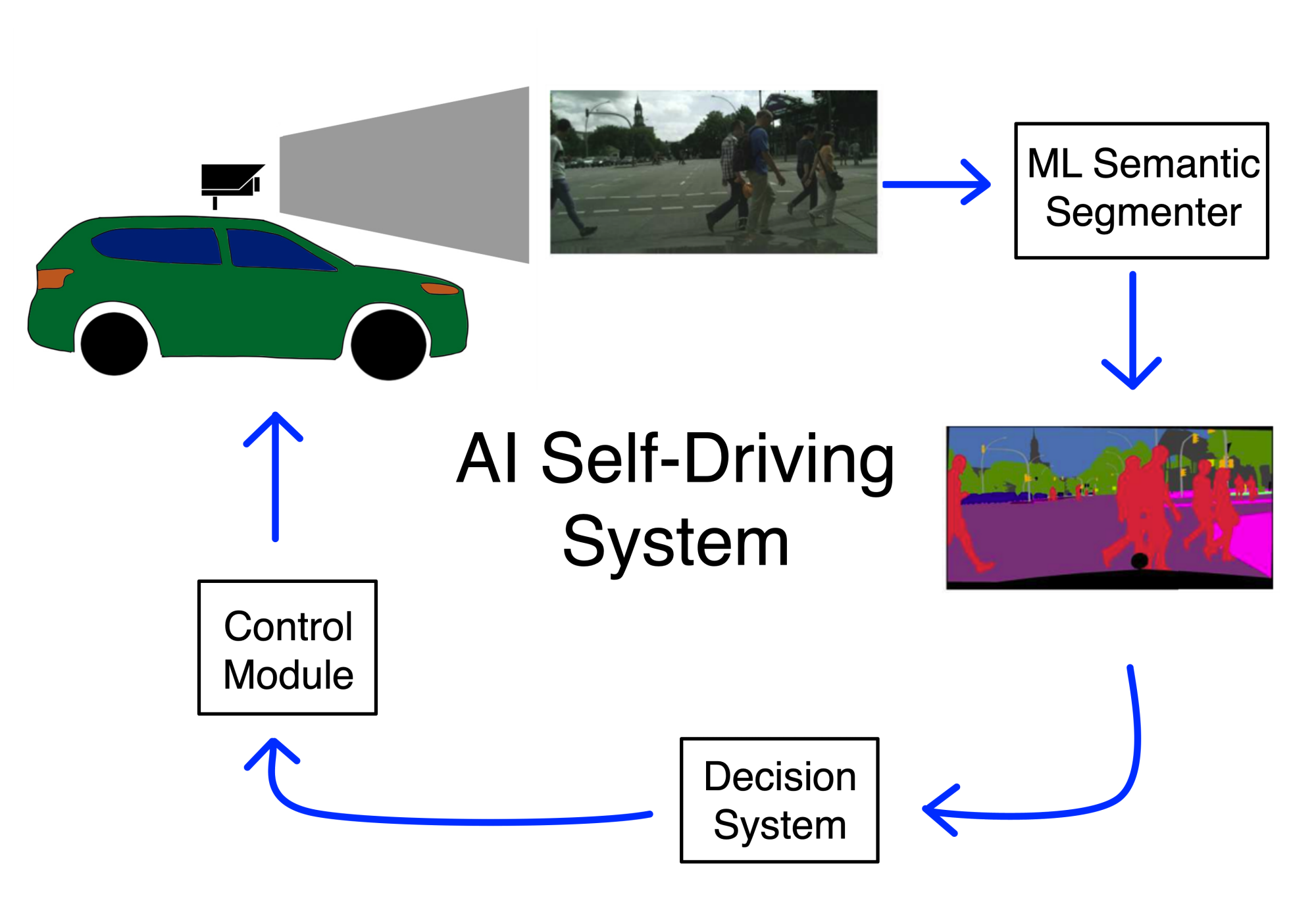}
\caption{Schematic of AI system for a self-driving car. Some illustration taken from \cite{yang2018_sem}.}
\label{fig:AI}
\end{center}
\end{figure}

AI refers to a system with human-like intelligence. While ML is a key component of an AI system, there are other ingredients involved. A self-driving car is a prototypical example of AI. Let's take a closer look at the design of such a system (see Figure \ref{fig:AI}). A car is mounted with a camera which takes lives images/video of the road ahead. These frames are then passed to an ML algorithm which performs a semantic segmentation, i.e., segments out different regions of the frame and classifies the type of object (car, tree, road, sky, etc) in each segment. Once this segmentation is done, it is passed to a \textit{decision system} that decides the next action of the car should be based on this segmented image. This information then passes though a control module that actually controls the mechanical actions of the car. This entire process is mimics what a real driver would do, and is thus artificial intelligence.

On the other hand, machine learning (ML) are the components of this system that are trained using data. That is they learn through data. In the example above, the Semantic Segmenter is one such system. There are many ML algorithms that can perform this task using data, and we will learn some in this course. The Decision System could also be an ML component - where the appropriate decision to be made is learnt from prior data. However, it could be non-ML. Perhaps a rules based expert system.

DL is a subset of ML algorithms. The simplest form of a DL architecture, known as a feed-forward network. It comprises a number of layers of non-linear transformations. The non-linear transformations are applied (component-wise) to an affine transformation of an intermediate output. This architecture is loosely motivated by how signals are transmitted by the central nervous in living organisms. We will study the DL architecture in greater detail in Chapter \ref{chap:dlintro}.


\section{Machine learning and computational physics}
Now that we have a better understanding of computational physics and ML, the next obvious question would be ``why do we need to look at a combination of the two?'' We list down a few motivations below:
\begin{itemize}
    \item For complex patterns of ``physical'' data, ML provides an alternate route to representing mathematical laws. Consider a physical process that contains two important components. Of these, one is well understood and has a trusted mathematical model, and the other is poorly understood and does not have a mathematical description. In this scenario, one may use computational physics for the first component and ML for the second. A concrete example of this would be a system governed by conservation of energy and a complex constitutive model. For the former we may have a well understood mathematical model, while for the latter we may have to rely on ML to develope a model. 
    \item ML in general is very data hungry. But the knowledge of physics can help restrict the manifold on which the input and solution/predictions lie. With such constraints, we can reduce the amount of data required to train the ML algorithm.
    \item Tools for analyzing computational physics (functional analysis, numerical analysis, notions of convergence to exact solutions, probabilistic frameworks) carry over to ML. Applying these tools to ML helps us better understand and design better ML algorithms.
\end{itemize}

We briefly summarize the various topics that will be covered in this course:
\begin{itemize}
    \item Deep Neural Networks (MLPs) and their convergence.
    \item Resnets and their connections with non-linear ODEs (Neural ODEs).
    \item Recurnets and their connections with nonlinear ODEs. 
    \item Convolutional neural networks and their connection to PDEs.
    \item Stochastic gradient descent and how it is related to ODEs.
    \item Deep Learning algorithms for solving PDEs.
    \item Deep Learning algorithms for approximation operators. 
    \item Generative adversarial algorithms and their connection to computational physics.
\end{itemize}

%% file: DLintro.tex

\chapter{Introduction to deep neural networks}
\label{chap:dlintro} 

In this chapter, we will take a closer look at the simplest network architecture that is available known as \textit{multilayer perceptron} (MLP). 

\section{MLP architecture}
\begin{figure}[htbp]
\begin{center}
\includegraphics[width=0.8\textwidth]{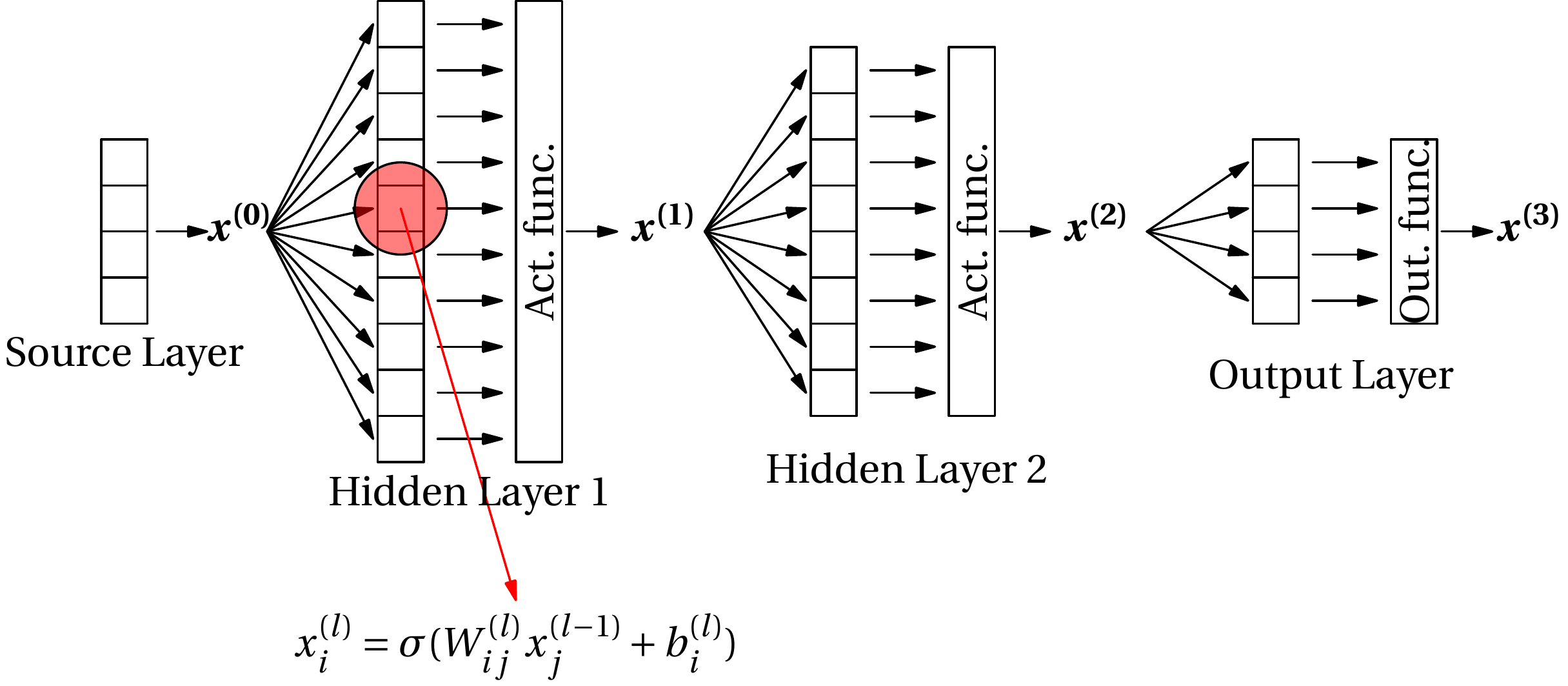}
\caption{MLP with 2 hidden layers}
\label{fig:MLP}
\end{center}
\end{figure}

Let us define our objective as the approximation of a function $\f : \x \in \Ro^d \mapsto \y \in \Ro^D$ using an MLP, which we denote as $\bm{\mathcal{F}}$. Computing units of an MLP, called \textit{artificial neurons}, are stacked in a number of consecutive layers. The zeroth layer of $\bm{\mathcal{F}}$ is called the \textit{source layer}, which is not a computing layer but is only responsible for providing an input (of dimension $d$) to the network. The last layer of $\bm{\mathcal{F}}$ is known as the \textit{output layer}, which outputs the network's prediction (of dimension $D$). Every other layer in between is known as a \textit{hidden layer}. The number of neurons in a layer defines the width of that layer. A schematic of an MLP with 2 hidden layers is shown in Figure \ref{fig:MLP}. 

To understand the operations occurring inside an MLP, let us define some notations. We consider a network with $L$ hidden layers, with the width of layer $(l)$ denoted as $H_l$ for $l=0,1,...,L+1$. Note that for consistency with the function $f$ that we are trying to approximate, we must have $H_0 = d$ and $H_{L+1} = D$. Let us denote the output vector for $l$-th layer by $\x^{(l)} \in \Ro^{H_{l}}$, which will serve as the input to the next layer. We set $\x^{(0)} = \x \in \Ro^d$ which will be the input signal provided by the input layer. In each layer $l$, $1 \leq l \leq L+1$, the $i$-th neuron performs an affine transformation on that layers input $\x^{(l-1)}$ followed by a non-linear transformation
\begin{equation}
x_i^{(l)} = \sigma \Big( \underbrace{W^{(l)}_{ij} x_j^{(l-1)} }_{\text{Einstein sum}}+ b^{(l)}_i\Big), \quad 1 \leq i \leq H_l, \quad 1 \leq j \leq H_{l-1}
\end{equation}
where $W^{(l)}_{ij}$ and $b^{(l)}_i$ are respectively known as the weights and bias associated with $i$-th neuron of layer $l$, while the function $\sigma(.)$ is known as the \textit{activation function}, and plays a pivotal role in helping the network to represent non-linear complex functions. If we set $\W^{(l)} \in \Ro^{H_{l-1} \times H_{l}}$ to be the weight matrix for layer $l$ and $\bb^{(l)} \in \Ro^{H_l}$ to be the bias vector for layer $l$, then we can re-write the action of the whole layer as
\begin{equation}
\x^{(l)} = \sigma\left(\mathcal{A}^{(l)}(\x^{(l-1)})\right), \quad \mathcal{A}^{(l)}(\x^{(l-1)}) = \W^{(l)} \x^{(l-1)}+ \bb^{(l)}
\end{equation}
where the activation function is applied component-wise. Thus, the action of the whole network $\bm{\mathcal{F}}: \Ro^d \mapsto \Ro^D$ can be mathematically seen as a composition of alternating affine transformations and component-wise activations
\begin{equation}
\bm{\mathcal{F}}(\x) = \mathcal{A}^{(L+1)} \circ \sigma \circ \mathcal{A}^{(L)} \circ  \sigma \circ \mathcal{A}^{(L-1)} \circ \cdots  \circ \sigma \circ \mathcal{A}^{(1)} (\x).
\end{equation}
We make a few remarks here:
\begin{enumerate}
\item For simplicity of the representation, we assume that the same activation function is used across all layers of the network. However, this is not a strict rule. In fact, there is recent evidence that suggests that alternating activation function from layer to layer leads to better neural networks \cite{Yarotsky2021}.
\item At times, there might be an output function $O$ instead of an activation function at the end of the output layer, which is typically used to reformulate the output into a suitable form. We will see examples of such functions later in the course.
\item We will use the term \textit{depth} of the network to denote the number of computing layers in the MLP, i.e. the number of hidden layers and the output layer, which would be $L+1$ as per the notations used above.
\end{enumerate}

The parameters of the network is all the weights and biases, which we will represent as 
\[
\btheta = \{\W^{(l)}, \bb^{(l)} \}_{l=1}^{L+1} \in \Ro^{N_\theta}
\]
where $N_\theta$ denotes the total number of parameters of the network. The network $\bm{\mathcal{F}}(\x ;\btheta)$ represents a family of parameterized functions, where $\btheta$ needs to suitably chosen such that the network approximates the target function $f(\x)$ at the input $\x$.

\begin{question}	
Prove that $N_\theta = \sum_{l=1}^{L+1} (H_{l-1} + 1) H_l$.
\end{question}

\section{Activation functions}
\begin{figure}[htbp!]
\begin{center}
\subfigure[Linear]{\includegraphics[width=0.3\textwidth]{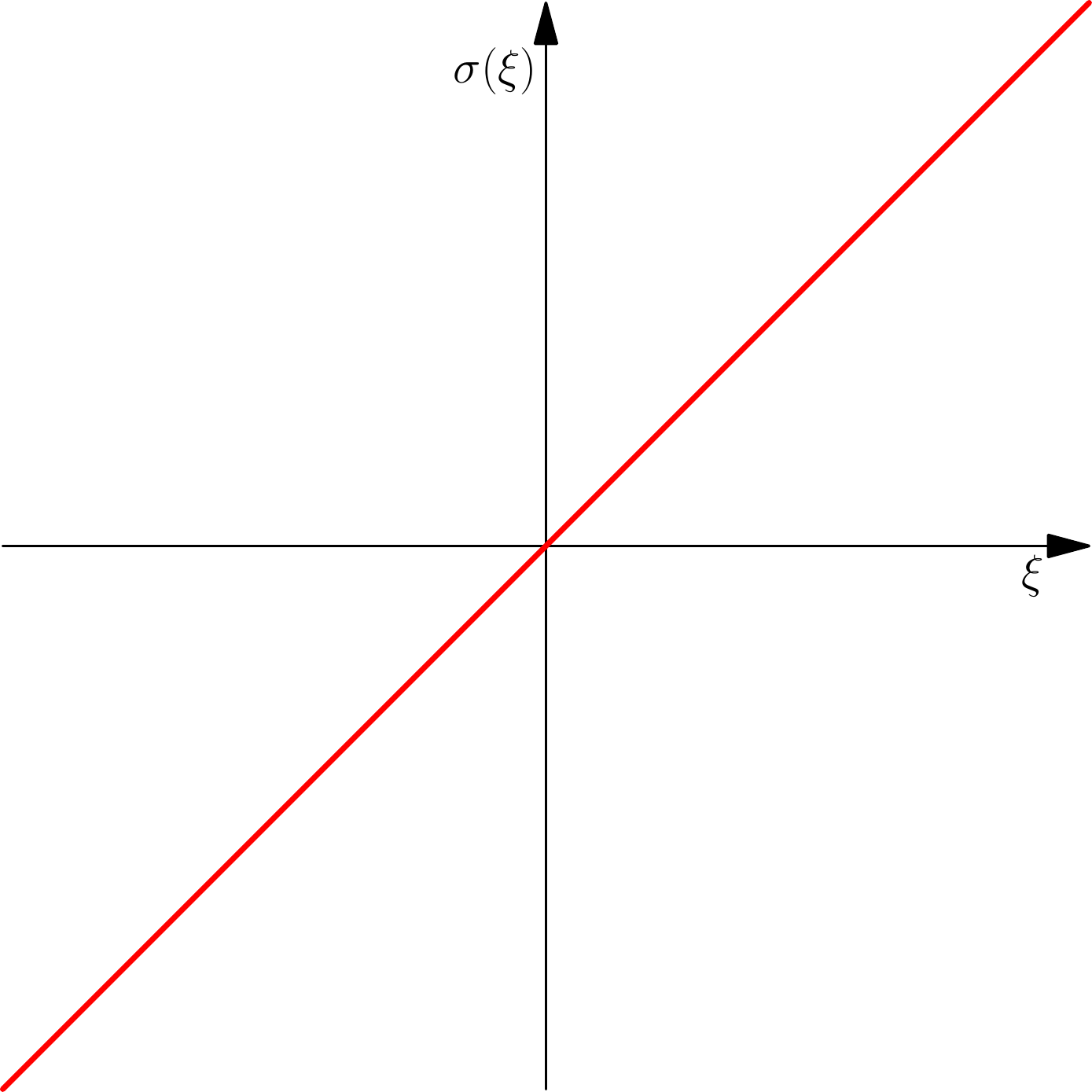}}
\subfigure[ReLU]{\includegraphics[width=0.3\textwidth]{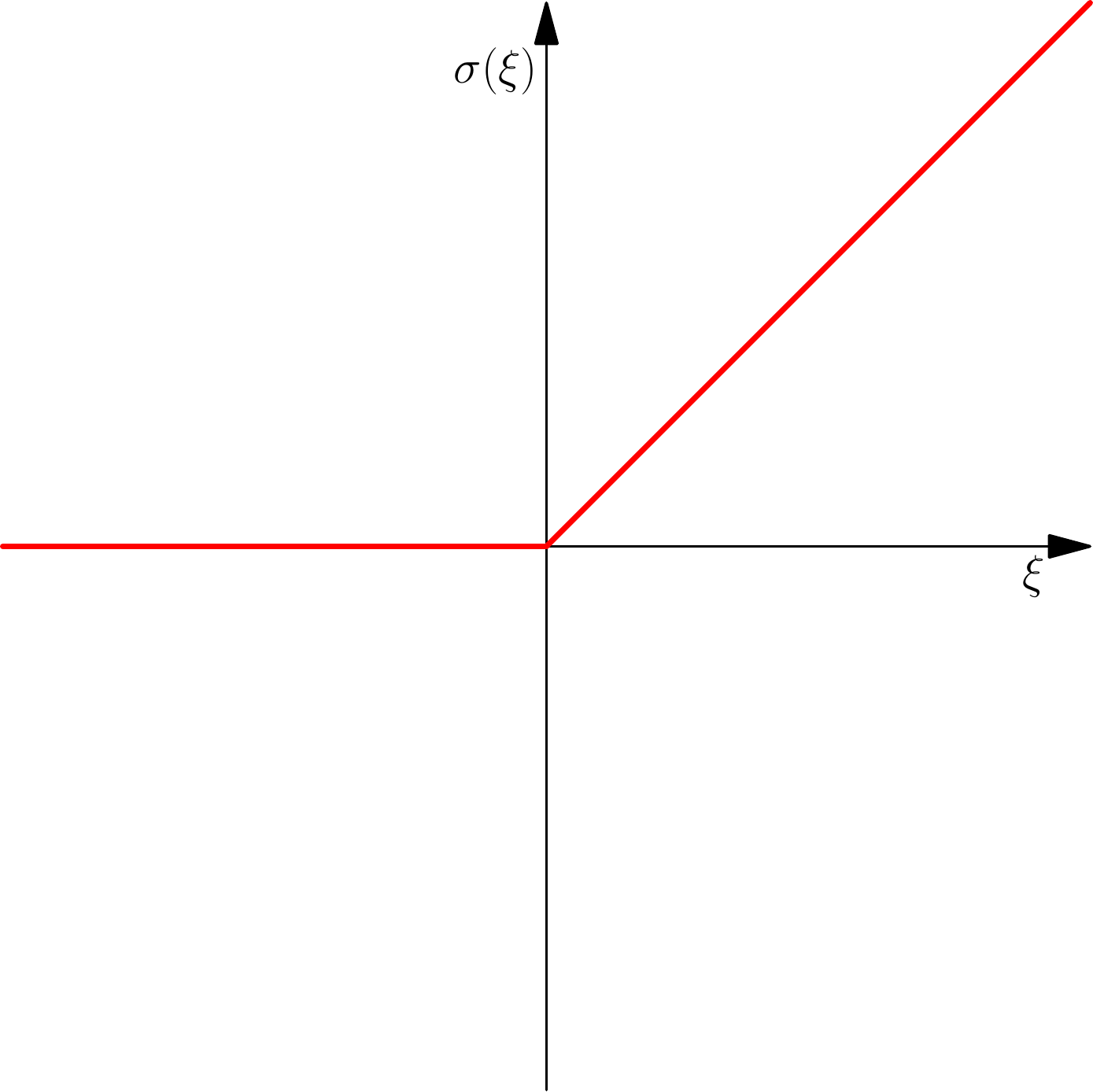}}
\subfigure[Leaky ReLU]{\includegraphics[width=0.3\textwidth]{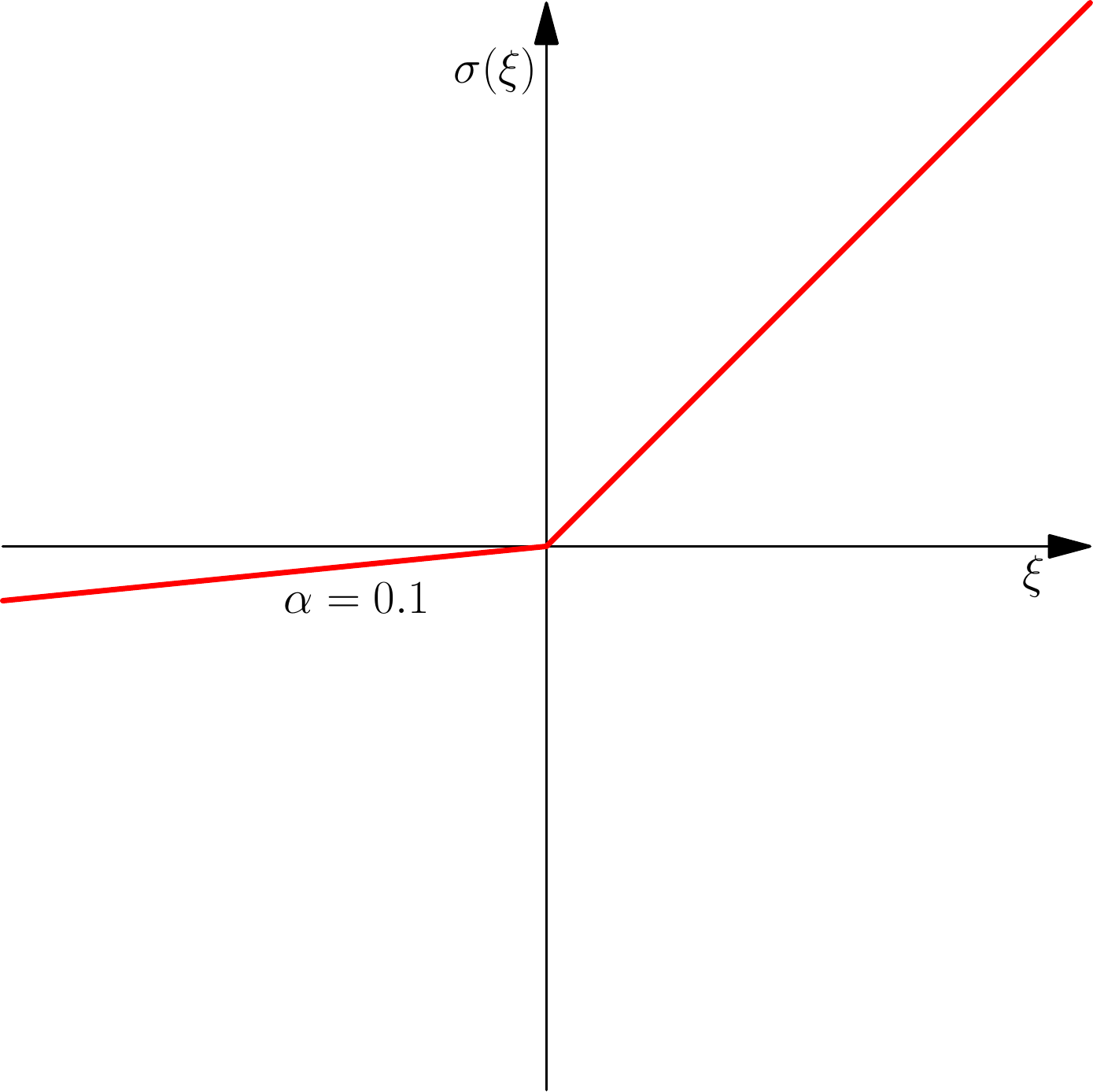}}\\
\subfigure[Logistic]{\includegraphics[width=0.3\textwidth]{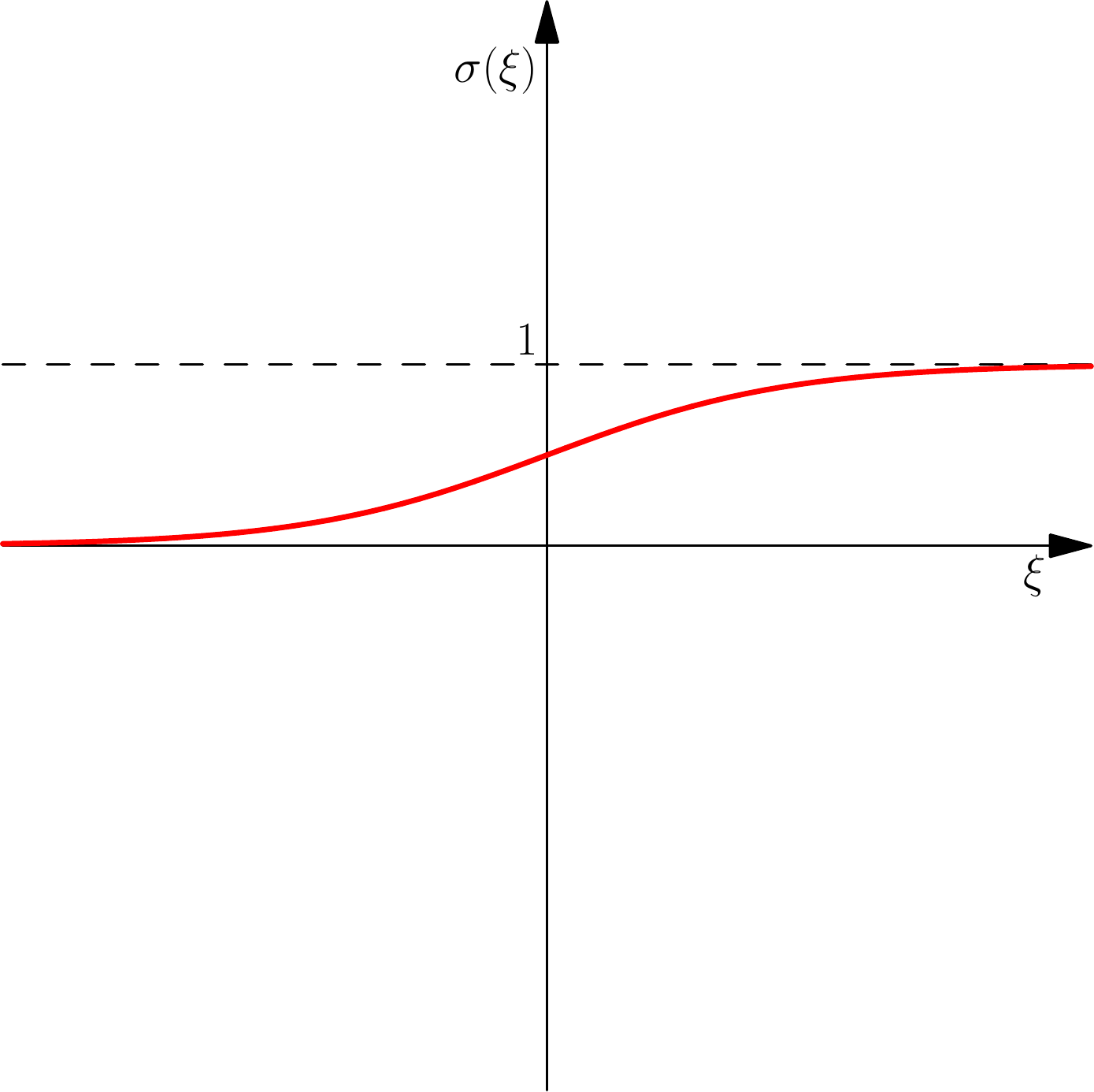}}
\subfigure[Tanh]{\includegraphics[width=0.3\textwidth]{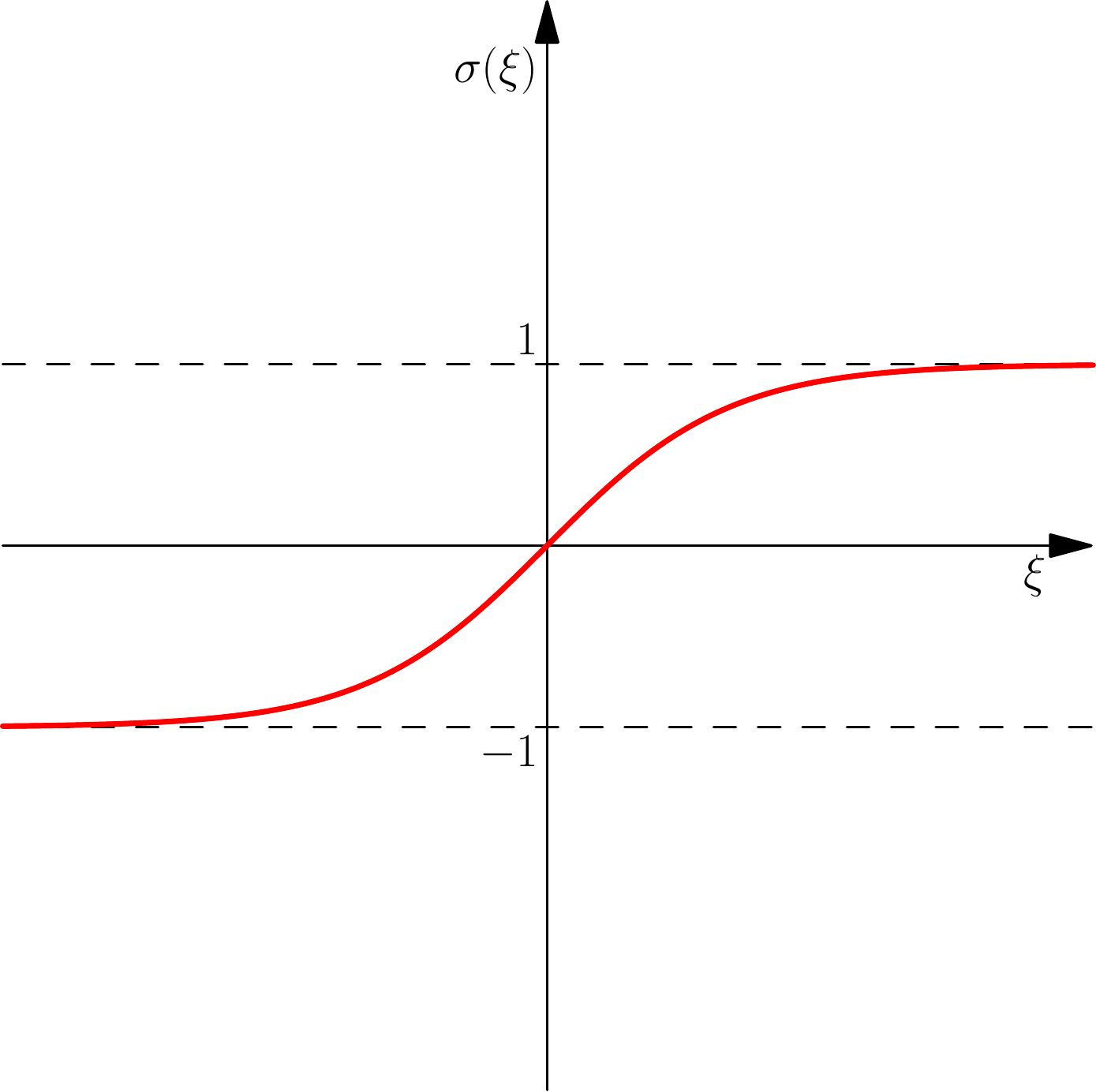}}
\subfigure[Sine]{\includegraphics[width=0.3\textwidth]{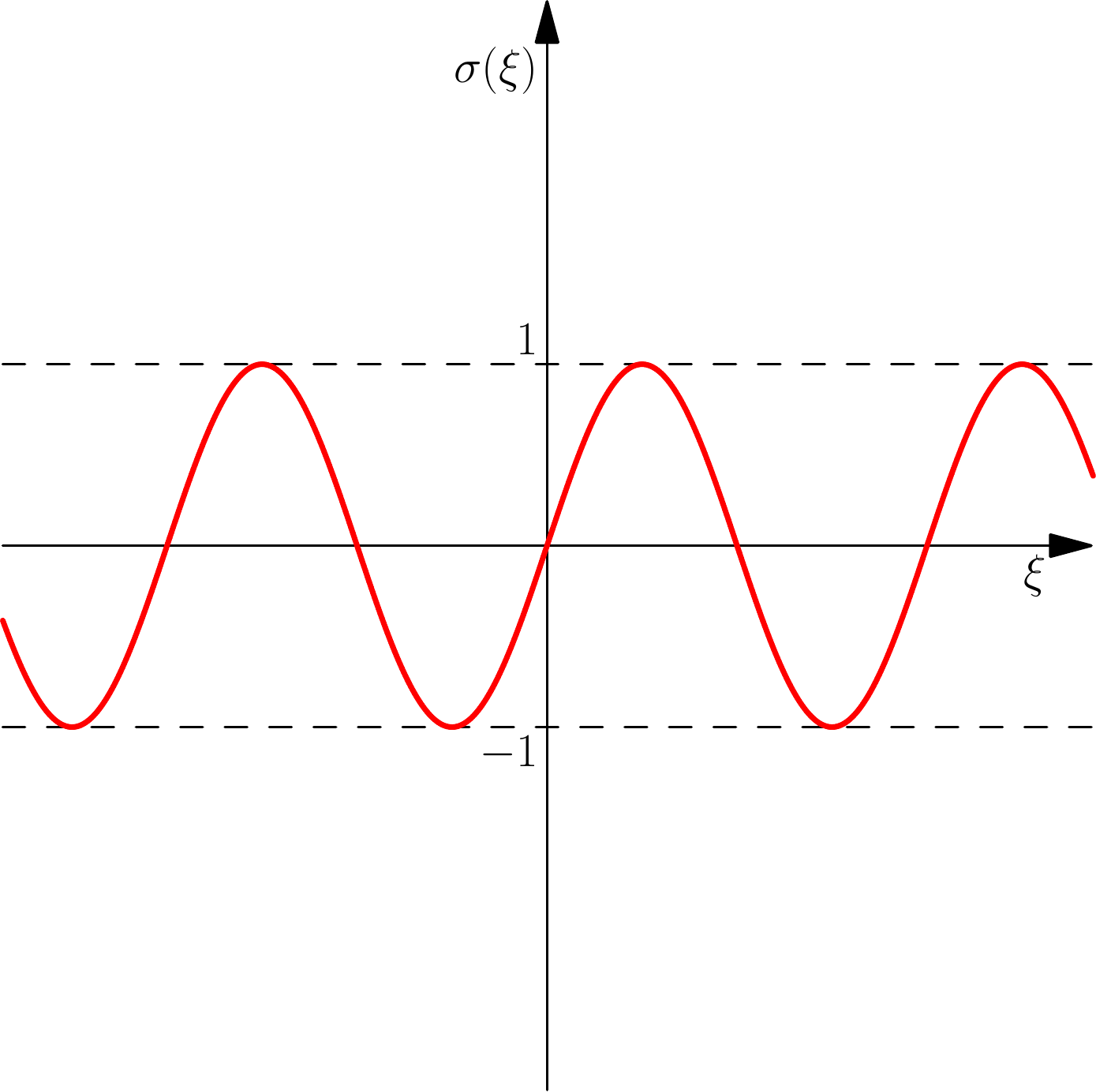}}
\caption{Examples of activation functions}
\label{fig:actfunc}
\end{center}
\end{figure}

The activation function is perhaps the most important component of an MLP. A large number of activations are available in literature, each with its own advantages and disadvantages. Let us take a look at a few of these options (also see Figure \ref{fig:actfunc}).

\subsection{Linear activation}
The simplest activation corresponds to $\sigma(\xi) = \xi$. Some features of this function are
\begin{itemize}
\item The function is infinitely smooth, but all derivatives beyond the second derivative are zero.
\item The range of the function is $(-\infty,\infty)$.
\item 
Using the linear activation function (in all layers) will reduce the entire network to a single affine transformation of the input $\x$. In other words, the network will be nothing more that a linear approximation of the target function $\f$, which is not useful if $\f$ is highly non-linear. 
\end{itemize}

\subsection{Rectified linear unit (ReLU)}
This function is piecewise linear and defined as
\begin{equation}
\sigma(\xi) = \max\{0,\xi\} = \begin{cases} \xi, & \quad \text{if } \xi \geq 0 \\ 0, & \quad \text{if } \xi < 0  \end{cases}
\end{equation}
This is one of the most popular activation functions used in practice. Some features of this function are:
\begin{itemize}
\item The function is continuous, while its derivative will be piecewise constant with a jump  $\xi = 0$. The second derivative will be a dirac function concentrated at $\xi=0$. In other words, the higher-order derivates (greater than 1) are not well-defined.
\item The range of the function is $[0,\infty)$.
\end{itemize}

\subsection{Leaky ReLU}
The ReLU activation leads to a null output from a neuron if the affine transformation of the neuron is negative. This can lead to the phenomena of \textit{dying neurons} \cite{Mass2013} while training a neural network, where neurons drops out completely from the network and no longer contribute to the final prediction. To overcome this challenge, a leaky version ReLU was designed
\begin{equation}
\sigma(\xi; \alpha) = \begin{cases} \xi, & \quad \text{if } \xi \geq 0 \\ \alpha \xi, & \quad \text{if } \xi < 0 \end{cases}
\end{equation}
where $\alpha$ becomes a network \textit{hyper-parameter}. Some features of this function are:
\begin{itemize}
\item The derivates of Leaky ReLU behave in the same way as those for ReLU.
\item The range of the function is $(-\infty,\infty)$.
\end{itemize}

\subsection{Logistic function}
The Logistic or Sigmoid activation function is given by
\begin{equation}
\sigma(\xi)  = \frac{1}{1 + e^{-\xi}}
\end{equation}
and has the following properties
\begin{itemize}
\item The function is infinitely smooth and monotonic.
\item The range of the function is $(0,1)$, i.e., the function is bounded. Such a function is useful in representing probabilities. 
\item Since the derivative quickly decays to zero away from $\xi=0$, this activation function can lead to slow convergence of the network while training.
\end{itemize}

\subsection{Tanh}
The tanh function is can be seen as a symmetric extension of the logistic function
\begin{equation}
\sigma(\xi)  = \frac{e^{\xi} - e^{-\xi}}{e^{\xi} + e^{-\xi}}
\end{equation}
and has the following properties
\begin{itemize}
\item The function is infinitely smooth and monotonic.
\item The range of the function is $(-1,1)$, i.e., the function is bounded. Note that it maps zeros input to zero, while pushing positive (negative) inputs to +1 (-1).
\item Similar to the logistic function, the derivative of tanh quickly decays to zero away from $\xi=0$ and can thus lead to slow convergence while training networks.
\end{itemize}

\subsection{Sine}
Recently, the sine function, i.e., $\sigma(\xi) = \sin(\xi)$ has been proposed as an efficient activation function \cite{Siren2020}. It has the best features of all the activation function discussed above:
\begin{itemize}
\item The function is infinitely smooth.
\item The range of the function is $(-1,1)$, i.e., the function is bounded. 
\item None of the derivatives of this function decay to zero.
\end{itemize}

\begin{question}	
Can you think of an MLP architecture with the sine activation function, which leads to an approximation very similar to a Fourier series expansion?
\end{question}

\section{Expressivity of a network}
\begin{figure}[htbp!]
\begin{center}
\subfigure[MLP with $L=1$,$W=2$]{\includegraphics[width=0.45\textwidth]{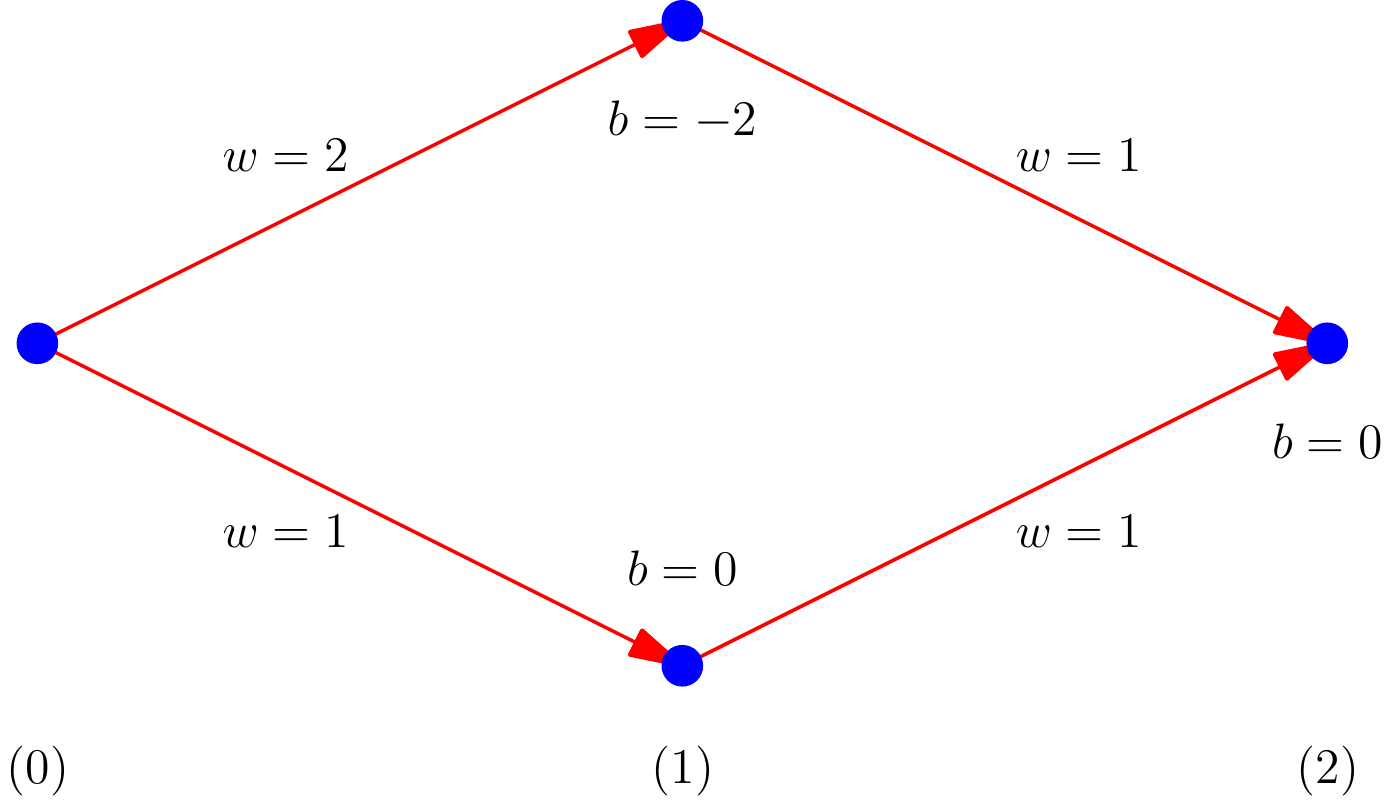}}
\subfigure[$x^{(1)}_1$ vs $x^{(0)}_1$]{\includegraphics[width=0.45\textwidth]{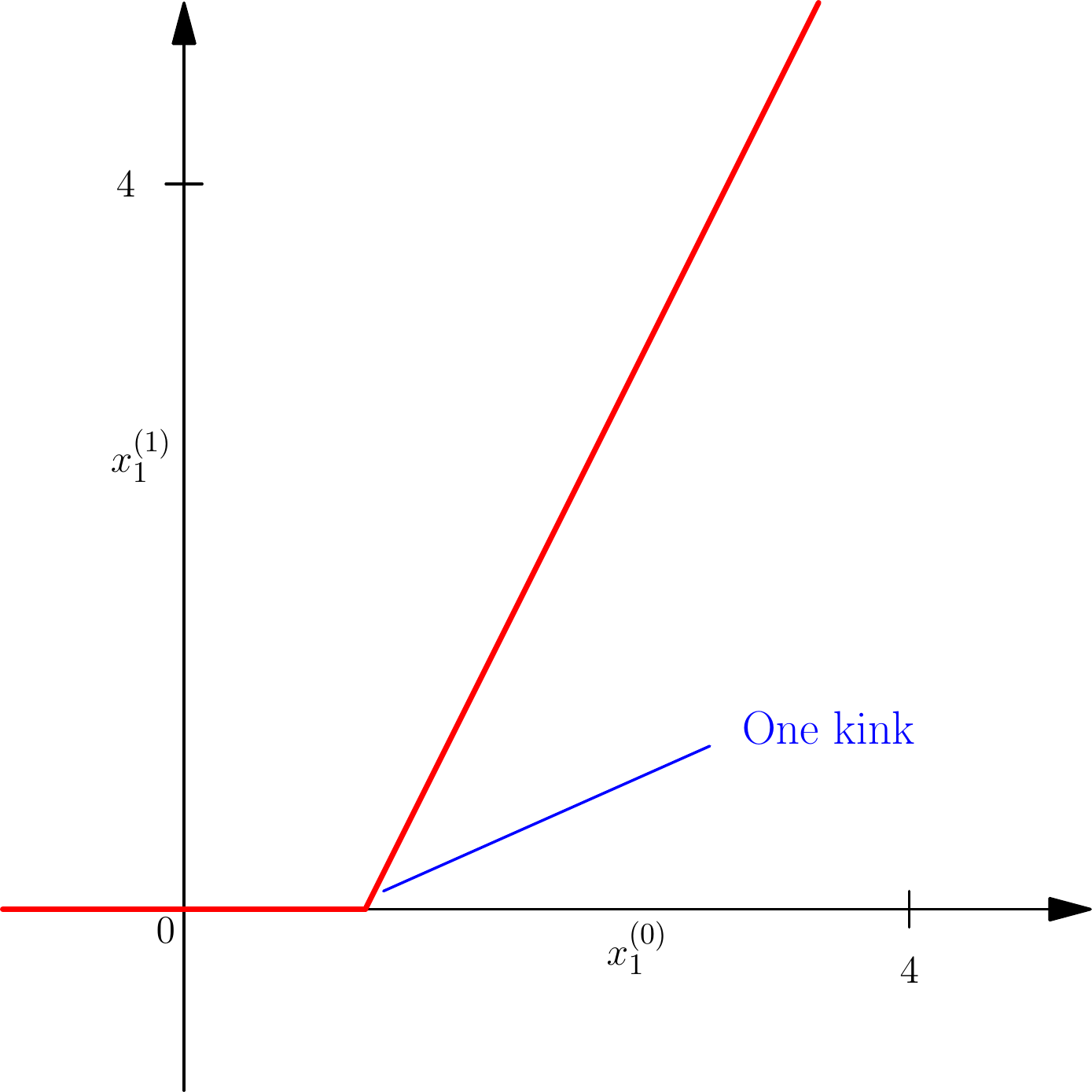}}
\subfigure[$x^{(1)}_2$ vs $x^{(0)}_1$]{\includegraphics[width=0.45\textwidth]{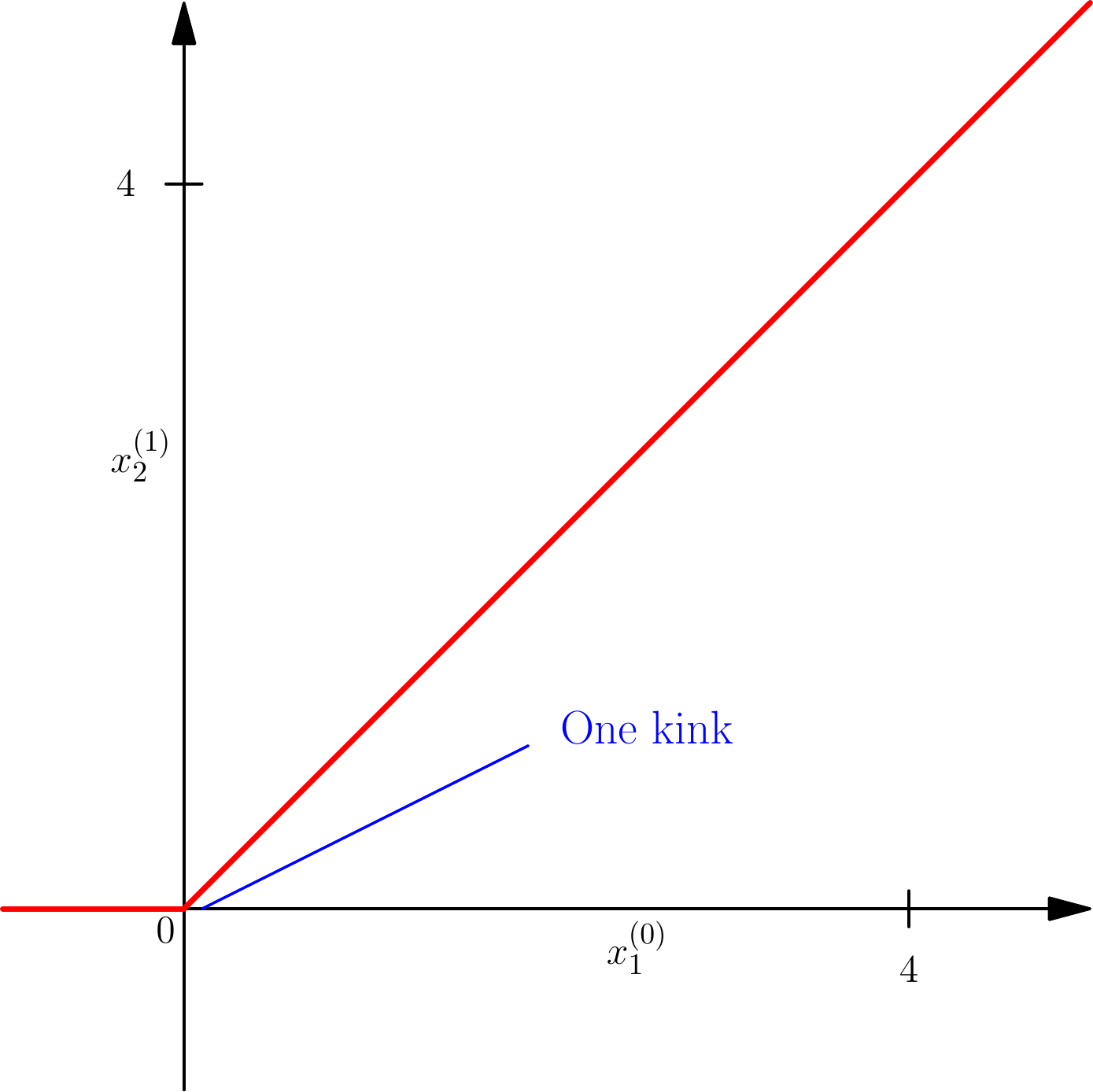}}
\subfigure[$x^{(2)}_1$ vs $x^{(0)}_1$]{\includegraphics[width=0.45\textwidth]{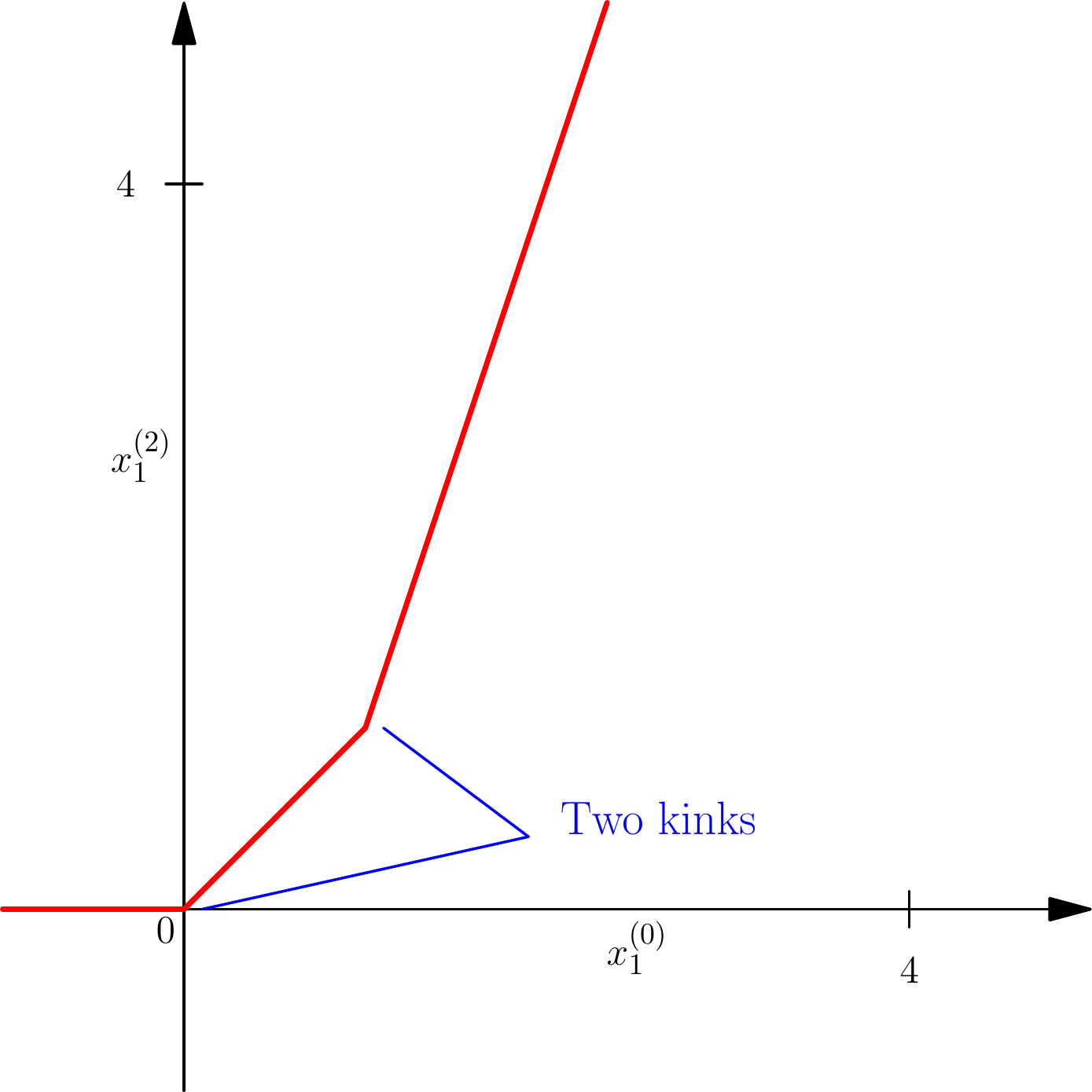}}
\caption{Examples to understand the expressivity of neural networks}
\label{fig:exp_eg}
\end{center}
\end{figure}

Let us try to understand the effects of $N_\theta$ increases. To see this, let us consider a simple example using the ReLU activation function, i.e., $\sigma(\xi) = \max\{\xi,0\}$. We set $d=D=1$, $L=1$ and the parameters
\[
\W^{(1)} = \begin{bmatrix} 2 \\ 1\end{bmatrix}, \quad \bb^{(1)} = \begin{bmatrix} -2 \\ 0 \end{bmatrix}, \quad \W^{(2)} = \begin{bmatrix} 1 &  1 \end{bmatrix}, \quad b^{(2)} =  0.
\]
as shown in Figure \ref{fig:exp_eg}(a). Then the various layer outputs are 
\[
x_1^{(1)} = \max\{2 x^{(0)}_1 - 2,0\},  \quad x_2^{(1)} = \max\{x^{(0)}_1,0\},  \quad x_1^{(2)} = \max\{2 x^{(0)}_1 - 2,0\} + \max\{x^{(0)}_1,0\}.
\]
Notice that while the the output $\x^{(1)}$ of the hidden layer (see Figures \ref{fig:exp_eg}(b) and (c)) have only one corner/kink, the final output ends up having two kinks (see Figures \ref{fig:exp_eg}(d)).

We generalize this formulation to a bigger network with $L$ hidden layers each of width $H$. Then one can expect that $x^{(1)}_i$, $1\leq i \leq H$ will have a single kink, with the location and angle of the kink depending on the weights and bias associated with each neuron of the hidden layer. The vector $\x^{(1)}$ is passed to the next hidden layer, where each neuron will combine the single kinks and give an output with possibly $H$ kinks. Once again, the location and angles of the $H$ kinks in the output from each neuron of the second hidden layer will be different. The location of the kinks will be different because each neuron is allowed a different bias, and therefore can induce a different shift. Continuing this argument, one can expect the number of kinks to increase as $H$, $H^2$, $H^3$ as it passes through the various hidden layers with width $H$. In general the total number of kinks can grow as $H^L$. In other words, the networks have the ability to become more expressive as the depth (and width) of the network is increased.

\subsection{Universal approximation results}\label{sec:uni_app_thms}
To quantify the expressivity of networks in a mathematically rigorous manner, we look at some results about the approximation properties of MLPs. For these results, we assume $K \subset \Ro^d$ is a closed and bounded set.

\begin{theorem}[Pinkus, 1999 \cite{Pinkus1999}]
Let $f:K \rightarrow \Ro$, i.e.,  $D=1$, be a continuous function. Then given an $\epsilon >0$, there exists an MLP with a single hidden layer ($L=1$), arbitrary width $H$ and a non-polynomial continuous activation $\sigma$ such that
\[
\max_{\x \in K} |\mathcal{F}(\x;\btheta) - f(\x)| \leq \epsilon.
\] 
\end{theorem}

\begin{theorem}[Kidger, 2020 \cite{Kidger2020}]
Let $\f:K \rightarrow \Ro^D$ be a continuous vector-valued function. Then given an $\epsilon >0$, there exists an MLP with arbitrary number of hidden layers $L$, each having width $H \geq d + D + 2$, a continuous activation $\sigma$ (with some additional mild conditions), such that
\[
\max_{\x \in K}  \|\bm{\mathcal{F}}(\x;\btheta) - \f(\x)\| \leq \epsilon.
\] 
\end{theorem}

\begin{theorem}[Yarotsky, 2021 \cite{Yarotsky2021}]
Let $f:K \rightarrow \Ro$ be a function with two continuous derivates, i.e., $f \in C^2(K)$. Consider an MLP with ReLU activations and $H \geq 2d + 10$. Then there exists a network with this configuration such that the error converges as
\[
\max_{\x \in K}  |\mathcal{F}(\x;\btheta) - \f(\x)| \leq C(N_\theta)^{-4}
\] 
where $C$ is a constant depending on the number of network parameters.
\end{theorem}

Numerical results like those mentioned above help demystify the ``black-box'' nature of neural network, and serve as useful practical guidelines when designing network architectures.

\section{Training, validation and testing of neural networks}
Now that we have a better understanding of the architecture of MLPs, we would now like to discuss how the parameters of these networks are set to approximate some target function. We restrict our discussions to the framework of supervised learning.

Let us assume that we are given a dataset of pairwise samples $\mathcal{S} = \{(\x_i,\y_i): 1 \leq i \leq N\}$ corresponding to a target function $\f: \x \mapsto \y$. We wish to approximate this function using the neural network
\[
\mathcal{F}(\x; \btheta, \Hp)
\]
where $\btheta$ are the network parameters defined before, while $\Hp$ corresponds to the \textit{hyper-parameters} of the network such as the depth $L+1$, width $H$, type of activation function $\sigma$, etc. The strategy to design a robust network involves three steps:
\begin{enumerate}
\item Find the optimal values of $\btheta$ (for a fixed $\Hp$)  in the \textit{training} phase.
\item Find the optimal values of $\Hp$ in the \textit{validation} phase.
\item Test the performance of the network on unseen data on the \textit{testing} phase.
\end{enumerate}

To accomplish these three tasks, it is first customary to split the dataset $\mathcal{S}$ into three distinct parts: a \textit{training set} with $N_{\text{train}}$ samples, a \textit{validation set} with $N_{\text{val}}$ samples and \textit{test set} with $N_{\text{test}}$ samples, with $N = N_{\text{train}} + N_{\text{val}} + N_{\text{test}}$. Typically, one uses around 60\% of the samples as training samples, 20\% as validation samples and the remaining 20\% for testing. 

Splitting the dataset is necessary as neural networks are heavily over-parameterized functions. The large number of degrees of freedom available to model the data can lead to over-fitting the data. This happens when the error or noise present in the data drives the behavior of the network more than the underlying input-output relation itself. Thus, a part of the data is used to determine $\btheta$, and another part to determine the hyper-parameters $\Hp$. The remainder of the data is kept aside for testing the performance 
 of the trained network on unseen data, i.e., the network's ability to \textit{generalize} well.

Now let us discuss how this split is used during the three phases in further details:

\paragraph{Training:}
Training the network makes use of the training set $\mathcal{S}_{\text{train}}$ to solve the following optimization problem:
Find
\[
\btheta^* = \amin_{\btheta} \Pi_\text{train}(\btheta), \quad \text{where} \quad 
\Pi_\text{train} (\btheta)= \frac{1}{N_\text{train}} \sum_{\substack{i=1\\(\x_i,\y_i) \in \mathcal{S}_{\text{train}}}}^{N_\text{train}} \|\y_i - \bm{\mathcal{F}}(\x_i; \btheta, \Hp)\|^2
\]
for some fixed $\Hp$. The optimal $\btheta^*$ is obtained using a suitable gradient based algorithm (will be discussed later). The function $\Pi_\text{train}$ is referred to as the loss function. In the example above we have used the mean-squared loss function. Later we will consider other types of loss functions. 

\paragraph{Validation:}
Validation of the network involves using the validation set $\mathcal{S}_{\text{val}}$ to solve the following optimization problem:
Find
\[
\Hp^* = \amin_{\Hp} \Pi_\text{val}(\Hp), \quad \text{where} \quad 
\Pi_\text{val} (\Hp)= \frac{1}{N_\text{val}} \sum_{\substack{i=1\\(\x_i,\y_i) \in \mathcal{S}_{\text{val}}}}^{N_\text{val}} \|\y_i - \bm{\mathcal{F}}(\x_i; \btheta^*, \Hp)\|^2.
\]
The optimal $\Hp^*$ is obtained using a techniques such as (random or tensor) grid search.

\paragraph{Testing:} Once the "best" network is obtained, characterized by $\btheta^*$ and $\Hp^*$, it is evaluated on the test set $\mathcal{S}_\text{test}$ to estimate the networks performance on data not used during the first two phases. 
\[
\Pi_\text{test} = \frac{1}{N_\text{test}} \sum_{\substack{i=1\\(\x_i,\y_i) \in \mathcal{S}_{\text{test}}}}^{N_\text{test}} \|\y_i - \bm{\mathcal{F}}(\x_i; \btheta^*, \Hp^*)\|^2.
\]
This testing error is also known as the (approximate) \textit{generalizing error} of the network.

Let's see an example to better understand how such a network is obtained
\begin{example}
Let us consider an MLP where all hyper-parameters are fixed except for the following flexible choices
\[
\sigma \in \{\text{ReLU, tanh}\}, \quad L \in \{10,20\}.
\]
We use the following algorithm
\begin{enumerate}
\item For each possible $\sigma, L$ pair:
\begin{enumerate}
\item Find $\btheta^* = \amin_{\btheta} \Pi_\text{train}(\btheta)$
\item With this $\btheta^*$, evaluate $\Pi_\text{val}(\Hp)$
\end{enumerate}
\item Select $\Hp^*$ to be the one that gave the smallest value of $\Pi_\text{val}(\Hp)$.	
\item Finally, report $\Pi_\text{test}$ for this $\Hp^*$ and the corresponding $\btheta^*$.
\end{enumerate}

\end{example}

\section{Generalizability} 
If we train a network that has a small value of $\Pi_\text{train}$ and $\Pi_\text{val}$, does it ensure that $\Pi_\text{test}$ will be small? This question is addressed by studying the \textit{generalizability} of the trained network, i.e., it capability to perform well on data not seen while training/validating the network. If the network is trained to overfit the training data, the network will typically lead to poor predictions on test data. Typically, if $\mathcal{S}_\text{train}$, $\mathcal{S}_\text{val}$ and $\mathcal{S}_\text{test}$ are chosen from the same distribution of data, then a small value of $\Pi_\text{train},\Pi_\text{val}$ can lead to small values of $\Pi_\text{test}$. Let us look at the commonly used technique to avoid data overfitting, called \textit{regularization}.

\subsection{Regularization}\label{sec:reg}
\begin{figure}[htbp!]
\begin{center}
\subfigure[Loss function landscape]{\includegraphics[width=0.4\textwidth]{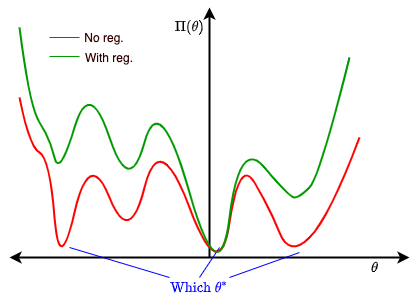}}
\subfigure[Network sensitivity]{\includegraphics[width=0.55\textwidth]{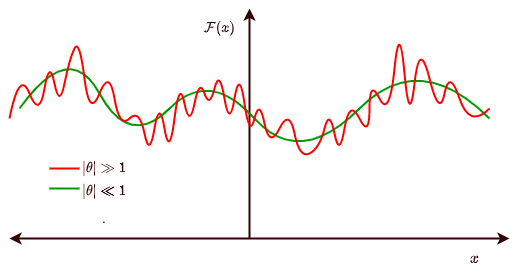}}
\caption{The effect of regularization on the loss function. We have assumed a scalar $\theta$ for easier illustration.}
\label{fig:reg}
\end{center}
\end{figure}

Neural networks, especially MLPs, are almost always \textit{over-parametrized}, i.e., $N_\theta \gg N$ where $N$ is the number of training samples. This would lead to a highly non-linear network model, for which the loss function $\Pi(\btheta)$ (where we omit the subscript "train" for brevity) can have a landscape with many local minimas (see Figure \ref{fig:reg}(a)). Then how do we determine which minima leads to a better generalization? To nudge the choice of $\btheta^*$ in a more favorable direction, a regularization technique can be employed.

The simplest method of regularization involves augmenting a penalty term to the loss function:
\[
\Pi(\btheta) \longrightarrow \Pi(\btheta) + \alpha \|\btheta\|, \quad \alpha \geq 0
\]
where $\alpha$ is a regularization hyper-parameter, and $\|\btheta\|$ is a suitable norm of the network parameters $\btheta$. This augmentation can change the landscape of $\Pi(\btheta)$ as illustrated in Figure \ref{fig:reg}(a). In other words, such a regularization encourages the selection of a minima corresponding to smaller values of the parameters $\btheta$. 

It is not obvious why a smaller value of $\btheta$ would be a better choice. To see why this is better, consider the intermediate network output
\[
x_1^{(1)} = \sigma( W_{1j}^{(1)} x_j^{(0)} + b^{(1)}_1),
\]
which gives
\[
\df{x_1^{(1)}}{x_1^{(0)}} = \sigma^\prime(W_{1j}^{(1)} x_j^{(0)} + b^{(1)}_1) W_{11}^{(1)}  \propto W_{11}^{(1)}.
\]
Since this derivate scales with $W_{11}^{(1)}$, this implies that $|\df{\bm{\mathcal{F}}(\x)}{x_1^{(0)}}|$ scales with $W_{11}^{(1)}$ as well. If $|W_{11}^{(1)}| \gg 1$, then network would be very sensitive to even small changes in the input $x_1^{(0)}$, i.e., the network would be ill-posed. As illustrated in Figure \ref{fig:reg}(b), using a proper regularization would help avoid over fitting.

Let us consider some common types of regularization:
\begin{itemize}
\item \textbf{$l_2$ regularization}: Here we use the $l_2$ norm in the regularization term
\[
\| \btheta \| = \|\btheta\|_2 = \left(\sum_{i=1}^{N_\theta} \theta_i^2\right)^{1/2}.
\]	
\item \textbf{$l_1$ regularization}: Here we use the $l_1$ norm in the regularization term
\[
\| \btheta \| = \|\btheta\|_1 = \sum_{i=1}^{N_\theta} |\theta_i|,
\]
which promotes the sparsity of $\btheta$.
\end{itemize}

\section{Gradient descent}
Recall that we wish to solve the minimization problem $\btheta^* = \amin \Pi(\btheta)$ in the training phase. This minimization problem can be solved using gradient descent (GD), also known as steepest descent. Consider the Taylor expansion about $\btheta_0$
\[
\Pi(\btheta_0 + \Delta \btheta) = \Pi(\btheta_0) + \df{\Pi}{\btheta}(\btheta_0) \cdot \Delta \btheta + \df{^2\Pi}{\theta_i\theta_j}(\hat{\btheta})  \Delta \theta_i \Delta \theta_j
\]
for some $\hat{\btheta}$ in a small neighbourhood of $\btheta_0$. When $|\Delta \btheta |$ is small and assuming $\df{^2\Pi}{\theta_i\theta_j}$ is bounded, we can neglect the second order term and just consider the approximation
\[
\Pi(\btheta_0 + \Delta \btheta) \approx \Pi(\btheta_0) + \df{\Pi}{\btheta}(\btheta_0) \cdot \Delta \btheta. 
\]
In order to lower the value of the loss function as much as possible compared to its evaluation at $\btheta_0$, i.e. minimize $\Delta \Pi = \Pi(\btheta_0 + \Delta \btheta) - \Pi(\btheta_0)$, we need to choose the step $\Delta \btheta$ in the opposite direction of the gradient, i.e.:
\[
\Delta \btheta = - \eta \df{\Pi}{\btheta}(\btheta_0)
\]
with the step-size $\eta \geq 0$, also known as the \textit{learning-rate}. This is yet another hyper-parameter that we need to tune during the validation phase. This is the crux of the GD algorithm, and can be summarized as follows:
\begin{enumerate}
\item Initialize $k=0$ and $\btheta_0$
\item While $|\Pi(\btheta_k)| > \epsilon_1$, do
\begin{enumerate}
  \item Evaluate $\df{\Pi}{\btheta}(\btheta_k)$
  \item Update $\btheta_{k+1} = \btheta_k - \eta \df{\Pi}{\btheta}(\btheta_k)$
  \item Increment $k=k+1$
\end{enumerate}
\end{enumerate}

\textbf{Convergence:} Assume that $\Pi(\btheta)$ is convex and differentiable, and its gradient is Lipschitz continuous with Lipschitz constant  $\mathcal{K}$. Then for a $\eta \leq 1/\mathcal{K}$ , the GD updates converges as
\[
\|\btheta^* - \btheta_k\|_2 \leq \frac{C}{k}.
\]

\begin{figure}[htbp!]
\begin{center}
\includegraphics[width=0.5\textwidth]{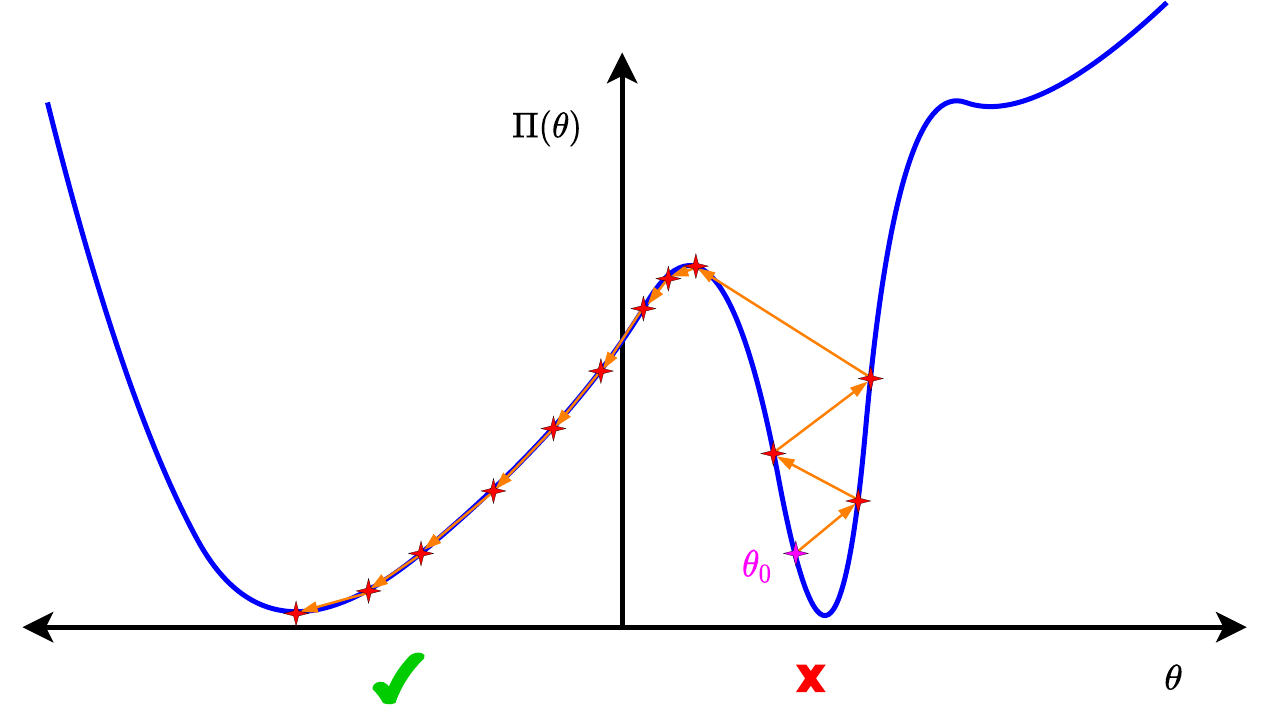}
\caption{GD prefers flatter minimas.}
\label{fig:GD}
\end{center}
\end{figure}

However, in most scenarios $\Pi(\btheta)$ is not convex. If there is more than one minima, then what kind of minima does GD like to pick? To answer this, consider the loss function for a scalar $\theta$ as  shown in Figure \ref{fig:GD}, which has two valleys. Let's assume that the profile of $\Pi(\theta)$ in the each valley can be approximated by a (centered) parabola
\[
\Pi(\theta) \approx \frac{1}{2} a \theta^2
\]
where $a > 0$ is the  curvature of each valley. Note that the curvature of the left valley is much smaller than the curvature of the right valley. Let's pick a constant learning rate $\eta$ and a starting value $\theta_0$ in either of the valleys. Then, 
\[
\df{\Pi}{\theta}(\theta_0) = a \theta_0
\]
and the new point after a GD update will be $\theta_1 = \theta_0(1-a\eta)$. Similarly, it is easy to see that all subsequent iterates write $\theta_{k+1} = \theta_k(1-a\eta)$. For convergence, we need
\[
\left| \frac{\theta_{k+1}}{\theta_k}\right| < 1 \quad \implies |1 - a \eta| < 1.
\]
Since $a>0$ in the valleys, we will need the following condition on the learning rate
\[
-1 < 1 - a \eta \implies a\eta < 2.
\]
If we fix $\eta$, then for convergence we need the local curvature to satisfy $a < 2/\eta$. In other words, GD will prefer to converge to a minima with a flat/small curvature, i.e., it will prefer the minima in the left valley. If the starting point is in the right valley, there is a chance that we will keep overshooting the right minima and bounce off the opposite wall till the GD algorithm slingshots $\theta_k$ outside the valley. After this it will enter the left valley with a smaller curvature and gradually move towards its minima.

While it is clear that GD prefers flat minima, what is not clear is why are flat minima better. There is empirical evidence that the parameter values obtained at flat minima tend to generalize better, and therefore are to be preferred. 

\section{Some advanced optimization algorithms}
We discussed how GD can be used to solve the optimization problem involved in training neural networks. Let us look at a few advanced and popular optimization techniques motivated by GD. 

\begin{figure}[htbp!]
\begin{center}
\includegraphics[width=0.5\textwidth]{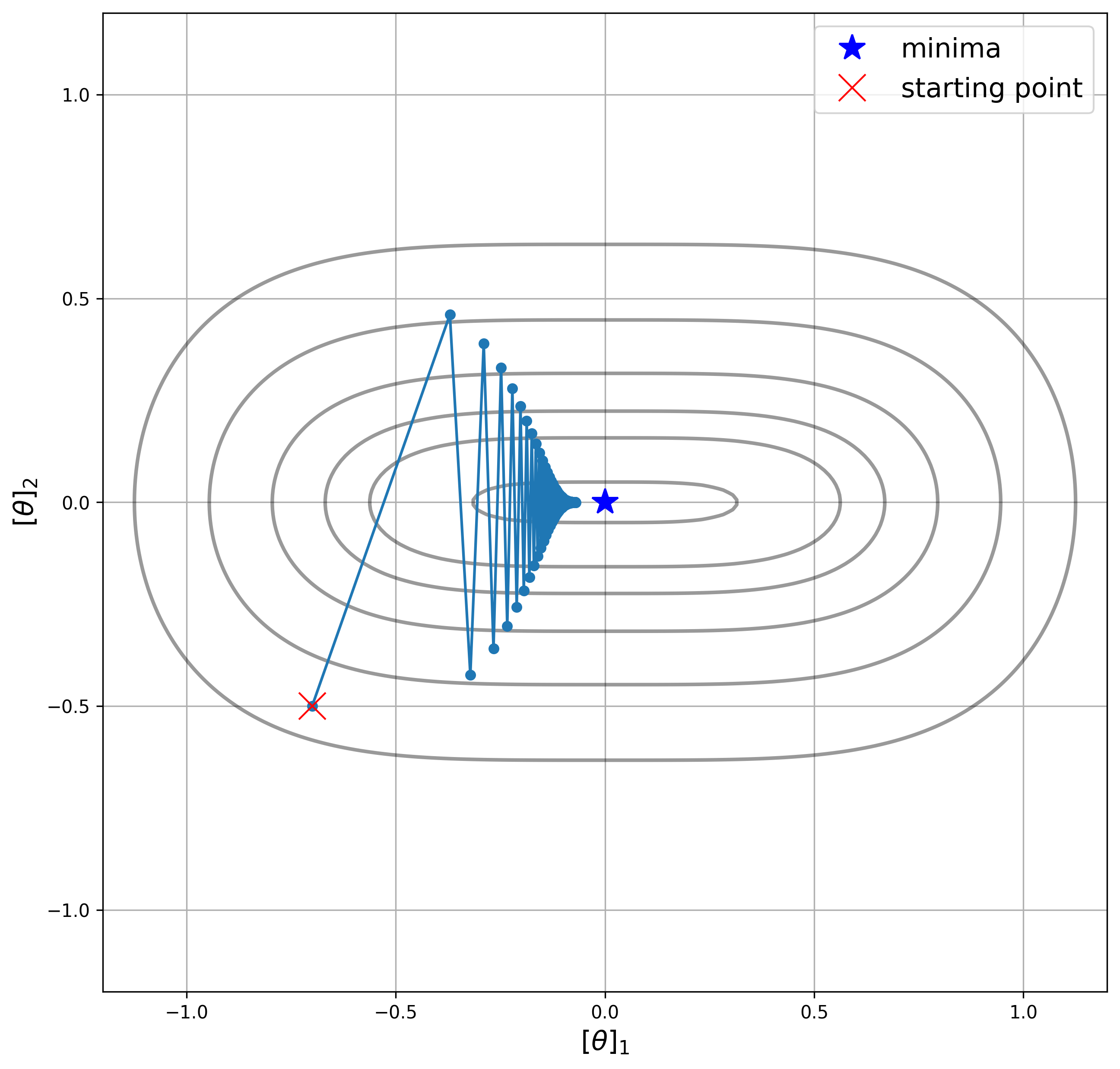}
\caption{Zig-zagging updates with GD.}
\label{fig:gd_zigzag}
\end{center}
\end{figure}

In general, the update formula for most optimization algorithms make use of the following formula
\begin{equation}
[\btheta_{k+1}]_i = [\btheta_{k}]_i - [\bm{\eta}_k ]_i [\g_k]_i, \quad 1 \leq i \leq N_\theta,
\end{equation}
where $[\bm{\eta}_k]_i$ is the component-wise learning rate and the vector-valued function $\g$ depends/approximates the gradient. Note that the notation $[.]_i$ is used to denote the $i$-th component of the vector. Also note that the learning rate is allowed to depend on the iteration number $k$. The GD method makes use of 
\[
[\bm{\eta}_k ]_i = \eta, \quad \g_k = \df{\Pi}{\btheta}(\btheta_k).
\]
An issue with the GD method is that the convergence to the minima can be quite slow if $\eta$ is not suitably chosen. For instance, consider the objective function landscape shown in Figure \ref{fig:gd_zigzag}, which has sharper gradients along the $[\theta]_2$ direction compared to the $[\theta]_1$ direction. If we start from a point, such as the one shown in the figure, then if $\eta$ is too large (but still within the stable bounds) the updates will keep zig-zagging its way towards the minima. Ideally, for the particular situation shown in Figure \ref{fig:gd_zigzag}, we would like the steps to take longer strides along the $[\theta]_1$ compared to the $[\theta]_2$ direction, thus reaching the minima faster. 

Let us look at two popular methods that are able to overcome some of the issues faced by GD.

\subsection{Momentum methods}
Momentum methods make use of the history of the gradient, instead of just the gradient at the previous step. The formula for the update is given by
\[
[\bm{\eta}_k ]_i = \eta, \quad \g_k = \beta_1 \g_{k-1} + (1-\beta_1) \df{\Pi}{\btheta}(\btheta_k), \quad \g_{-1} = 0
\]
where $\g_k$ is a weighted moving average of the gradient. This weighting is expected to smoothen out the zig-zagging seen in Figure \ref{fig:gd_zigzag} by cancelling out the components of gradient along the $[\theta]_2$ direction and move more smoothly towards the minima. A commonly used value for $\beta_1$ is 0.9. 

\subsection{Adam}
The Adam optimization was introduced by Kingma and Ba \cite{kingma2017adam}, which makes use of the history of the gradient as well the second moment (which is a measure of the magnitude) of the gradient. For an initial learning rate $\eta$, the updates are given by
\begin{equation}
\begin{aligned}
\g_k &= \beta_1 \g_{k-1} +  (1-\beta_1) \df{\Pi}{\btheta}(\btheta_k) \\
[\G_k]_i &= \beta_2[\G_{k-1}]_i + (1-\beta_2) \left(\df{\Pi}{\btheta_i}(\btheta_k)\right)^2\\
[\bm{\eta}_k ]_i &=\frac{\eta}{\sqrt{[\G_k]_i} + \epsilon} 
\end{aligned}
\end{equation}
where $\g_k$ and $\G_k$ are the weighted running averages of the gradients and the square of the gradients, respectively. The recommended values for the hyper-parameters are $\beta_1 = 0.9$, $\beta_2 = 0.999$ and $\epsilon=10^{-8}$. Note that the learning rate for each component is different. In particular, the larger the magnitude of the gradient for a component the smaller is its learning rate. Referring back to the example in Figure \ref{fig:gd_zigzag}, this would mean a smaller learning rate for $\theta_2$ in comparison to $\theta_1$, and therefore will help alleviate the zig-zag path of the optimization algorithm. 

\begin{remark}
The Adam algorithm also has additional correction steps for $\g_k$ and $\G_k$ to improve the efficiency of the algorithm. See \cite{kingma2017adam} for details.
\end{remark}

\subsection{Stochastic optimization}
\begin{figure}[htbp!]
\begin{center}
\subfigure[Function contours and paths]{\includegraphics[width=0.45\textwidth]{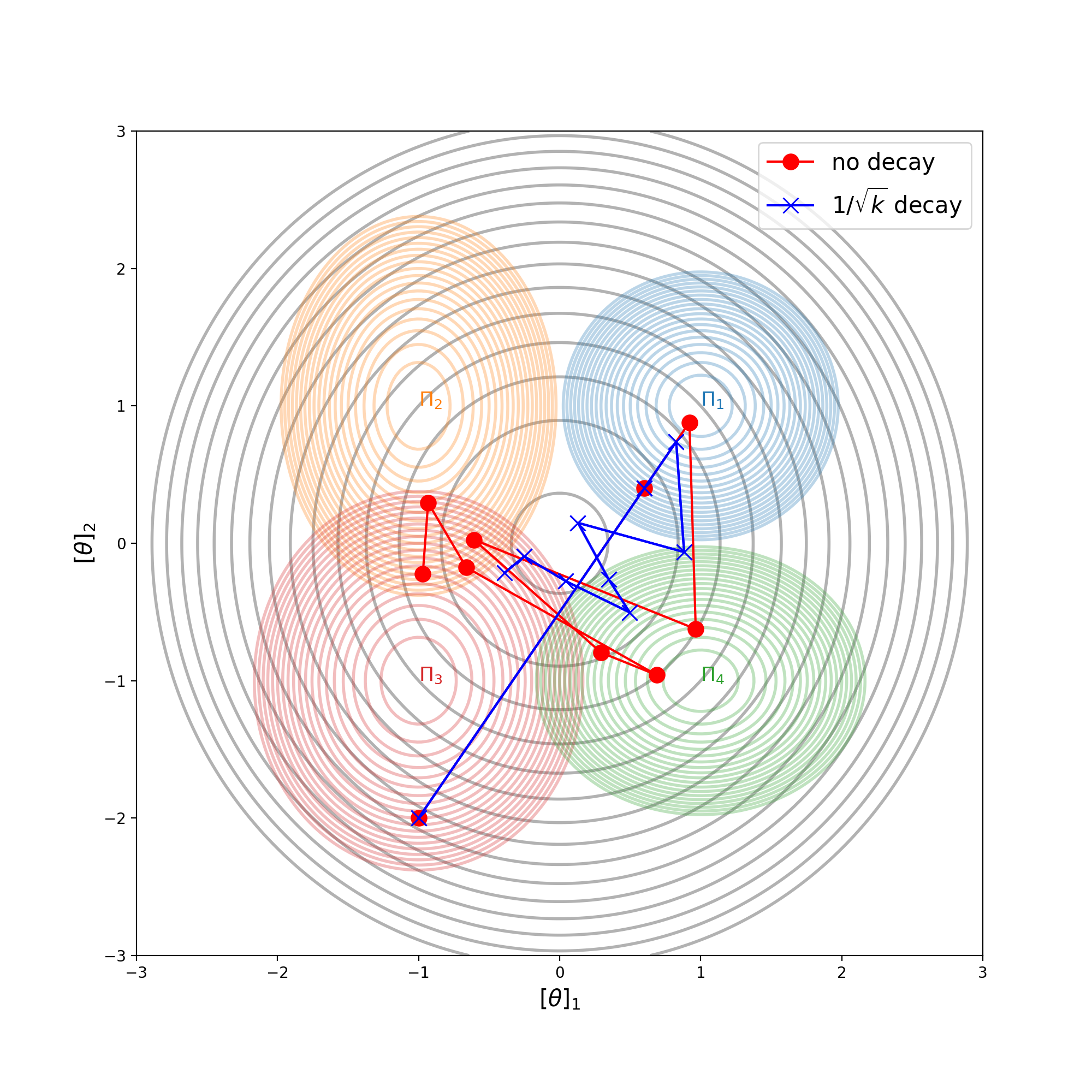}}
\subfigure[Norm of updates]{\includegraphics[width=0.45\textwidth]{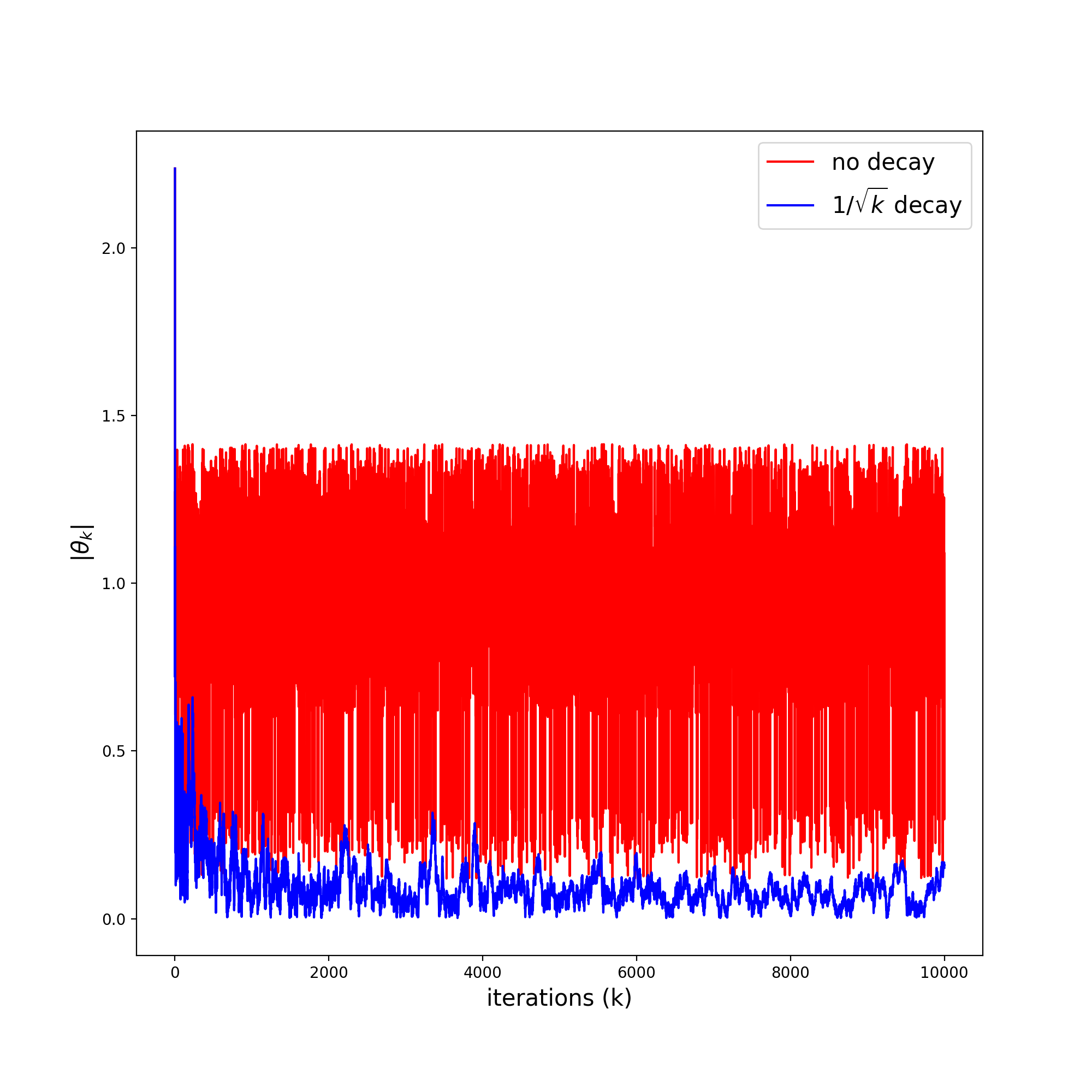}}
\caption{SGD algorithm with and without a decay in the learning rate.}
\label{fig:sgd}
\end{center}
\end{figure}

We note that the training loss can be rewritten as
\[
\Pi(\btheta) = \frac{1}{N_\text{train}} \sum_{i=1}^{N_\text{train}} \Pi_i(\btheta), \quad \Pi_i(\btheta) = \|\y_i - \bm{\mathcal{F}}(\x_i; \btheta, \Hp)\|^2
\]
Thus, the gradient of the loss function is
\[
\df{\Pi}{\btheta}(\btheta) = \frac{1}{N_\text{train}} \sum_{i=1}^{N_\text{train}}  \df{\Pi_i}{\btheta}(\btheta)
\]
However, taking the summation of gradients can be very expensive since $N_\text{train}$ is typically very large, $N_\text{train} \sim 10^6$. One easy way to circumvent this problem is to use the following update formula (shown here for the GD method)
\begin{equation}\label{eqn:sgd}
\btheta_{k+1} = \btheta_k - \eta_k \df{\Pi_i}{\btheta}(\btheta_k),
\end{equation}
where $i$ is randomly chosen for each update step $k$. This is known as  \textit{stochastic gradient descent}. Remarkably, this modified algorithm does converge assuming that $\Pi_i(\btheta)$ is convex and differentiable, and $\eta_k \sim 1/\sqrt{k}$ \cite{nemirovski90}. To illustrate why $\eta_k$ needs to decay, consider the toy function(s) for $\btheta \in \Ro^2$
\begin{equation}
\begin{aligned}
\Pi_1(\btheta) &= ([\theta]_1 - 1)^2 + ([\theta]_2 - 1)^2, \quad \Pi_2(\btheta) = ([\theta]_1 + 1)^2 + 0.5([\theta]_2 - 1)^2,\\
\Pi_3(\btheta) &= 0.7([\theta]_1 +1)^2 + 0.5([\theta]_2 +1)^2, \quad \Pi_4(\btheta) = 0.7([\theta]_1 - 1)^2 + \frac{1}{2}([\theta]_2 + 1)^2,\\
\Pi(\btheta)& = \frac{1}{4} \left( \Pi_1(\btheta) + \Pi_2(\btheta) + \Pi_3(\btheta) + \Pi_4(\btheta)\right).
\end{aligned}
\end{equation}
The contour plots of these functions in shown in Figure \ref{fig:sgd}(a), where the black contour plots corresponds to $\Pi(\btheta)$.  Note that the $\btheta^* = (0,0)$ is the unique minima for $\Pi(\btheta)$. We consider solving with the SGD algorithm with a constant learning rate $\eta_k=0.4$ and a decaying learning rate $\eta_k = 0.4/\sqrt{k}$. Starting with $\btheta_0=(-1.0,2.0)$ and randomly selecting $i \in {1,2,3,4}$ for each step $k$, we run the algorithm for 10,000 iterations. The first 10 steps with each learning rate is plotted in Figure \ref{fig:sgd}(a). We can clearly see that without any decay in the learning rate, the SGD algorithm keeps overshooting the minima. In fact, this behaviour continues for all future iterations as can be seen in Figure \ref{fig:sgd}(b) where the norm of the updates does not decay (we expect it to decay to $|\btheta^*| = 0$). On the other hand, we quickly move closer to $\btheta^*$ if the learning rate decays as $1/\sqrt{k}$. 

The reason for reducing the step size as we approach closer to the minima is that far away from the minima for $\Pi$ the gradient vector for $\Pi$ and all the individual $\Pi_i$'s align quite well. However, as we approach closer to the minima for $\Pi$ this is not the case and therefore one is required to take smaller steps so as not be thrown off to a region far away from the minima.

In practice, stochastic optimization algorithms are not used for the following reasons:
\begin{enumerate}
\item Although the loss function decays with the number of iterations, it fluctuates in a chaotic manner close the the minima and never manages to reach the minima.
\item While handling all samples at once can be computationally expensive, handling a single sample at a time severly under-utilizes the computational and memory resources.
\end{enumerate}
However, a compromise can be made by using \textit{mini-batch optimization}. In this strategy, the dataset of $N_\text{train}$ samples is split into $N_\text{batch}$ disjoint subsets known as mini-batches. Each mini-batch contains $\overline{N}_\text{train} = N_\text{train}/N_\text{batch}$ samples, which also refered to as the batch-size. Thus, the gradient of the loss function can be approximated by 
\begin{equation}\label{eqn:mbs_grad}
\df{\Pi}{\btheta}(\btheta) = \frac{1}{N_\text{train}} \sum_{i=1}^{N_\text{train}}  \df{\Pi_i}{\btheta}(\btheta) \approx \frac{1}{\overline{N}_\text{train}} \sum_{i \in \text{batch}(j)}  \df{\Pi_i}{\btheta}(\btheta).
\end{equation}
Note that taking $N_\text{batch} = 1$ leads to the original optimization algorithms, while take $N_\text{batch} = N_\text{train}$ gives the stochastic gradient descent algorithm. One typically chooses a batch-size to maximize the amount of data that can be loaded into the RAM at one time. We define an \textit{epoch} as one full pass through all samples (or mini-batches) of the full training set. The following describes the \text{mini-batch stochastic optimization} algorithm:
\begin{enumerate}
\item For epoch = 1, ..., J
  \begin{enumerate}
     \item Randomly shuffle the full training set
     \item Create $N_\text{batch}$ mini-batches
     \item For $ i = 1, \cdots, N_\text{batch}$
     \begin{enumerate}
        \item Evaluate the batch gradient using \eqref{eqn:mbs_grad}.
        \item Update $\btheta$ using this gradient and your favorite optimization algorithm (gradient descent, momentum, or Adam).
      \end{enumerate}   
   \end{enumerate}
\end{enumerate}

\begin{remark}
There is an interesting study \cite{WuSGD} that suggests that stochastic gradient descent might actually help in selecting minima that generalize better. In that study the authors prove that SGD prefers minima whose curvature is more homogeneous. That is, the distribution of the curvature of each of the components of the loss function is sharp and centered about a small value. This is contrast to minima where the overall curvature might be small; however the distribution of the curvature of each component of loss function is more spread out. Then they go on to show (empirically) that the more homogeneous minima tend to generalize better than their heterogeneous counterparts. 

\end{remark}

\section{Calculating gradients using back-propagation} \label{sec:backprop}
\begin{figure}[htbp!]
\begin{center}
\includegraphics[width=0.90\textwidth]{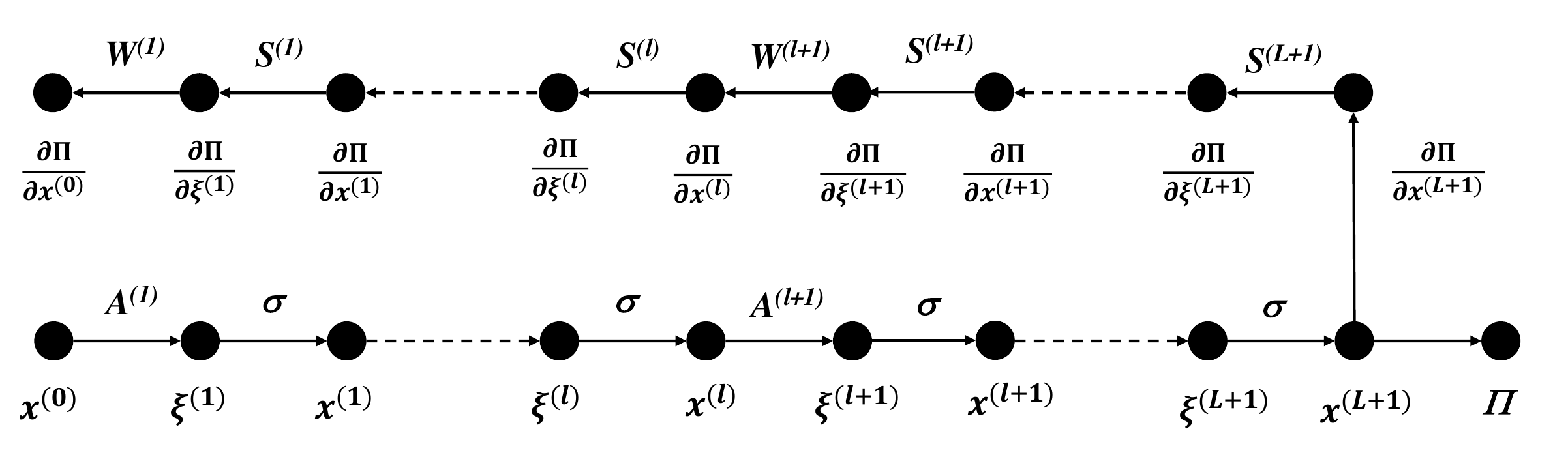}
\caption{Computational graph for computing the loss function and its derivatives with respect to hidden/latent vectors.}
\label{fig:comp_graph}
\end{center}
\end{figure}

The final piece of the training algorithm that we need to understand is how the gradients are actually evaluated while training the network. Recall the output $\x^{(l+1)}$ of layer $l+1$ is given by
\begin{align}
\text{Affine transform:} & \quad \xi^{(l+1)}_i= W_{ij}^{(l+1)} x_j^{(l)}+ b_i^{(l+1)}, \quad 1 \leq i \leq H_{l+1} \label{eqn:at}\\
\text{Non-linear transform:} & \quad x^{(l+1)}_i = \sigma\left(\xi^{(l+1)}_i\right), \quad 1 \leq i \leq H_{l+1} \label{eqn:nlt}. 
\end{align}
Given a training sample $(\x,\y)$, set $\x^{(0)} = \x$. The value of the loss/objective function (for this particular sample) can be evaluated using the forward pass:
\begin{enumerate}
\item For $l = 1, ..., L+1$
  \begin{enumerate}
     \item Evaluate $\xi^{(l)}$ using \eqref{eqn:at}.
     \item Evaluate $\x^{(l)}$ using \eqref{eqn:nlt}.
  \end{enumerate}
\item Evaluate the loss function for the given sample 
\[
\Pi(\btheta) =\|\y - \bm{\mathcal{F}}(\x; \btheta, \Hp)\|^2.
\]   
\end{enumerate}
This operation can be written succinctly in the form of a computational graph as shown in Figure \ref{fig:comp_graph}. In this figure, the lower portion of the graph represents the evaluation of the loss function $\Pi$. 

We would of course need to repeat this step for all samples in the training set (or a mini-batch for stochastic optimization). For simplicity, we restrict the discussion to the evaluation of the loss and its gradient for a single sample.

In order to update the network parameters, we need $\df{\Pi}{\btheta}$, or more precisely $\df{\Pi}{\W^{(l)}}, \df{\Pi}{\bb^{(l)}}$ for $1\leq l \leq {L+1}$. We will derive expressions for these derivatives by first deriving expressions for $\df{\Pi}{\bm{\xi}^{(l)}}$ and $\df{\Pi}{\bm{\x}^{(l)}}$. 

From the computational graph it is easy to see how each hidden variable in the network is transformed to the next. Recognizing this, and applying the chain rule repeatedly yields

\begin{align}
\df{\Pi}{\bm{\xi}^{(l)}}  &=  \label{eqn:dpidxi} \df{\Pi}{\bm{\x}^{(L+1)}} \cdot \df{\bm{\x}^{(L+1)}}{\bm{\xi}^{(L+1)}} \cdot \df{\bm{\xi}^{(L+1)}}{\bm{\x}^{(L)}} \cdots \df{\bm{\x}^{(l+1)}}{\bm{\xi}^{(l+1)}} \cdot \df{\bm{\xi}^{(l+1)}}{\bm{\x}^{(l)}} \cdot \df{\bm{\x}^{(l)}}{\bm{\xi}^{(l)}}. 
\end{align}
In order to evaluate this expression we need to evaluate the following terms:
\begin{align}
\df{\Pi}{\bm{\x}^{(L+1)}} &= -2 (\y-\x^{(L+1)})^T  \\
\df{\bm{\xi}^{(l+1)}}{\bm{\x}^{(l)}}  &= \W^{(l+1)}\\
\df{\bm{\x}^{(l)}}{\bm{\xi}^{(l)}}  &= \bm{S}^{(l)} \equiv {\rm diag}[\sigma'(\xi^{(l)}_1), \cdots, \sigma'(\xi^{(l)}_{H_l})],
\end{align}
where the last two relations hold for any network layer $l$, $H_l$ is the width of that particular layer, and $\sigma'$ denotes the derivative of the activation with respect to its argument. Using these relations in (\ref{eqn:dpidxi}), we arrive at,
\begin{align}
\df{\Pi}{\bm{\xi}^{(l)}}  &=  \label{eqn:dpidxi1} \df{\Pi}{\bm{\x}^{(L+1)}} \cdot \bm{S}^{(L+1)} \cdot \W^{(L+1)}  \cdots \bm{S}^{(l+1)} \cdot \W^{(l+1)} \cdot \bm{S}^{(l)}. 
\end{align}
Taking the transpose, and recognizing that $\bm{\Sigma}^{(l)}$ is diagonal and therefore symmetric, we finally arrive at
\begin{align}
\df{\Pi}{\bm{\xi}^{(l)}}  &=  \label{eqn:dpidxifinal}  \bm{S}^{(l)} \W^{(l+1)T} \bm{S}^{(l+1)} \cdots   \W^{(L+1)T} \bm{S}^{(L+1)} [-2 (\y-\x^{(L+1)})]. 
\end{align}
This evaluation can also be represented as a computational graph. In fact, as shown in Figure \ref{fig:comp_graph}, it can be appended to the original graph, where this part of the computation appear in the upper row of the graph. Note that we are now traversing in the backward direction.  Hence the name back propagation. 

The final step is to evaluate an explicit expression for $\df{\Pi}{\W^{(l)}}$. This can be done by recognizing,
\begin{align}
\df{\Pi}{\W^{(l)}} &= \df{\Pi}{\bm{\xi}^{(l)}} \cdot \df{ \bm{\xi}^{(l)} }{\bm{W}^{(l)}}  = \df{\Pi}{\bm{\xi}^{(l)}} \otimes \x^{(l-1)}, 
\label{eqn:bpr1}
\end{align}
where $ [\x \otimes \y]_{ij} = x_iy_j$  is the outer product. Thus, in order to evaluate $\df{\Pi}{\W^{(l)}}$ we need $\x^{(l-1)}$ which is evaluted during the forward phase and $\df{\Pi}{\bm{\xi}^{(l)}}$ which is evaluated during back propagation. 




\begin{question}
Can you derive a similar set of expressions and the corresponding algorithm to evaluate $\df{\Pi}{\bb^{(l)}}$?
\end{question}

\begin{question}
Can you derive an explicit expression for $\df{\x^{(L+1)}}{\x^{(0)}}$. That is the an expression for the derivative of the output of the network with respect to its input? This is a very useful quantity that finds use in algorithms like physics informed neural networks and Wasserstein generative adversarial networks. 
\end{question}

\section{Regression versus classification}
Till now, given the labelled dataset $\mathcal{S} = \{(\x_i,\y_i): 1 \leq i \leq N\}$, we have considered two types of losses
\begin{itemize}
\item The mean square error (MSE)
\[
\Pi(\btheta) =  \frac{1}{N_\text{train}} \sum_{i=1}^{N_\text{train}}\|\y_i - \bm{\mathcal{F}}(\x_i; \btheta, \Hp)\|^2.
\]
\item The mean absolute error (MAE)
\[
\Pi(\btheta) =  \frac{1}{N_\text{train}} \sum_{i=1}^{N_\text{train}}\|\y_i - \bm{\mathcal{F}}(\x_i; \btheta, \Hp)\|.
\]
\end{itemize}

Neural networks with the above losses can be used to solve various regression problems where the underlying function is highly nonlinear and the inputs/outputs are multi-dimensional.
\begin{example}
Given the house/apartment features such as the zip code, the number of bedrooms/bathrooms, carpet area, age of construction, etc, predict the outcomes such as the market selling price, or the number of days on the market.
\end{example}

Now let us consider some examples of classification problems, where the output of the network typically lies in a discrete finite set.
\begin{example}
Given the symptoms and blood markers of patients with COVID-19, predict whether they will need to be admitted to ICU. So the input and output for this problem would be
\begin{equation*}
\begin{aligned}
\x &= [\text{pulse rate}, \text{temperature}, \text{SPO}_2, \text{procalcitonin}, ... ]\\
\y &= [p_1,p_2]
\end{aligned}
\end{equation*}
where $p_1$ is the probability of being admitted to the ICU, while $p_2$ is the probability of not being admitted. Note that $0 \leq p_1, p_2 \leq1$ and $p_1 + p_2 = 1$.
\end{example}

\begin{example}\label{ex:3class}
Given a set of images of animals, predict whether the animal is a dog, cat or bird. In this case, the input and output should be
\begin{equation*}
\begin{aligned}
\x &= \text{the image}\\
\y &= [p_1,p_2,p_2]
\end{aligned}
\end{equation*}
where $p_1,p_2,p_3$ is the probability of being a dog, cat or bird, respectively.
\end{example}

Since the output for the classification problem corresponds to probabilities, we need to make a few changes to the network
\begin{enumerate}
\item Make use of an output function at the end of the output layer that suitably transforms the output vector into the desired form, i.e, a vector of probabilities. This is typically done using the \textit{softmax function}
\[
x_i^{(L+1)} = \frac{\exp{(\xi_i^{(L+1)})}}{\sum_{j=1}^C\exp{(\xi_j^{(L+1)})}}
\]
where $C$ is the number of classes (and also the output dimension). Verify that with this transformation, the components of the $\x^{(L+1)}$ form a convex combination, i.e., $x_i^{(L+1)} \in [0,1]$ and $\sum_{i=1}^C x_i^{(L+1)} = 1$.
\item The output labels for the various samples need to be one-hot encoded. In other words, for the sample $(\x,\y)$, the output label $\y$ should have dimension $D=C$, and whose component is 1 only for the component signifying the class $\x$ belongs  to, otherwise 0. For instance, in Example \ref{ex:3class} 
\[
\y = \begin{cases}
[1,0,0]^\top & \quad \text{if $\x$ is a dog},\\ 
[0,1,0]^\top & \quad \text{if $\x$ is a cat},\\
[0,0,1]^\top & \quad \text{if $\x$ is a pig}.
\end{cases}
\]

\item Although the MSE or MSA can still be used as the loss function, it is preferable to use the \text{cross-entropy} loss function
\begin{equation}\label{eqn:crossentropy}
\Pi(\btheta) = \frac{1}{N_\text{train}} \sum_{i=1}^{N_\text{train}} \sum_{c=1}^{C} - y_{ci} \log(\mathcal{F}_c(\x_i;\btheta)),
\end{equation}
where $y_{ci}$ is the $c$-th component of the true label for the $i$-th sample.
The loss function in \eqref{eqn:crossentropy} treats $y_c$ and $\mathcal{F}_c$ as probability distributions and measures the discrepancy between the two. It can be shown to be related to the Kullback-Liebler divergence between the two distributions. Compared to MSE, this loss function severely penalizes strongly confident incorrect predictions. This is demonstrated in Example \ref{ex:cr_ent}
\end{enumerate}

\begin{example}\label{ex:cr_ent}
Let us consider a binary classification problem, i.e., $C=2$. For a given $\x$, let $\y = [0,1]$ and let the prediction be $\bm{\mathcal{F}} = [p, 1-p]$. Clearly, a small value of $p$ is preferred. Therefore any reasonable cost function should penalize large values of $p$. Now let us evaluate the error using various loss functions
\begin{itemize}
\item MSE Loss $= (0-p)^2 + (1 - 1 +p)^2 = 2p^2$.
\item Cross-entropy Loss $= -(0 \log(p) + 1 \log(1-p) = - log(1-p)$.
\end{itemize}
Note that both losses penalize large values of $p$. Also when $p=0$, both losses are zero. However, as $p \rightarrow 1$ (which would lead the wrong prediction), the MSE loss $\rightarrow 2$, while the cross-entropy loss $\rightarrow \infty$. That is, it strongly penalizes incorrect confident predictions. 
\end{example}

%% file: ResNets.tex
\newcommand{\red}[1]{\textcolor{red}{#1}}

\chapter{Residual neural networks}

Residual networks (or ResNets) were introduced by He et al. \cite{he2015deep} in 2015. In this chapter, we will discuss what these networks are, why they were introduced and their relation to ODEs.

\section{Vanishing gradients in deep networks}
While training neural networks, the gradients $\df{\Pi}{\W^{(l)}}, \df{\Pi}{\bb^{(l)}}$ might become very small. 
For instance, consider a very deep network, say $L \geq 20$. If $\left|\df{\Pi}{\W^{(l)}}\right| \ll 1$ for $l \le \bar{l}$, then the contribution of first $\bar{l}$ layers of the network will be negligible, as the influence of their weights on the loss function is small. Because of this depth cut-off, the benefit in terms of expressivity of deep networks is lost.

So why does this happen? Recall from Section \ref{sec:backprop} that 
\[
\df{\Pi}{\W^{(l)}} = \df{\Pi}{\bm{\xi}^{(l)}} \otimes \x^{(l-1)}
\]
and 
\begin{align}
\df{\Pi}{\bm{\xi}^{(l)}}  &=  \label{eqn:dpidxifinal1}  \bm{\Sigma}^{(l)}   \prod_{m=l+1}^{L+1} 
 ( \W^{(m)T} \bm{\Sigma}^{(m)}) \df{\Pi}{\bm{\xi}^{(L+1)}} .
\end{align}
%
For any matrix, $\bm{A}$, let $\tau (\bm{A})$ denote the largest singular value. Then we can bound $|\df{\Pi}{\bm{\xi}^{(l)}}|$ by 
\begin{align}
|\df{\Pi}{\bm{\xi}^{(l)}}|  &\le  \label{eqn:bounddpidxi1}  \tau(\bm{\Sigma}^{(l)})   \prod_{m=l+1}^{L+1} 
 ( \tau(\W^{(m)}) \tau(\bm{\Sigma}^{(m)}))  |\df{\Pi}{\bm{\xi}^{(L+1)}}|. 
\end{align}

Recall that $\bm{\Sigma}^{(m)} \equiv {\rm diag}[\sigma'(\xi^{(m)}_1), \cdots, \sigma'(\xi^{(m)}_{H_l})]$, where $\sigma'$ denotes the derivative of $\sigma$ with respect to its argument. For ReLU its value is either $0$ or $1$. Therefore $\tau(\bm{\Sigma}^{(m)})) = 1$. 

Also, for stability we would want $\tau(\W^{(m)}) < 1$. Otherwise the output of the network can become unbounded. In practise this is enforced by the regularization term. 

Using this in the equation above we have 
\begin{align}
|\df{\Pi}{\bm{\xi}^{(l)}}|  &\le  \label{eqn:bounddpidxi2}     \prod_{m=l+1}^{L+1} 
 ( \tau(\W^{(m)}))  |\df{\Pi}{\bm{\xi}^{(L+1)}}|,
\end{align}
where each term in the product is a scalar less than 1. As the number of terms increases, that is $L-l \gg 1$, this product can, and does, become very small. This typically happens when $L-l \approx 20$, in which case $|\df{\Pi}{\bm{\xi}^{(l)}}| $, and therefore $|\df{\Pi}{\W^{(l)}}|$, become very small. This issue is called the problem of vanishing gradients. It manifests itself in deep networks where the weights in the inner layers (say $L-l > 20$) do not contribute to the network. 

\begin{figure}[htbp!]
\begin{center}
\includegraphics[width=0.8\textwidth]{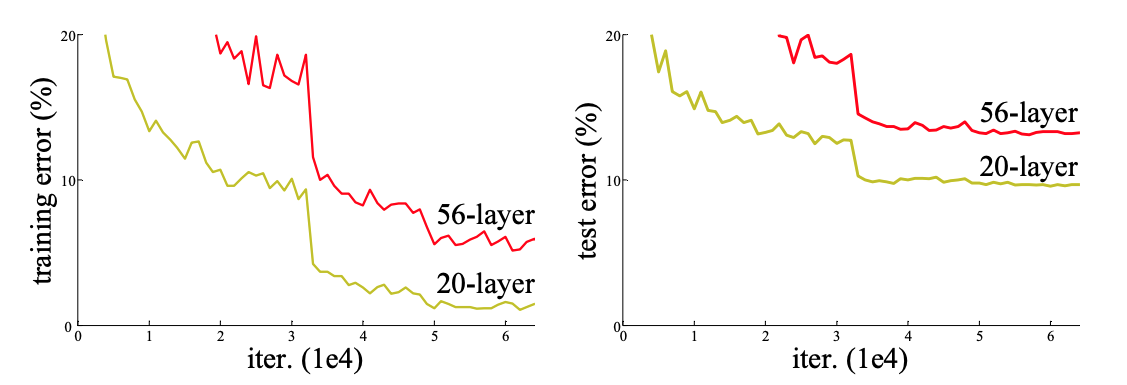}
\caption{Training error (left) and test error (right) on CIFAR-10 data with ``plain" deep networks (taken from \cite{he2015deep}).}
\label{fig:resnet_paper1}
\end{center}
\end{figure}

In \cite{he2015deep}, the authors demonstrate that taking a deeper network can actually lead to an increase in training and validation error (see Figure \ref{fig:resnet_paper1}). Thus, beyond a certain point, increasing the depth of a network can be counterproductive. Based on our previous discussion on vanishing gradients we know why this is the case. Given this, we would like to come up with a network architecture that addresses the problem of vanishing gradients by ensuring $\left| \df{\Pi}{\bm{\xi}^{(L+1)}} \right| \approx \left| \df{\Pi}{\bm{\xi}^{(1)}} \right|$. This means requiring that when the weights of the network approach small values, the network should approach the identity mapping, and not the null mapping. This is the core idea behind a ResNet architecture.

\section{ResNets}

\begin{figure}[htbp!]
\begin{center}
\includegraphics[width=\textwidth]{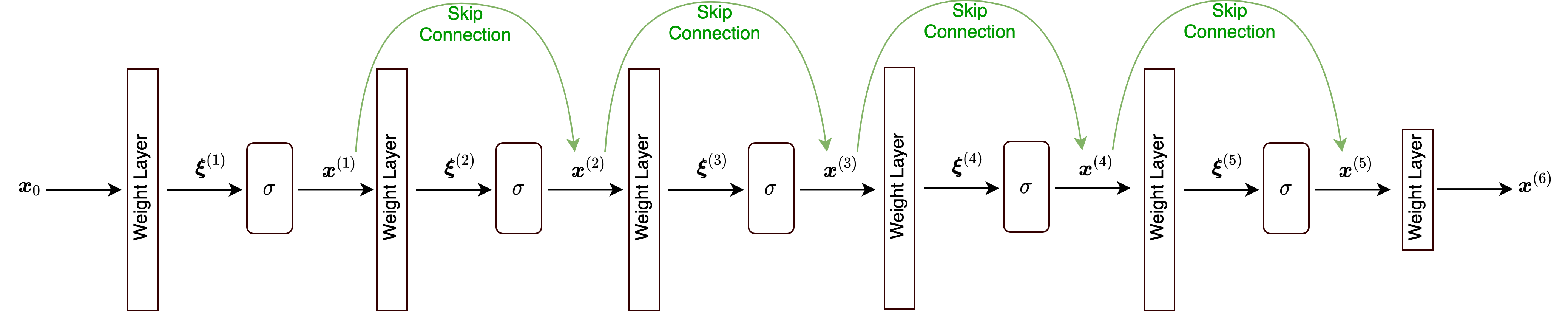}
\caption{ResNet of depth 6 with skip connections.}
\label{fig:resnet}
\end{center}
\end{figure}

Consider an MLP with depth 6 (as shown in Figure \ref{fig:resnet}) with a fixed width $H$ for each hidden layer. We add skip connections between the hidden layers in the following manner
\begin{equation}\label{eqn:resnet}
x_i^{(l)} = \sigma(W_{ij}^{(l)} x_j^{(l-1)} + b_i^{(l)}) + x_i^{(l-1)} , \quad 2 \leq l \leq L.
\end{equation}
We can make the following observations:
\begin{enumerate}
\item If all weights (and biases) were null, then $\x^{(5)} = \x^{(1)}$, which in turn would imply
\[
\df{\Pi}{\x^{(1)}} = \df{\Pi}{\x^{(5)}},
\]  
i.e., we will not have the issue of vanishing gradients.

\begin{figure}[htbp!]
\begin{center}
\includegraphics[width=\textwidth]{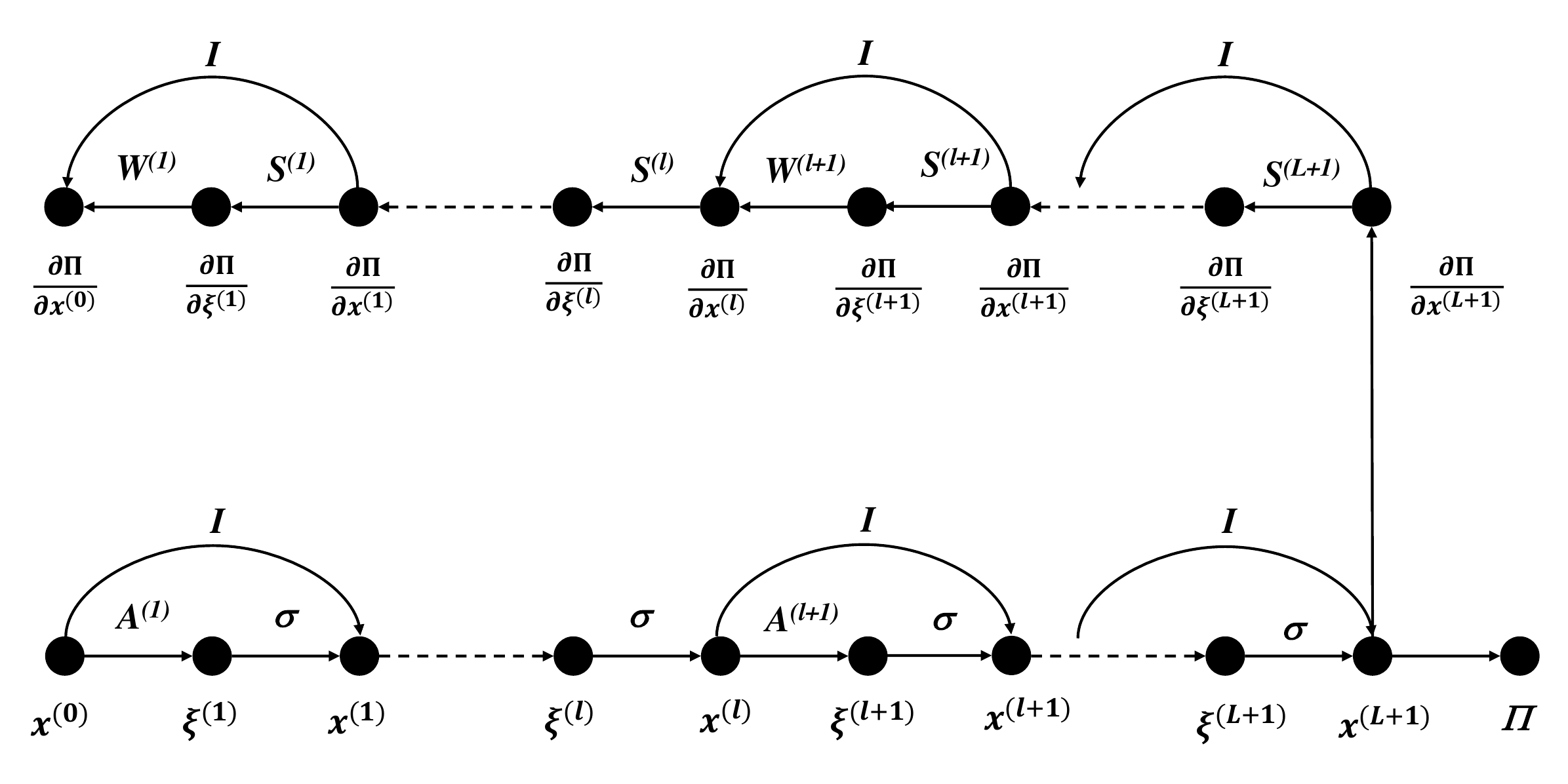}
\caption{Computational graph for forward and backpropagation in a Resnet.}
\label{fig:resnet_graph}
\end{center}
\end{figure}

\item The computational graph for forward and back-propagation of a ResNet is shown in Figure \ref{fig:resnet_graph}. Looking at this graph, it is clear that the expression for $ \df{\x^{(l+1)}}{\x^{(l)}}$ now involves traversing two branches and adding their sum. Therefore, we have 
\begin{align}
\df{\Pi}{\bm{\xi}^{(l)}}  &=  \label{eqn:dpidxiresnet}  \bm{\Sigma}^{(l)}   \prod_{m=l+1}^{L+1} 
 ( \bm{I} + \W^{(m)T} \bm{\Sigma}^{(m)}) \df{\Pi}{\bm{\xi}^{(L+1)}} .
\end{align}
Now, if we assume that $|\W^{(m)}| \ll 1$ via regularization, we have 
\begin{align}
\df{\Pi}{\bm{\xi}^{(l)}}  &=  \label{eqn:dpidxiresnet1}  \bm{\Sigma}^{(l)}    
 \big(\bm{I} + \sum_{m=l+1}^{L+1} \W^{(m) \ T} \bm{\Sigma}^{(m)}  + \text{ higher order terms} \big) \df{\Pi}{\bm{\xi}^{(L+1)}} .
\end{align}
In the expression above, even if the individual matrices have small entries, their sum need not approach a zero matrix. This implies that we can create a finite (and significant) change between the gradients near the input and output layers, while still requiring the weights to be small (via regularization).


\end{enumerate}

\begin{remark}
The above analysis can be extended to cases when $H$ is not fixed, but the analysis is not as clean. See \cite{he2015deep} on how we can do this.
\end{remark}

\section{Connections with ODEs}
Let us first consider the special case of a ResNet with $d=D=H$.  Recall the relation \eqref{eqn:resnet}, which we can rewrite as
\begin{equation}\label{eqn:resnet_ode1}
\frac{ \x^{(l)} -  \x^{(l-1)}}{\Delta t}= \frac{1}{\Delta t} \sigma(\W^{(l)} \x^{(l-1)} + \bm{b}^{(l)}) = \frac{1}{\Delta t} \sigma(\bm{\xi}^{(l)})
\end{equation}
for some scalar $\Delta t$, where we note that $\bm{\xi}^{(l)}$ is a function of $\x^{(l-1)}$ parameterized by $\btheta^{(l)} = [\W^{(l)}, \bb^{(l)}]$. Thus, we can further rewrite \eqref{eqn:resnet_ode1} as
\begin{equation}\label{eqn:resnet_ode2}
\frac{\x^{(l)} -  \x^{(l-1)}}{\Delta t} = \bm{V} (\x^{(l-1)}; \btheta^{(l)}).
\end{equation}

Now consider a first-order system of (possibly non-linear) ODEs, where given $\x(0)$ and
\begin{equation}\label{eqn:ode}
\dot{\x} \equiv \dd{\x}{t} = \V(\x,t) 
\end{equation}
we want to find $\x(T)$. In order to solve this numerically, we can uniformly divide the temporal domain with a time-step $\Delta t$ and temporal nodes $t^{(l)}=l \Delta t$, $0 \leq l \leq L+1$, where $(L+1)\Delta t = T$. Define the discrete solution as
$\x^{(l)} = \x(l\Delta t)$. Then, given $\x^{(l-1)}$, we can use a time-integrator to approximate the solution $\x^{(l)}$.  We can consider a method motivated by the forward Euler integrator, where the the LHS of \eqref{eqn:ode} is approximated by
\[
LHS \approx \frac{\x^{(l)} - \x^{(l-1)}}{\Delta t}.
\]
while the RHS is approximated using a parameter $\btheta^{(l)}$ as
\[
RHS \approx  \V(\x^{(l-1)}; t^{(l)}) = \V(\x^{(l-1)}; \btheta^{(l)}).
\]
where we are allowing the parameters to be different at each time-step.
Putting these two together, we get exactly the relation of the ResNet given in \eqref{eqn:resnet_ode2}. In other words, a ResNet is nothing but a descritization of a non-linear system of ODEs. We make some comments to further strengthen this connection.
\begin{itemize}

\item In a fully trained ResNet we are given $\x^{(0)}$ and the weights of a network, and we predict $\x^{(L+1)} $.

\item In a system of ODEs, we are given $\x(0)$ and $\bm{V}(\x,t)$, and we predict $\x{(T)}$.

\item Training the ResNet means determining the parameters $\btheta$ of the network so that $\x^{(L+1)}$ is as close as possible to $\y_i$ when $\x^{(0)} = \x_i$, for $i = 1, \cdots, N_{\rm train}$. 

\item When viewed from the analogous ODE point of view, training means determining the right hand side $\V(\x,t)$ by requiring $\x(T)$ to be as close as possible to $\y_i$ when $\x(0) = \x_i$, for $i = 1, \cdots, N_{\rm train}$.

\item In a ResNet we are looking for "one" $\bm{V}(\x,t)$ that will map $\x_i$ to $\y_i$, for all $1 \leq i \leq N_\text{train}$.
\end{itemize}

\section{Neural ODEs}

\begin{figure}[htbp!]
\begin{center}
\includegraphics[width=0.6\textwidth]{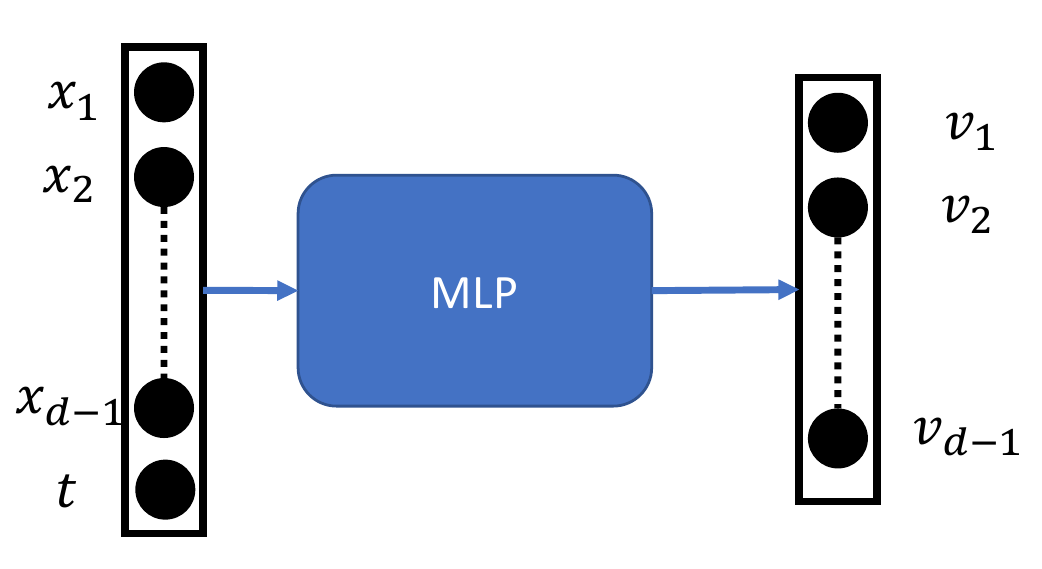}
\caption{Feed-forward neural network used to model the right hand side in a Neural ODE. The number of dependent variables $= d-1$.}
\label{fig:neural_ode}
\end{center}
\end{figure}

Motivated by the connection between ResNets and ODEs, neural ODEs were proposed in \cite{chen2019neural}. 
Consider a system of ODEs given by
\begin{equation}\label{eqn:neural_ODE}
\begin{aligned}
\dd{\x}{t} = \V(\x,t)
\end{aligned}
\end{equation}
Given $\x(0)$, we wish to find $\x(T)$. In \cite{chen2019neural}, the RHS, i.e., $\V(\x,t)$, is defined using a feed-forward neural network with parameters $\btheta$ (see Figure \ref{fig:neural_ode}). The input to the network is $(\x,t)$ while the output is $\V(\x,t)$ (having the same dimension as $\x$). With this description, the system \eqref{eqn:neural_ODE} is solved using a suitable time-marching scheme, such as forward Euler, Runge-Kutta, etc.

So how do we use this network to solve a regression problem? Assume that you are given the labelled training data $\mathcal{S} = \{(\x_i,\y_i): 1 \leq i \leq N_{\rm train}\}$. Here both $\x_i$ and $\y_i$ are assumed to have the same dimension $d-1$. The key idea is to think of $\x_i$ as points in the $d-1$-dimensional space that represent the initial state of the system, and to think of $\y_i$ as points that represent the final state. Then the regression problem becomes finding the RHS of \eqref{eqn:neural_ODE} that will map the initial points to the final points with minimal amount of error. In other words, find the parameters $\btheta$ such that
\[
\Pi(\btheta) = \frac{1}{N} \sum_{i=1}^N |\x_i(T; \btheta) - \y_i |^2
\]
is minimized. Here, $\x_i (T; \btheta)$ denotes the solution (at time $t=T$) to \eqref{eqn:neural_ODE} with $\x(0) = \x_i$ and the RHS  represented by a feed-forward neural network $\V(\x,t; \btheta)$. Note that $\y_i$ is the output value that is measured. There is a relatively straightforward way of extending this approach to the case when $\x_i$ and $\y_i$ have different dimensions. In summary, in Neural ODEs one transforms a regression problem to one of finding the nonlinear, time-dependent RHS of a system of ODEs. 
\begin{figure}[]
\begin{center}
\includegraphics[width=0.9\textwidth]{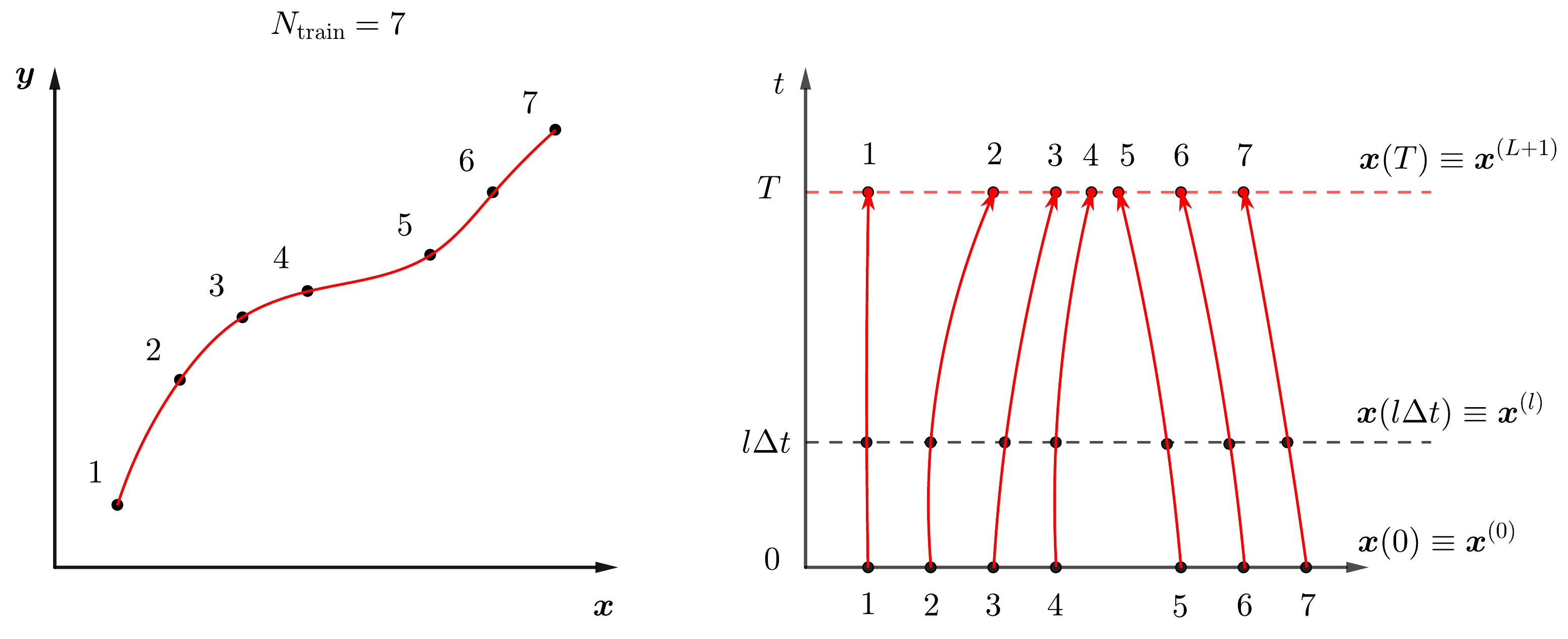}
\caption{Analogy between regression problems and Neural ODEs.}
\label{fig:neural_ode2}
\end{center}
\end{figure}

Let us list the advantages and differences when comparing Neural ODEs to ResNets:
\begin{itemize}
\item If we interpret the number of time-steps in the Neural ODE as the number of hidden layers $L$ in a ResNet, then the computational cost for both methods is $\bigO(L)$. This is the cost associated with performing one forward propagation and one backward propagation. However the memory cost (the cost associated with storing the weights of each layer), is different. For the neural ODE all the weights are associated with the feed-forward network used to represent the function $\V(\x,t; \btheta)$. Thus the number of weights are independent of the number of time-steps used to solve the ODE. On the other hand, for a ResNet the number of weights increases linearly with the number of layers, therefore the cost of storing them scales as $\bigO(L)$. 
\item In Neural ODEs, we can take the limit $\Delta t \rightarrow 0$ and study the convergence, since this will not change the size of the network used to represent the RHS. However, this is not computationally feasible to do for ResNets, where $\Delta t \rightarrow 0$ corresponds to the network depth $L \rightarrow \infty$!
\item ResNet uses a forward Euler type method, but in a Neural ODE one can use any time-integrator. Especially, other higher-order explicit time-integrator like the Runge-Kutta methods that converge to the ``exact'' solution at a faster rate. 

\end{itemize}

%% file: PDEs.tex

\chapter{Solving PDEs with MLPs}\label{chap:PINNs}
A number of numerical methods exist to solve PDEs. Some of these are:
\begin{itemize}
\item Finite difference methods
\item Finite volume methods
\item Finite element methods
\item Spectral Galerkin and collocation methods
\item Deep neural networks!
\end{itemize}

To better appreciate some of these methods, especially deep neural networks, let us consider a simple model problem describing the scalar advection-diffusion problem in one-dimension:\\
Find find $u(x)$ in the interval $x \in (0,l)$ such that
\begin{equation}\label{eqn:adeq}
\begin{aligned}
a \dd{u}{x} - \kappa \dd{^2u}{x^2} &= f(x), \quad x \in (0,\ell)\\
u(0) &=0\\
u(\ell) &=1  
\end{aligned}
\end{equation}
where $a$ denotes the advective velocity, $\kappa$ is the diffusion coefficient while $f(x)$ is the source. Such equations are used to model many physical phenomena, such as the transport of pollutant by fluids, or modelling the flow of electrons through semiconductors. The multi-dimensional version of this problem will take the form 
\begin{equation}\label{eqn:adeq_md}
\begin{aligned}
\bm{a} \cdot \nabla \bm{u}(\bm{s}) - \kappa \Delta \bm{u}(\bm{s}) &= \bm{f}(\bm{s}), \quad \bm{s} \in \Omega\\
\bm{u}(\bm{s}) &= \bm{g}(\bm{s}), \quad \bm{s} \in \partial \Omega
\end{aligned}
\end{equation}
Note that the model problem is a linear PDE (ODE in the one-dimensional case). Replacing the velocity $a$ by $u$ leads to the viscous Burgers equation.

\begin{figure}[htbp!]
\begin{center}
\subfigure[Fixed $\kappa$]{\includegraphics[width=0.45\textwidth]{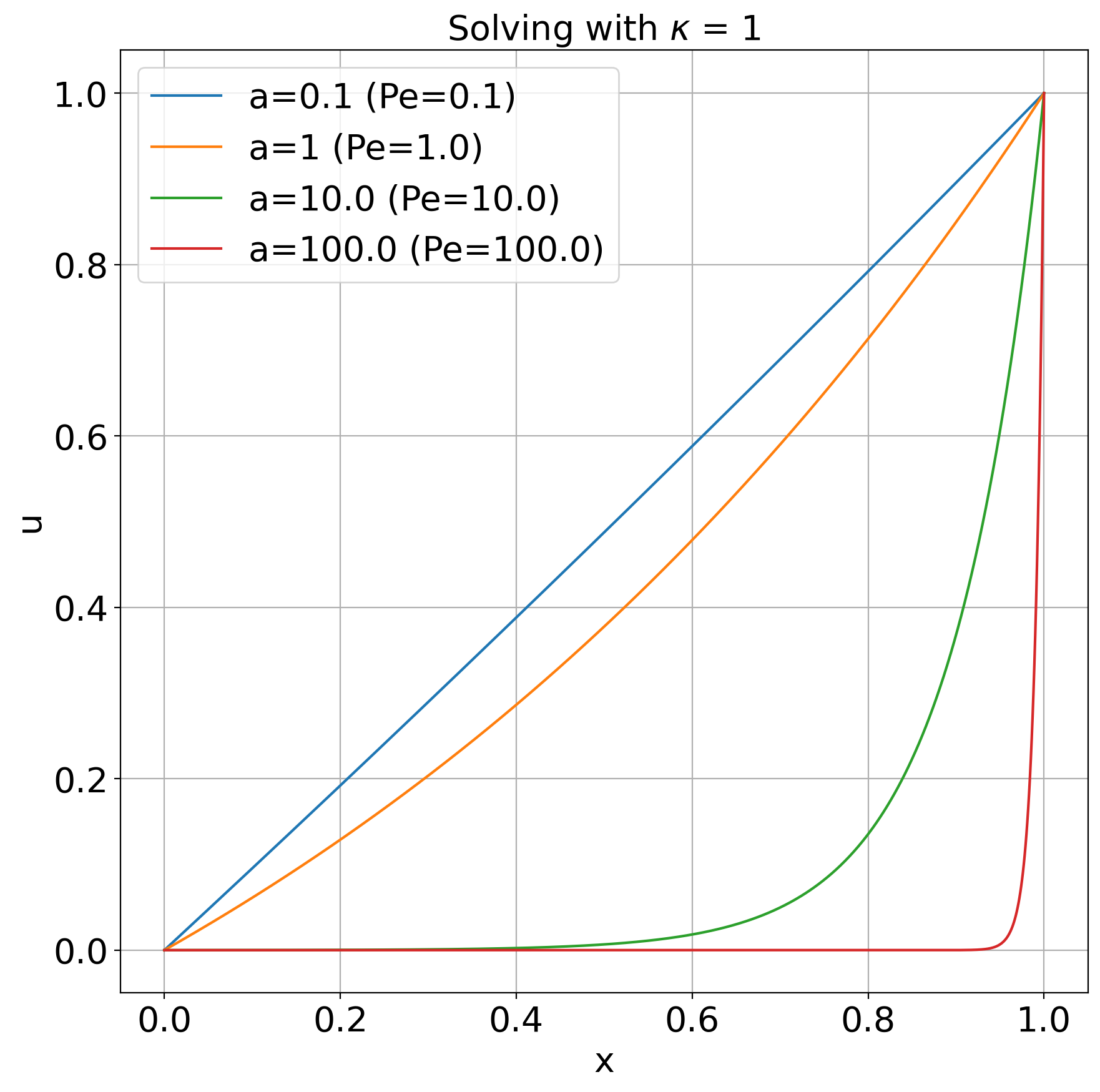}}
\subfigure[Fixed $a$]{\includegraphics[width=0.45\textwidth]{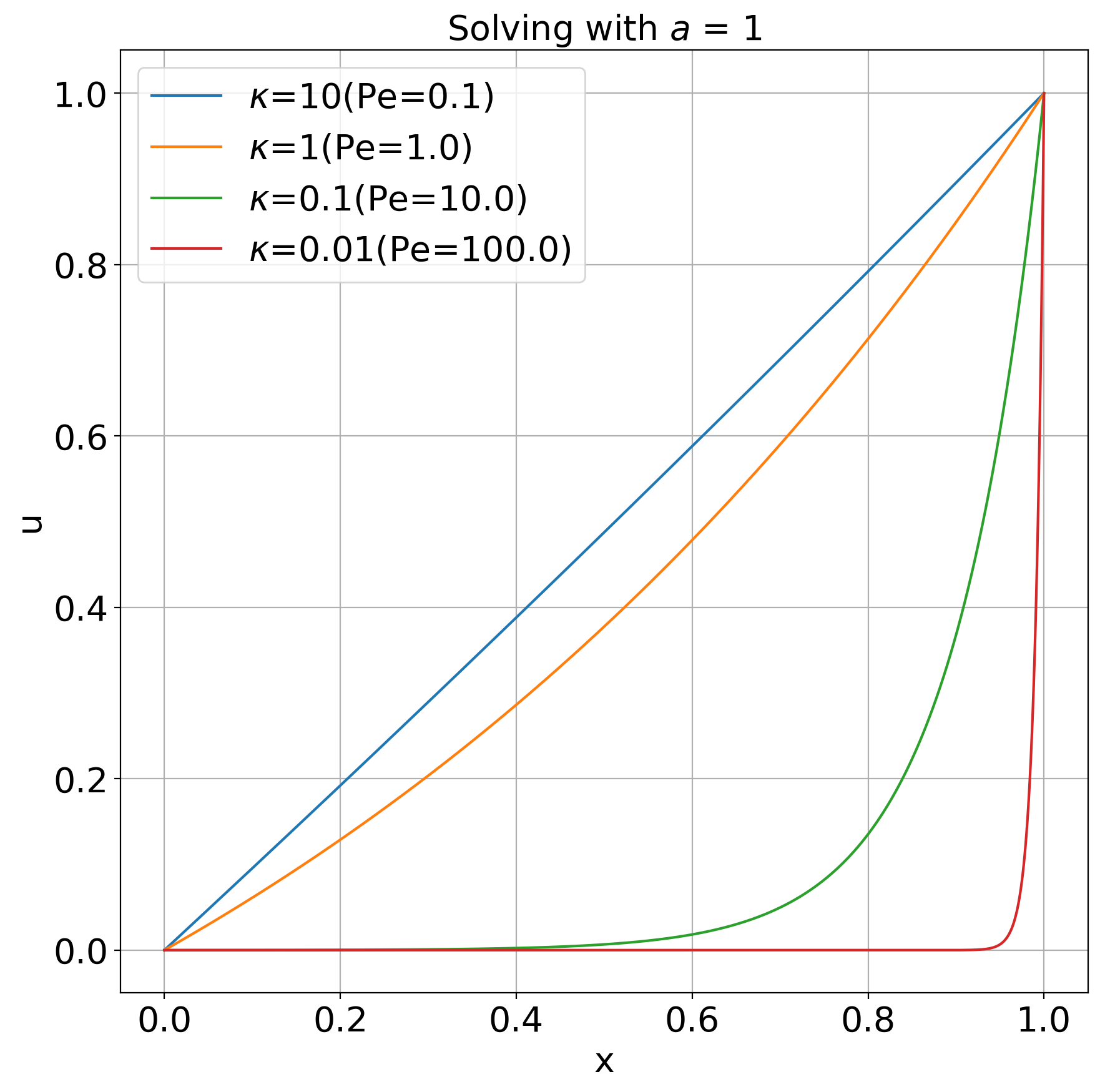}}
\caption{Exact solution of \eqref{eqn:adeq} with $\ell=1$.}
\label{fig:adeq_exact_soln}
\end{center}
\end{figure}

The solution to \eqref{eqn:adeq} for $f\equiv0$ can be analytically written as
\[
u(x) = \frac{1 - \exp(ax/\kappa)}{1 - \exp(a\ell/\kappa)}
\]
where $a\ell/\kappa$ is known as the Peclet number (Pe) and measures the ratio of the strength of advection to the strength of diffusion. We plot the solution for varying values of $a$ and $\kappa$ in Figure \ref{fig:adeq_exact_soln}. Note that for small Pe, the solution is essentially a straight line. But as Pe increases, the solution starts to bend and forming a steeper boundary layer near the right boundary. The thickness of this boundary layer is given by $\delta \approx {\rm Pe} \times l$. 

We will now consider a few methods to numerically solve this toy problem.

\section{Finite difference method}
The key steps of a finite difference scheme are as follows:
\begin{enumerate}
\item Discretize the domain into a grid of points, with the goal being to find the solution at these points. 
\item Approximate the derivates with finite difference approximations at these points. This leads to a system of (linear or non-linear) algebraic equations.
\item Solve this system using a suitable algorithm to find the solution.
\end{enumerate}
Applying these steps to \eqref{eqn:adeq} leads to:
\begin{enumerate}
\item Discretize the domain into $N+1$ points, with $x_i = ih$, $0\leq i\leq N$ where $h=\ell/N$. We wish to solve for $u(x_i) = u_i$.  We also know from the boundary conditions that $u_0 = 0$ and $u_N = 1$.
\item Use the approximations
\begin{align*}
\dd{u}{x}(x_i) &= \frac{u_{i+1} - u_{i-1}}{2h} + \bigO(h^2)\\
\dd{^2u}{x^2}(x_i) &= \frac{u_{i+1} - 2u_i +u_{i-1}}{h^2} + \bigO(h^2)
\end{align*}
Note that both the approximation used above are second order accurate. They are ``central difference'' approximations, as they weigh points on either side of the $i$-th point with the same magnitude. It is worth mentioning that in the limit of large Peclet number, a central difference approximation of the advective term is not ideal since it leads to numerical instability. In such a case, an ``upwind'' approximation is preferred. 
Applying the approximations to the PDE at $x_i$, $1 \leq i \leq N-1$
\begin{align*}
a \frac{u_{i+1} - u_{i-1}}{2h} - \kappa \frac{u_{i+1} - 2u_i +u_{i-1}}{h^2} = f_i\\
\iff u_{i+1} \underbrace{\left( \frac{a}{2h} - \frac{\kappa}{h^2}\right)}_{\gamma} + u_{i} \underbrace{\left( \frac{2\kappa}{h^2}\right)}_{\beta} + u_{i-1} \underbrace{\left( -\frac{a}{2h} - \frac{\kappa}{h^2}\right)}_{\alpha} = f_i
\end{align*}
Looking at each node where the solution is unknown (recall that $u_0 = 0$ and $u_N = 1$ are known),
\begin{equation}\label{eqn:fd}
\begin{aligned}
\beta u_1 + \gamma u_2 &= - \alpha u_0 + f_1\\
\alpha u_{i-1} + \beta u_i + \gamma u_{i+1} &=  f_i, \quad \forall \ 2 \leq i \leq N-2\\
\alpha u_{N-2} + \beta u_{N-1} &= - \gamma u_{N} + f_{N-1}
\end{aligned}
\end{equation}
Combining all the $N-1$ equations in \eqref{eqn:fd}, we get the following linear system
\begin{equation}\label{eqn:fd_sys}
\bm{K} \bm{u} = \bm{f}
\end{equation}
where the tridiagonal matrix $\bm{K}$ and the other vectors in \eqref{eqn:fd_sys} are defined as
\begin{align*}
\bm{K} &= \begin{bmatrix} \beta & \gamma &  & 0  \\ \alpha & \ddots &\ddots & \\ & \ddots & \ddots & \gamma \\ 0 & & \alpha & \beta \end{bmatrix} \in \Ro^{(N-1) \times (N-1)}, \\
 \bm{u} &= \begin{bmatrix} u_1 & u_2 & \cdots & u_{N-2} & u_{N-1} 
\end{bmatrix}^\top \in \Ro^{N-1}, \\
 \bm{f} &= \begin{bmatrix} -\alpha u_0 + f_1 & f_2 & f_3 \cdots & f_{N-2} & f_{N-1} 
& - \gamma u_{N} + f_{N-1} \end{bmatrix}^\top \in \Ro^{N-1}
\end{align*}

\item Solve $\bm{u} = \bm{K}^{-1} \f$. 
\end{enumerate}
Note that:
\begin{itemize}
\item In practice, we never actually invert $\bm{K}$ as it is computationally expensive. We instead use smart numerical algorithms to solve the system \eqref{eqn:fd_sys}. For instance, one can use the Thomas tridiagonal algorithm for this particular system, which is a simplified version of Gaussian elimination. 
\item We only obtain an approximation $u_i \approx u(x_i)$. To reduce the approximation error, we could reduce the mesh size $h$. Alternatively, we could use higher-order finite difference approximations which would lead to a "wider stencil" to approximate the derivates at each point.
\item We can think of each point where we ``apply'' the PDE as a collocation point. This idea of applying the PDE at collocation points is shared by the next method we consider. It is also shared by the method with uses MLPs to solve PDEs. 
\end{itemize}

\section{Spectral collocation method}
Spectral collocation methods seek a solution written as an expansion in terms of a set of smooth and global basis functions. The basis functions are chosen a priori, whereas the coefficients of the expansion are unknowns, and are computed by requiring that the numerical solution of the PDE is exact at a set of so-called collocation points. 
More specifically, this approach involves the following steps. 

\begin{enumerate}
\item Select a set of global basis functions with the following properties:
\begin{enumerate}
    \item It forms complete basis in the space of functions being considered.
    \item Is smooth enough so that derivatives can be evaluated.
    \item Easy to evaluate.
    \item Derivatives that are easy to evaluate.
\end{enumerate}
For instance, one can use the Chebyshev polynomials defined on $\xi \in (-1,1)$, given by the following recurrence relation
\[
T_0(\xi) = 1, \quad T_1(\xi) = \xi, \quad T_{n+1}(\xi) = 2 \xi T_n(\xi) - T_{n-1}(\xi)
\] 
The first few Chebyshev polynomials are shown in Figure \ref{fig:cheb}. Note that this basis satisfies all the required properties listed above. It is easy to evaluate at any point because one can use the recurrence relation above and the values of the two lower-order polynomials to evaluate the Chebyshev polynomial of the subsequent order. One can also take derivatives of the recurrence relation above to evaluate a recurrence relation for derivatives of all orders.

\begin{figure}[htbp!]
\begin{center}
\includegraphics[width=0.4\textwidth]{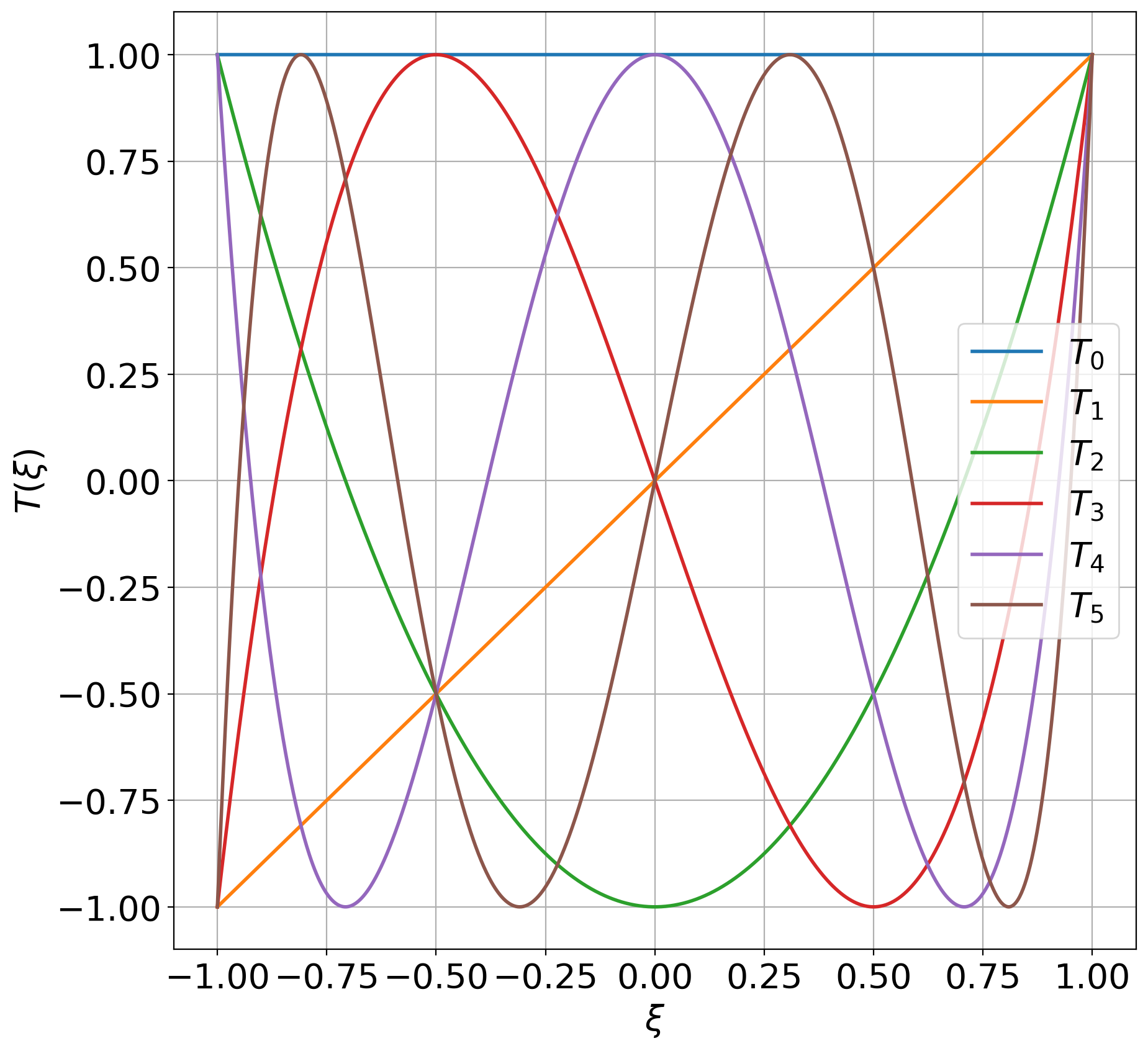}
\caption{First few Chebyshev polynomials.}
\label{fig:cheb}
\end{center}
\end{figure}

\item Write the solution as a linear combination of the basis functions $\{\phi_n(x)\}_{n=0}^N$
\begin{equation}\label{eqn:sc_approx}
u(x)  = \sum_{n=0}^N u_n \phi_n(x)
\end{equation}
where $u_n$ are the basis coefficients. For our toy problem \eqref{eqn:adeq} (assuming $\ell=1$), we will use the Chebyshev polynomials $\phi_n(x) = T_n(2x-1)$, where the argument is transformed to use these functions on the interval $(0,1)$. 

\item Evaluate the derivates for the PDE, which for our toy problem will be
\begin{equation}\label{eqn:scm_diff}
\begin{aligned}
\dd{u}{x}(x) &= \sum_{n=0}^N u_n \phi_n^\prime(x) = \sum_{n=0}^N u_n 2 T^\prime_n(2x-1)\\
\dd{^2u}{x^2}(x) &= \sum_{n=0}^N u_n \phi_n^{\prime \prime} (x) = \sum_{n=0}^N u_n 4 T^{\prime \prime}_n(2x-1)
\end{aligned}
\end{equation}

\item Use the boundary conditions of the PDE. For the specific case of \eqref{eqn:adeq},
\begin{equation}\label{eqn:scm_bc}
\begin{aligned}
u(0) = 0 &\implies \sum_{n=0}^N u_n \phi_n(0) = \sum_{n=0}^N u_n T_n(-1) = 0,\\
u(1) = 1 &\implies \sum_{n=0}^N u_n \phi_n(1) = \sum_{n=0}^N u_n T_n(1) = 1\\
\end{aligned}
\end{equation}
which leads to 2 linear equations for $N+1$ coefficients. We then consider a set of (suitably chosen) nodes $x_i$, $1 \leq i \leq N-1$ in the interior  of the domain, i.e. the collocation points, and use the derivatives found in step 3. in the PDE evaluated at these $N-1$ nodes 
\begin{equation}\label{eqn:scm_pde}
\begin{aligned}
a \sum_{n=0}^N u_n \phi_n^\prime(x_i) - \kappa \sum_{n=0}^N u_n \phi_n^{\prime \prime} (x_i) = f(x_i)\\
\implies  \sum_{n=0}^N u_n \left( 2 a T_n^\prime(2x_i-1) - 4 \kappa T_n^{\prime \prime} (2x_i-1) \right)  = f(x_i)
\end{aligned}
\end{equation}
This leads to an additional $N-1$ equations for the $N+1$ coefficients. Combining \eqref{eqn:scm_bc} and \eqref{eqn:scm_pde} leads to the following linear system
\begin{equation}\label{eqn:sc_sys}
\bm{K} \bm{u} = \bm{f}
\end{equation}
where
\begin{align*}
\bm{K} & \in \Ro^{(N+1) \times (N+1)}, \\
 \bm{u} &= \begin{bmatrix} u_0 & u_2 & \cdots & u_{N-1} & u_{N} 
\end{bmatrix}^\top \in \Ro^{N+1}, \\
 \bm{f} &= \begin{bmatrix} 0 & f(x_1) & f(x_1) & \cdots & f(x_{N-2} )& f(x_{N-1}) & 1
\end{bmatrix}^\top \in \Ro^{N+1}
\end{align*}
\item Solve $\bm{u} = \bm{K}^{-1} \f$. 
\end{enumerate}
We need to choose the collocation/quadrature points $x_i$ properly, so that $\bm{K}$ has desirable properties that make the linear system \eqref{eqn:sc_sys} easier to solve. These include invertibility, positive-definiteness, sparseness, etc.

\begin{remark}
The method is called a collocation method as the PDE is evaluated at the specific collocation/quadrature points $x_i$.
\end{remark}

\begin{remark}
When working with a non-linear PDE, we will end up with a non-linear systems of algebraic equations for the coefficients $u_0,..., u_N$, which is typically solved by Newton's method.
\end{remark}

Let us look at a least-square variant for finding the coefficients of the expansion of the spectral methods. As done earlier, we still represent the solution using \eqref{eqn:sc_approx} and compute its derivates. Then, the coefficients $\bm{u}$ are found by minimizing the following loss function
\begin{equation}\label{eqn:sclsq_loss}
\begin{aligned}
\Pi(\bm{u}) &= \Pi_{\text{int}}(\bm{u}) + \lambda \Pi_{\text{bc}}(\bm{u})\\
\Pi_{\text{bc}}(\bm{u}) &= \left| \sum_{n=0}^N u_n \phi_n(0) - 0\right|^2 + \left| \sum_{n=0}^N u_n \phi_n(1) - 1\right|^2\\
\Pi_{\text{int}}(\bm{u}) &= \frac{1}{N_\text{train}} \sum_{i=1}^{N_\text{train}} \left| a \sum_{n=0}^N u_n \phi^\prime_n(x_i) - \kappa \sum_{n=0}^N u_n \phi^{\prime \prime}_n(x_i)  - f(x_i)\right|^2
\end{aligned}
\end{equation}
This can be solved using any of the gradient-based methods we have seen in Chapter 2. This approach is especially useful when treating non-linear PDEs. In fact, in those cases it is not be possible to write a linear system in terms of the coefficients such as (\ref{eqn:sc_sys}). A few things to note here
\begin{itemize}
\item $\lambda$ is a parameter used to scale the interior loss and boundary loss differently.
\item The number of interior point $x_i$ can be chosen independently of the number of basis functions. In other words, $N_\text{train}$ does not have to be the same as $N$. 
\item We will see in the next section how this variant of the spectral method is very similar to how deep neural networks are used to solve PDEs.
\end{itemize}

\section{Physics-informed neural networks (PINNs)}
The idea of using neural networks to solve partial differential equations was introduced in 1990-2000s by Lagaris et al. \cite{PINNs_Lagaris}. With the renewed interest in using machine learning tools in solving PDEs, this idea was rediscovered in 2019 by Raissi et al. \cite{raissi2019}, and was given the term PINNs (physics-informed neural networks). The basic idea of PINNs is similar to regression, except that the loss function $\Pi(\btheta)$ contains derivate operators arising in the PDE being considered. We outline the main steps below for a one-dimensional scalar PDE, which can easily be extended to multi-dimensional systems of PDEs. We recommend that the reader thinks about the similarities and differences between this method and spectral collocation method described in the previous section. 

\begin{enumerate}
\item Select a neural network as a function representation of the PDE solution:

\begin{equation}\label{eqn:pinns}
u = \mathcal{F}(x; \btheta).
\end{equation}
Some crucial properties required by this representation are:
\begin{enumerate}
   \item Do we have completeness with the representation, i.e., can we accurately approximate the necessary class of function using the representation? The answer is yes, because of the universal approximation theorems of neural networks (see Section \ref{sec:uni_app_thms}).
   \item Is the representation smooth? The answer is yes if the activation function is smooth, such as tanh, sin, etc. Note that we cannot use ReLU since it does not enough number of smooth derivatives. 
   \item Is it easy to evaluate? The answer is yes, due to a quick forward propagation pass. 
   \item Is it easy to evaluate derivates? The answer is yes, due to back-propagation. This will be discussed in detail below. 
\end{enumerate}   

\item Given the representation \eqref{eqn:pinns}, we need to find $\btheta$ such that the PDE is satisfied in some suitable form. Compare this with spectral collocation approximation given by \eqref{eqn:sc_approx}, where we need to determine the coefficients $u_n$. Note that while the dependence on the coefficients $u_n$ in \eqref{eqn:sc_approx} is linear, the dependence on $\btheta$ in \eqref{eqn:pinns} can be highly non-linear.

\item Next we want to find the derivatives of the representation. Consider the computational graph of the network as shown in Figure \ref{fig:dd_graph}. It comprises alternate steps of affine transformations and component-wise nonlinear transformation. The derivative of the output with respect to the input can be evaluated by back-propagation. The graph in Figure \ref{fig:dd_graph} is obtained by simply setting $\Pi = \x^{(L+1)}$ in the graph shown in Figure \ref{fig:comp_graph}. Further, once we recognize that $\df{\x^{(L+1)}}{\x^{(L+1)}} = \bm{1}$, the identity matrix, we can easily read from this graph that 
\begin{eqnarray*}
\df{\x^{(L+1)}}{\x^{(0)}} = \bm{W}^{(L+1)} \bm{S}^{(L+1)} 
\bm{W}^{(L)} \bm{S}^{(L)} \cdots \bm{W}^{(2)} \bm{S}^{(2)} \bm{W}^{(1)} \bm{S}^{(1)}
\end{eqnarray*}
%
Hence, the evaluation of $\dd{u}{x}$ requires the extention of the original graph with a backward branch used to evaluate the derivative of the activation function for each component of the vectors $\bm{\xi}^{(l)}$ (see Figure \ref{fig:dd_graph}). 
The second derivative $\dd{^2u}{x^2}$ is evaluated by performing back-propagation of the extended graph. To evaluate higher order derivatives, the graph will need to be extended further in a similar manner.  This is what happens behind the scenes in Pytorch when a call to "autograd" is made. 

\begin{figure}[htbp!]
\begin{center}
\includegraphics[width=\textwidth]{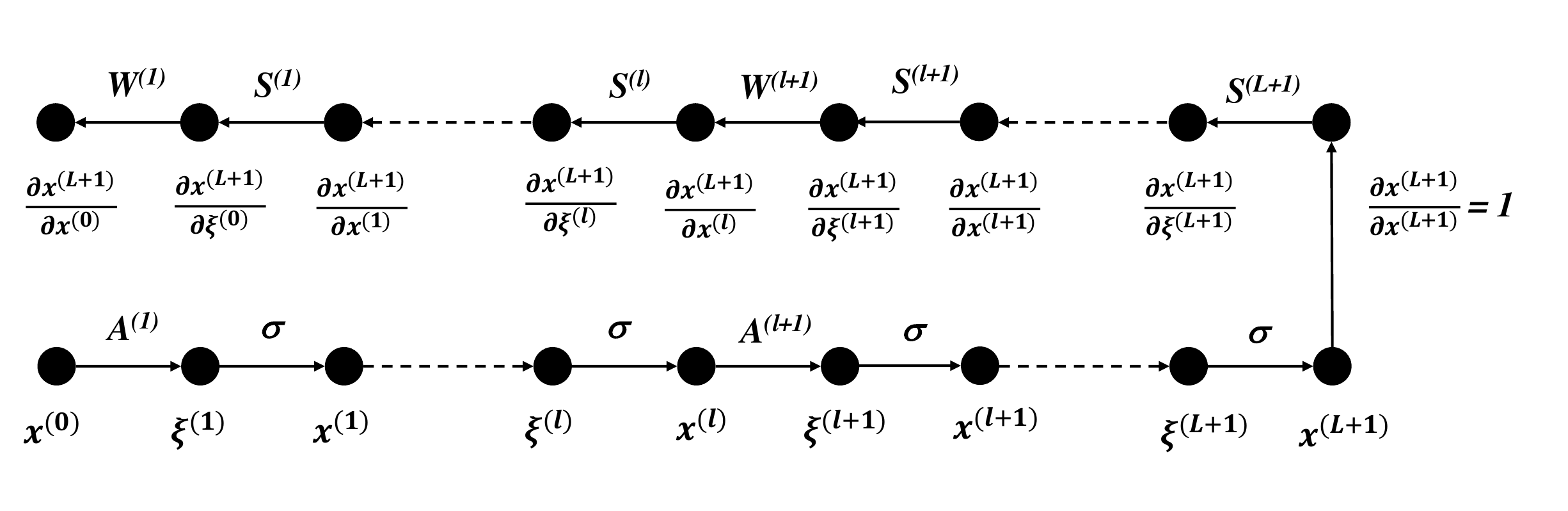}
\caption{Extended graph to evaluate derivatives with respect to network input.}
\label{fig:dd_graph}
\end{center}
\end{figure}

\item Insert the functional representation of the solution \eqref{eqn:pinns} into the PDE to find the parameters $\btheta$. To do this, we first define a set of points $\mathcal{S} = \{x_i: 1 \leq i \leq N_\text{train}\}$ used to train the network, analogous to the set of collocation points in the spectral collocation methods. Thereafter, we need to define the loss function (specialized to our toy problem \eqref{eqn:adeq})
\begin{equation}\label{eqn:pinns_loss}
\begin{aligned}
\Pi(\btheta) &= \Pi_{\text{int}}(\btheta) + \lambda_b \Pi_{\text{b}}(\btheta),\\
\Pi_{\text{b}}(\btheta) &= \left(  \mathcal{F}(0; \btheta) - 0\right)^2 + \left( \mathcal{F}(1; \btheta) - 1\right)^2,\\
\Pi_{\text{int}}(\btheta) &= \frac{1}{N_\text{train}} \sum_{i=1}^{N_\text{train}} \left( a  \mathcal{F}^\prime(x_i; \btheta) - \kappa  \mathcal{F}^{\prime \prime}(x_i; \btheta)_n  - f(x_i)\right)^2.
\end{aligned}
\end{equation}
After training the network, i.e. solving the minimization problem $\btheta^* = \argmin{\btheta} \ \Pi(\btheta)$, the solution writes $u^*(x) = \mathcal{F}(x;\btheta^*)$.
Note that this is exactly what is done for the least squares variant of the spectral collocation method, where the coefficients $u_n$ are solved by minimizing a similar loss.
\end{enumerate}

We make a few remarks:
\begin{itemize}
\item When we are able to find $\btheta^*$ for which $\Pi(\btheta^*) = 0$, this implies $\Pi_\text{int}(\btheta^*) = 0$ and $\Pi_\text{b}(\btheta^*) = 0$. In other words, the PDE residuals are zero at the collocation points. This will lead to a good solution as long as the collocation points cover the domain well.
\item There are various ways to improve the accuracy of PINNs, such as
\begin{itemize}
  \item Increasing the number of collocation points.
  \item Changing the hyper-parameter $\lambda_b$ weighting the boundary loss.
  \item Increasing the size of the network. That is, increasing $N_\theta$.
\end{itemize}

\item The boundary conditions (BCs) of a differential equation carry fundamental physical properties of the phenomena we are trying to describe, and it is paramount that those are satisfied by our numerical solution. 
In the framework of PINNs, BCs are enforced as a soft constrained via the penalization term $\Pi_{\text{b}}(\btheta)$. Hence, the hyper-parameter $\lambda_b$ plays a crucial role in the training of the network, as it balances the interplay between the two loss terms during the minimization process. If the gradients of the different loss terms are not adequately scaled, the convergence to a solution that satisfies both the BCs and the PDE itself can be extremely slow. This is particularly exacerbated for stiff PDEs. To address this issue, different self-adaptive techniques to tune the value of $\lambda_b$ along the training have been proposed \cite{wang2021understanding, mcclenny2020self, bischof2021multi}.
\end{itemize}

\section{Extending PINNs to a more general PDE}
Consider a general PDE: Find the solution $\bm{u}:\Omega\subset\Ro^d \rightarrow \Ro^D$ such that
\begin{equation}\label{eqn:md_pde}
\begin{aligned}
L(\bm{u}(\x)) &= \f(\x), \quad \x \in \Omega \\
B(\bm{u}(\x)) &= \g(\x), \quad \x \in \partial \Omega
\end{aligned}
\end{equation}
where $L$ is the differential operator, $\f$ is the known forcing term, $B$ is the boundary operator, and $\g$ is the non-homogeneous part of boundary condition (also prescribed). 

As an example, we can consider the three-dimensional incompressible Navier-Stokes equation solving for the velocity field $\bm{v} = [v_1,v_2,v_3]$ and pressure $p$ on $\Omega = \Omega_S \times [0,T]$. Here $\Omega_S$ is the three dimensions spatial domain and $[0,T]$ is the time interval of interest. The equation is given by 
\begin{equation}\label{eqn:NS}
\begin{aligned}
\df{\bm{v}}{t} + \bm{v} \cdot \nabla \bm{v} + \nabla p - \mu \Delta \bm{v} &= \f, \quad \forall \ (\s,t) \in \Omega \\
\nabla \cdot \bm{u} &=0, \quad \forall \ (\s,t) \in \Omega\\
\bm{v} &=\bm{0}, \quad \forall \ (\s,t)\in \partial \Omega_S \times [0,T]\\
\bm{v}(\s,0) &= \bm{v}_0(\s), \quad \forall \ \s \in \Omega_S.
\end{aligned}
\end{equation}
The first equation above is the balance of linear moment. The second equation enforces the conservation of mass. The third equation is the no-slip boundary condition which is used when the boundary is rigid and fixed. The fourth equation is the prescription of the initial velocity field. 

To design a PINN for \eqref{eqn:md_pde}, the input to the network should be the independent variables $\x$ and the output should be the solution vector $\bm{u}$. For the specific case of the Navier-Stokes system \eqref{eqn:NS}, the input to the network would be $[s_1,s_2,s_3, t] \in \Ro^{4}$, while the output vector would be $\bm{u} = [v_1,v_2,v_3,p] \in \Ro^4$. The steps would be the following:
\begin{enumerate}
\item Construct the loss functions
\begin{itemize}
   \item Define the interior residual $R(\bm{u}) = L(\bm{u}) - \f$.
   \item Define the boundary residual $R_b(\bm{u}) = B(\bm{u}) - \g$.  
   \item Select suitable $N_v$ collocation points in the interior of the domain and $N_b$ points on the domain boundary to evaluate the residuals. These could be chosen as based on quadrature rules, such as Gaussian, Lobatto, Uniform, Random, etc.
\end{itemize}
Then the loss function is
\begin{equation*}
\begin{aligned}
\Pi(\btheta) &= \Pi_\text{int}(\btheta) + \lambda_b \Pi_\text{b}(\btheta)\\
\Pi_\text{int}(\btheta) &= \frac{1}{N_v} \sum_{i=1}^{N_v} \left | R(\bm{\mathcal{F}}(\x_i;\btheta)\right |^2\\
\Pi_\text{b}(\btheta) &= \frac{1}{N_b} \sum_{i=1}^{N_b} \left | R_b(\bm{\mathcal{F}}(\x_i;\btheta)\right |^2
\end{aligned}
\end{equation*}

\item Train the network: find $\btheta^* = \argmin{\btheta} \Pi(\btheta)$, and set the solution as $\bm{u}^* = \bm{\mathcal{F}}(\x;\btheta^*)$
\end{enumerate}

We make some remarks here:
\begin{itemize}
\item It is implicitly assumed that a weight regularization term is also added to the loss $\Pi(\btheta)$.
\item Is $\bm{u}^*(\x)$ the exact solution to the PDE? The answer is No! 
\begin{itemize}
   \item Firstly, $\Pi(\btheta^*)$ may not be zero. 
   \item Even if $\Pi(\btheta^*)$ is identically zero, it only means that the residuals vanishes at the collocation points. However, that does not guarantee that the residuals will vanish everywhere in the domain. For that $N_v, N_b \rightarrow \infty$.
   \item Also, with a fixed network ($N_\theta$ fixed) we cannot represent all functions. For that, we will need $N_\theta \rightarrow \infty$.
\end{itemize}   

\item In practice, we only compute $\Pi(\btheta^*)$. Is the solution error $\|\bm{e}\| = \|\bm{u}^* - \bm{u} \|$ related to this loss value? And if it is, can we say that this error will be small as long as the loss is small? This is what we try to answer in next section.
\end{itemize}

\section{Error analysis for PINNs}
In order to evaluate the error $\bm{e} = \bm{u}^* - \bm{u}$, we need to know the exact solution $\bm{u}$ which is not available in general. We consider a way of overcoming this issue, by restricting our discussion to linear PDEs i.e., $L$ and $B$ are linear operators.

Note that if $\bm{u}$ is the exact solution, then
\begin{equation}\label{eqn:pinns_err1}
L(\bm{e}) = L(\bm{u}^* - \bm{u}) = L(\bm{u}^*) - L(\bm{u}) = L(\bm{u}^*) - \f = R(\bm{u}^*)
\end{equation}
and
\begin{equation}\label{eqn:pinns_err2}
B(\bm{e}) = B(\bm{u}^* - \bm{u}) = B(\bm{u}^*) - B(\bm{u}) = B(\bm{u}^*) - \g = R_b(\bm{u}^*)
\end{equation}
Thus, \eqref{eqn:pinns_err1} \eqref{eqn:pinns_err2} lead to a PDE for $\bm{e}$ driven by the residuals of the MLP solution,
\begin{equation}\label{eqn:err_pde}
\begin{aligned}
L(\bm{e}) &= R(\bm{u}^*), \quad \text{in } \Omega \\
B(\bm{e}) &= R_b(\bm{u}^*), \quad \text{on } \Omega
\end{aligned}
\end{equation}
If the residuals of $\bm{u}^*$ were zero, then $\bm{e} = 0$. Unfortunately, these residuals are not zero. The most that we can say is that they are small at the collocation points. However, from the theory of stability of well-posed PDEs, we have
\begin{equation}\label{eqn:err_bd1}
\|\bm{e}\|_{L^2(\Omega)} \leq C_1 \left( \| R(\bm{u}^*) \|_{L^2(\Omega)} + \| R_b(\bm{u}^*) \|_{L^2(\partial \Omega)} \right)
\end{equation}
where $C_1$ is a stability constant that depends on the PDE, the domain $\Omega$, etc. This is a condition that hols for all well-posed PDEs. It says that if the terms driving the PDE are small, then the solution to the PDE will also be small. This equation tells us that we can control the error if we can control the residuals for the MLP solution. However, in practise we know and control  $\Pi_\text{int}, \Pi_\text{b}$ and not $\| R(\bm{u}^*) \|^2_{L^2(\Omega)}, \| R_b(\bm{u}^*) \|^2_{L^2(\partial \Omega)}$. The question then becomes, are these quantities related. This is answered in the analysis below,
\begin{equation}\label{eqn:err_bd2}
\begin{aligned}
\| R(\bm{u}^*) \|_{L^2(\Omega)} &= \left | m_\Omega \Pi_\text{int}(\btheta^*)^{1/2} + \| R(\bm{u}^*) \|_{L^2(\Omega)} - m_\Omega \Pi_\text{int}(\btheta^*)^{1/2} \right| \\
& \leq m_\Omega \Pi_\text{int}(\btheta^*)^{1/2}  + \left | \| R(\bm{u}^*) \|_{L^2(\Omega)} - m_\Omega \Pi_\text{int}(\btheta^*)^{1/2} \right| \\
& \leq m_\Omega \Pi_\text{int}(\btheta^*)^{1/2} + C_2 (N_v)^{-\alpha}
\end{aligned}
\end{equation}
where $m_\Omega$ is the measure of the domain $\Omega$, $C_2$ and $\alpha > 0$ will depend on the type of quadrature points chosen. In the equation above the first line is obtained by adding and subtracting the term $m_\Omega \Pi_\text{int}(\btheta^*)^{1/2}$. The second line is obtained by using the triangle inequality. The third line is a statement of error in approximating an integral with a finite sum of values evaluated at quadrature points. 

Similarly for the boundary residual
\begin{equation}\label{eqn:err_bd3}
\begin{aligned}
\| R_b(\bm{u}^*) \|_{L^2(\partial \Omega)} & \leq m_{\partial \Omega} \Pi_\text{b}(\btheta^*)^{1/2} + C_3 (N_b)^{-\beta}
\end{aligned}
\end{equation}
where $m_{\partial \Omega}$ is the measure of $\partial \Omega$, $C_3$ and $\beta > 0$ will depend on the type of boundary quadrature points. 

Using \eqref{eqn:err_bd1}, \eqref{eqn:err_bd2} and \eqref{eqn:err_bd3}, we get
\begin{equation}\label{eqn:err_bd4}
\|\bm{e}\|_{L^2(\Omega)} \leq C_1 \left(  \underbrace{m_\Omega \Pi_\text{int}(\btheta^*)^{1/2} + m_{\partial \Omega} \Pi_\text{b}(\btheta^*)^{1/2}}_{\text{reduced by } N_\theta \uparrow } + \underbrace{C_2 (N_v)^{-\alpha} + C_3 (N_b)^{-\beta}}_{\text{reduced by } N_v, N_b \uparrow} \right)
\end{equation}
This equation tells us that it is possible to control the error in the PINNs solution by reducing the loss functions (by increasing $N_\theta$) and by increasing the number of interior and boundary collocation points.  For further details about this analysis, the reader is referred to \cite{mishraPINNs}.

\section{Data assimilation using PINNs}
The problem of data assimilation is often encountered in the science and engineering. In this problem, we are able to make a few sparse measurements of a quantity, and using these we wish to evaluate it everywhere on a fine grid. We are also given a physical principal (in the form of a PDE) that the variable of interest must adhere to. 

Let us assume that we are given a set of sparse measurements of some quantity $u$ on the domain $\Omega$
\[
u_i = u(\x_i), \quad \x_i \in \Omega, \quad 1 \leq i \leq M
\]
Furthermore, we are given that $u$ satisfies some constraint $R(u)=0$ on $\Omega$. Then, data assimilation corresponds to using this information to find the value of $u$ at any $\x \in \Omega$.

We can solve this problem using PINNs. First, we represent $u$ using a neural network $\mathcal{F}(\x,\btheta)$. Next, we define a loss function
\[
\Pi(\btheta) = \underbrace{\frac{\lambda_I}{M} \sum_{i=1}^M (u_i - \mathcal{F}(\x_i,\btheta))^2}_{\text{data matching}} + \underbrace{\frac{1}{N_v} \sum_{i=M+1}^{M+N_v} \left| R(\mathcal{F}(\x_i,\btheta))\right|^2}_{\text{physical constraint}} + \underbrace{\lambda \|\btheta\|^2}_{\text{smoothness}}
\]
where $\x_i, \ M+1 \leq i \leq M+ N_v$ are some collocation points chosen to evaluate the residual, while  $\lambda_I, \lambda$ are hyper-parameters. Then we train the network by finding $\btheta^* = \argmin{\btheta} \Pi(\btheta)$, and set the PINNs solution as $\bm{u}^* = \bm{\mathcal{F}}(\x;\btheta^*)$.

%% file: CNNs.tex

\chapter{Convolutional Neural Networks}
In the previous chapters, we have seen how to construct neural networks using fully-connected layers. We will now look at a different class of layers, called convolution layers, which are very useful when handling inputs which are images. These tasks include classifying images into categories, performing semantic segmentation on images, and transforming images from one type to another.

\section{Functions and images}
Consider a function $u(\x)$ defined on $\x \in [a,b] \times [c,d] \subset \Ro^2$. Then we can visualize the discretized version of this function as an image $\bm{U} \in \Ro^{N_1 \times N_2}$, where
\begin{equation}\label{eqn:grayscale} 
U[i,j] = u(ih,jh), \quad 1 \leq i \leq N_1, \ 1 \leq j \leq N_2, \ h= \text{ pixel size}.
\end{equation}
Note that the image in \eqref{eqn:grayscale} defines a grayscale image where the value of $u$ at each pixel is just the intensity. If we work with color images, then it would be a three-dimensional tensor, with the third dimension corresponding to the red, blue and green channels. In other words, $\bm{U} \in \Ro^{N_1 \times N_2 \times 3}$.

If we want to use a fully-connected neural network (MLP) which takes as input a colored 2D image of size $100\times100$, then the input dimension after unravelling the entire image as a single vector would be $3 \times 10^4$, which is very large. This would in turn lead to very large connected layers which is not computationally feasible. Secondly, when unravelling the image, we lose all spatial context of the initial image. Finally, one would expect that local operations, such as edge detection, would be the same in any region of the image. Consider the weights for a fully connected layer. These would be represented by the matrix $W_{ij}$, where the $i$ index represent the output of the linear transform and the $j$ index represents the input. If the operation was the same for every output index, we would apply the same operation for every $i$ and therefore not need the matrix. To address all these issues, we can use the convolution operator on functions.

\section{Convolutions of functions}
\begin{figure}[htbp!]
\begin{center}
\subfigure[Kernel $g(x)$]{\includegraphics[width=0.4\textwidth]{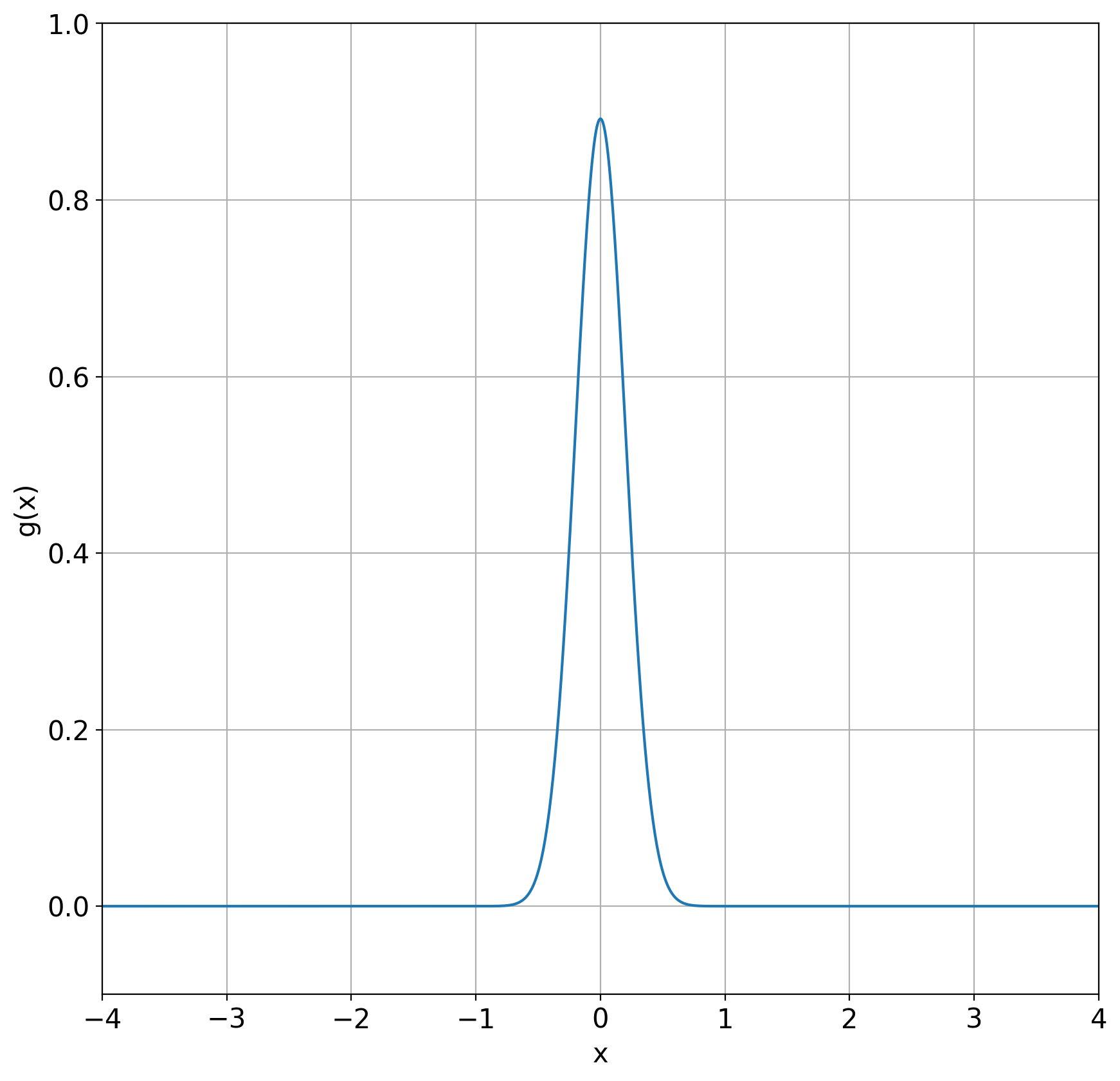}}
\subfigure[Sampling]{\includegraphics[width=0.4\textwidth]{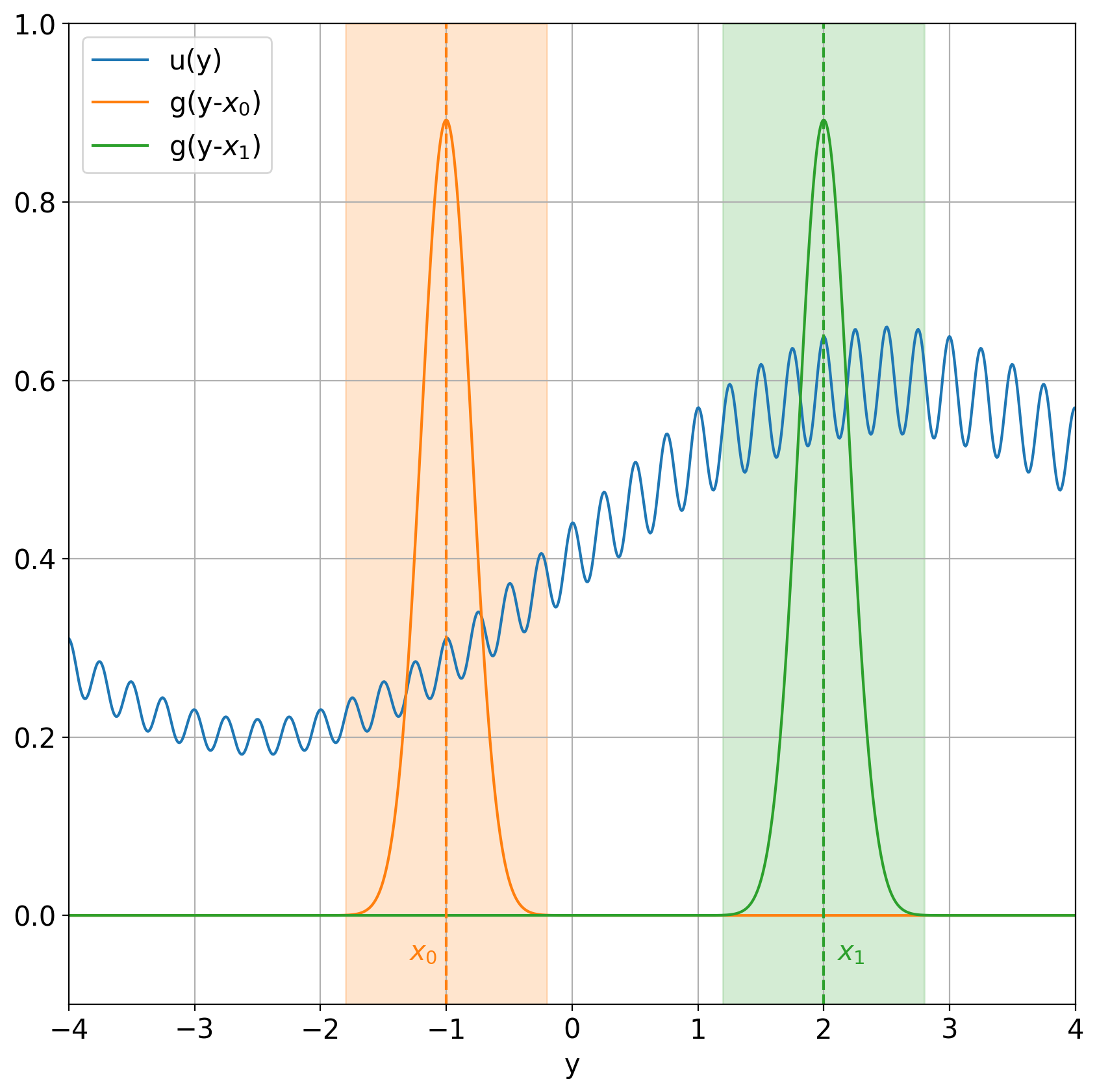}}
\caption{Sampling with shifted kernel in 1D.}
\label{fig:1dkernel}
\end{center}
\end{figure}

The convolution operator maps functions to functions. Consider the function $u(\x)$, $\x \in \Ro^d$, and a sufficiently smooth kernel function $g(\x)$ which decays as $|\x| \rightarrow \infty$.
Then the convolution operator is given by
\begin{equation}\label{eqn:cont_conv}
\overline{u}(\x) = \int_{\Ro^d} g(\y - \x) u(\y) \ud \y
\end{equation}
We can interpret the convolution operator as sampling $u$ by varying $\x$. For example, in 1D, let $u(x)$ and $g(x)$ be as shown in Figure \ref{fig:1dkernel}, and
\[
\overline{u}(x) = \int_{\Ro} g(y - x) u(y) \ud y.
\]
Consider a point $x_0$. Then $g(y-x_0)$ shifts the kernel to the location $x_0$ which will sample the function $u$ in the orange shaded region. Similarly, for another point $x_1$, $g(y-x_1)$ shifts the kernel to the location $x_1$ which will sample the function $u$ in the green shaded region. So as the kernel moves, it samples $u$ in different windows. Note that the same operation is applied regardless of the value of $x$. Lets now consider a few typical kernel functions. 

\subsection{Example 1} 
\begin{figure}[htbp!]
\begin{center}
\includegraphics[width=0.7\textwidth]{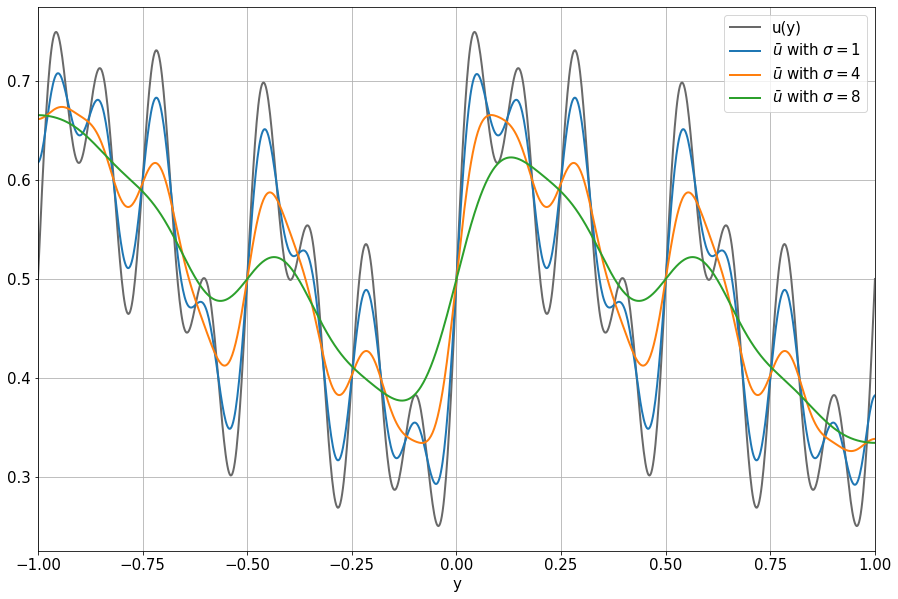}
\caption{1D convolution with Gaussian kernel.}
\label{fig:1dkernel_gauss}
\end{center}
\end{figure}

A popular choice is the Gaussian kernel
\[
g(\bm{\xi}) = \rho(|\bm{\xi}|),
\]
where for any scalar $r$, 
\[
\rho(r) = \frac{\exp(-r^2/2\sigma^2)}{\sqrt{(2 \pi \sigma^2 )^d}  }.
\]
which is used as a smoothing/blurring filter. Here $d$ is the dimension while $\sigma$ denotes the spread. This kernel in 1D is shown in \ref{fig:1dkernel}(a)  for $\sigma=0.2$. Some important properties of this kernel are:
\begin{itemize}
\item It is isotropic, as it only depends on the magnitude of $\bm{\xi}$.
\item The integral over the whole domain is unity.
\item It is parameterized by $\sigma$, which represents a frequency cut-off, as scales finer than $\sigma$ are filtered out by the convolution. This smoothing effect is demonstrated in Figure \ref{fig:1dkernel_gauss}
\end{itemize}

\subsection{Example 2 }
\begin{figure}[htbp!]
\begin{center}
\subfigure[1D]{\includegraphics[width=0.4\textwidth]{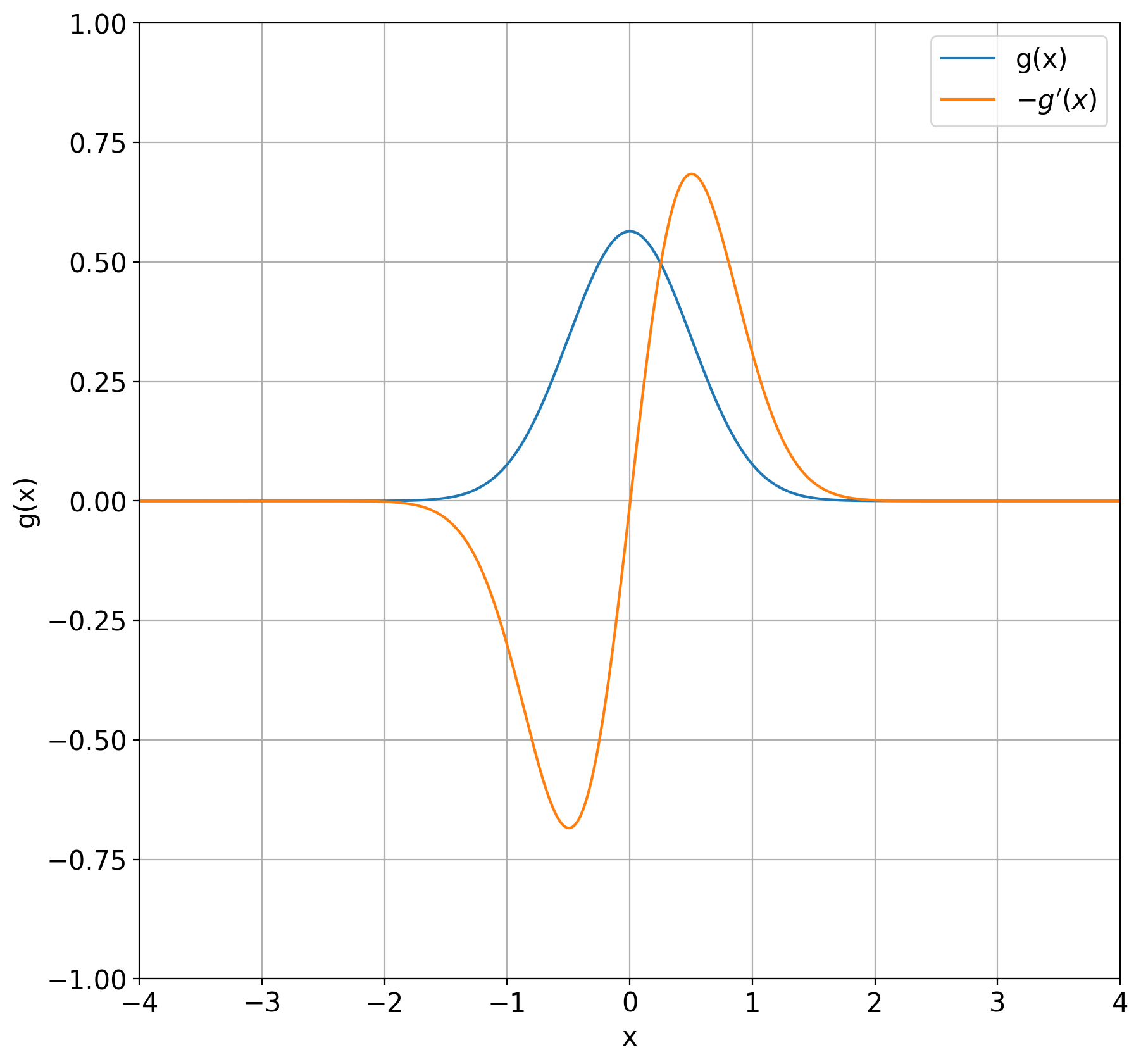}}
\subfigure[2D, with a derivative along the 1-direction]{\includegraphics[width=0.4\textwidth]{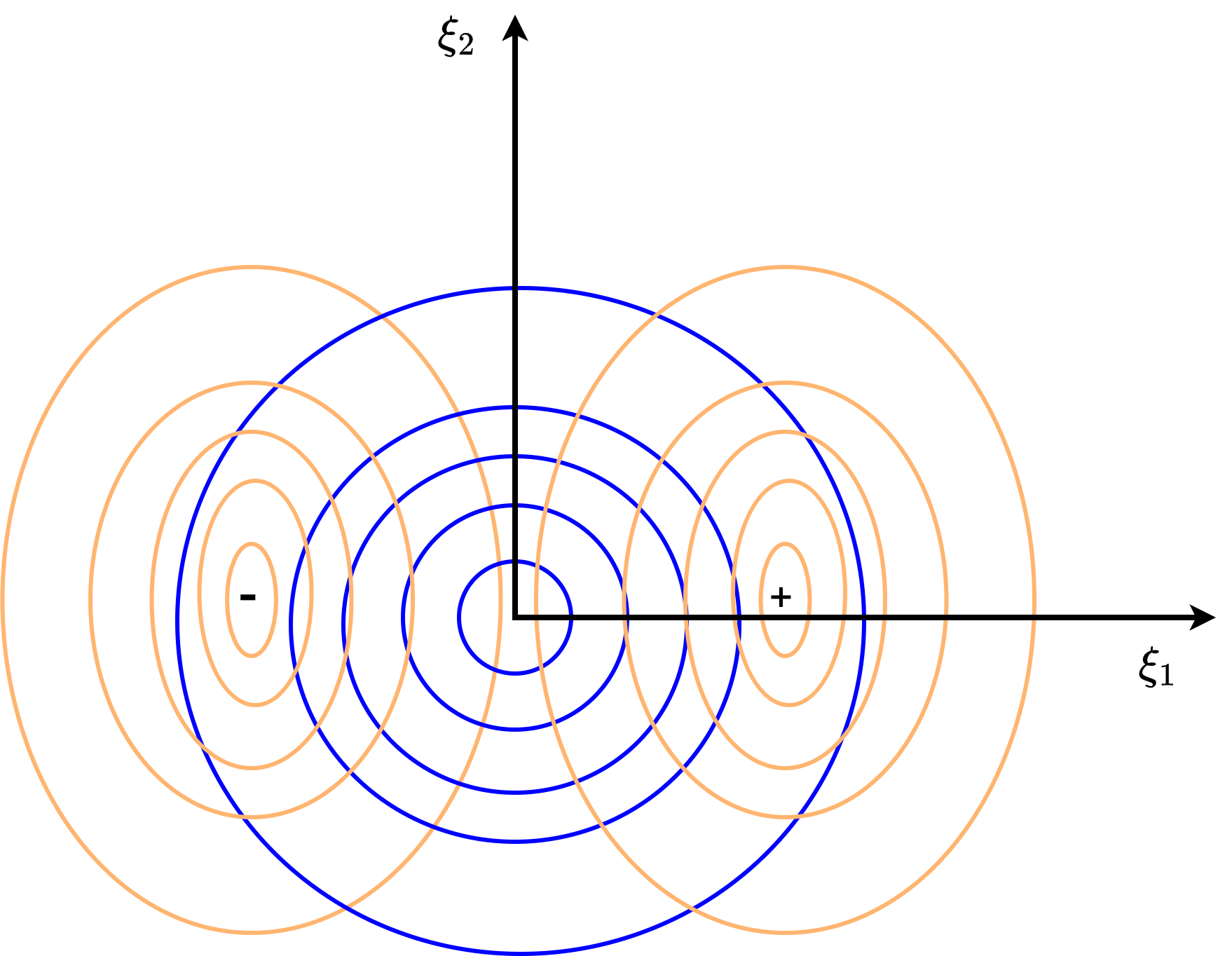}}
\caption{Derivative kernel. The blue curve denotes the Gaussian kernel and the orange curve denotes the derivative. }
\label{fig:kernel_dx}
\end{center}
\end{figure}

Let us consider another example of a kernel that would produce the derivative of a smooth version of $u$. In 2D, we want this to look like
\begin{equation}
\begin{aligned}
\overline{u}(\x) &= \df{}{x_1} \left( \int_{-\infty}^\infty \int_{-\infty}^\infty \rho(|\y-\x|)u(\y) dy_1 dy_2\right) \\
&=\int_{-\infty}^\infty \int_{-\infty}^\infty \df{\rho(|\y-\x|)}{x_1}u(\y) dy_1 dy_2 \\
&=\int_{-\infty}^\infty \int_{-\infty}^\infty \underbrace{\left(-\df{\rho(|\y-\x|)}{y_1}\right)}_{\text{required kernel}}u(\y) dy_1 dy_2 \\
\end{aligned}
\end{equation}
This kernel is shown in both 1D and 2D in Figure \ref{fig:kernel_dx}. Note that the action of this kernel look like a smoothed finite difference operation. That is the region to the left of the center of the kernel is weighted by a negative value and the region to the right is weighted by a positive value. 

\section{Discrete convolutions}
To evaluate the discrete convolution in 2D, consider \eqref{eqn:cont_conv} and discretize it using quadrature. Then, we will have
\begin{equation}
\overline{U}[i,j] = \sum_{m=1}^{N_1} \sum_{n=1}^{N_2} g[m-i,n-j] U[m,n]
\end{equation}
where we have absorbed the measure $h^2$ into the definition of the kernel. As in the continuous case, we will assume that $g$ vanishes after a certain distance
\[
g[p,q] = 0, \quad |p|,|q| > \bar{N} \ \text{(measure of width of the kernel)}.
\]
Thus, the limits of the sum are reduced by exclusing all the pixels over which the convolution will be zero, 
\begin{equation}
\overline{U}[i,j] = \sum_{m=i-\bar{N}}^{i+\bar{N}} \sum_{n=j-\bar{N}}^{j+\bar{N}} g[m-i,n-j] U[m,n].
\end{equation}
Let $m^\prime  = m-i$ and $n^\prime  = n-j$. Then
\begin{equation}
\overline{U}[i,j] = \sum_{m^\prime=-\bar{N}}^{\bar{N}} \sum_{n^\prime=-\bar{N}}^{\bar{N}} g[m^\prime,n^\prime] U[i+m^\prime,j+n^\prime].
\end{equation}
This is precisely how a convolution is applied in deep learning. 
Thus, the convolution is entirely determined by 
\[
g[m,n], \quad |m|,|n| \leq \bar{N},
\]
which become the weights of the convolution layer, with the number of weight being $(\bar{N}+1)^2$.

Let's consider some examples:
\begin{itemize} 
\item A smoothing kernel would be
\[
\frac{1}{8}\begin{bmatrix} \frac{1}{4} & 1 & \frac{1}{4} \\ 1 & 3 & 1\\ \frac{1}{4} & 1 & \frac{1}{4}\end{bmatrix} \approx \text{Gaussian kernel with some $\sigma$}
\]

\item Kernels that lead to the derivative along the $x$-direction and $y$-direction are given by
\[
\begin{bmatrix} 0 & 0 & 0 \\ -1 & 0 & 1 \\ 0 & 0 & 0\end{bmatrix}  \quad \text{and} \quad \begin{bmatrix} 0 & 1 & 0 \\ 0 & 0 & 0\\ 0 & -1 & 0\end{bmatrix}
\]
\item Similarly, the second derivatives anlong the $x$ and $y$-directions are given by kernels of the form
\[
\begin{bmatrix} 0 & 0 & 0 \\ -1 & 2 & -1 \\ 0 & 0 & 0\end{bmatrix}  \quad \text{and} \quad \begin{bmatrix} 0 & -1 & 0 \\ 0 & 2 & 0\\ 0 & -1 & 0\end{bmatrix}
\]
\item While the Laplacian is given by the kernel of the type
\[
\begin{bmatrix} 0 & -1 & 0 \\ -1 & 4 & -1 \\ 0 & -1 & 0\end{bmatrix}  
\]
\end{itemize}

\begin{remark}
We can have different $\bar{N}$ in different directions. That is, we can have kernels with different widths along each direction. 
\end{remark}

\section{Connection to finite differences}
There is a very strong connection between the concept of convolution and the stencil of a finite difference scheme. This is made clear in the discussion below.

Say some function $u(x_1,x_2)$ is represented on a finite grid, where the grid points are indexed by $(i,j)$ with a grid size $h$. Then we use the notation $U[i,j] = u(x_1^i,x_2^j)$. Using Taylor series expansion about $(i,j)$
\[
\frac{U[i+1,j] - U[i-1,j]}{2h}= \frac{u(x_1^{i+1},x_2^j) - u(x_1^{i-1},x_2^j)}{2h}= \df{}{x_1}u(x_1^i,x_2^j) + \mathcal{O}(h^2).
\]
The operation above is {\em identical to } the operation of a discrete convolution with weights given by 
\[
\begin{bmatrix} 0 & 0 & 0 \\ -1 & 0 & 1 \\ 0 & 0 & 0\end{bmatrix} \frac{1}{2h} \approx \df{u}{x_1}.
\]
Thus we may say that this convolution operation approximates a derivative along the 1-direction. 


Similarly, we can show that 
\[
\frac{U[i+1,j] - 2U[i,j] + U[i-1,j]}{h^2}= \df{}{x_1^2}u(x_1^i,x_2^j) + \mathcal{O}(h^2).
\]
and thus the convolution with the kernel given by 
\[
\begin{bmatrix} 0 & 0 & 0 \\ 1 & -2 & 1 \\ 0 & 0 & 0\end{bmatrix} \frac{1}{h^2} \approx \df{u}{x^2_1},
\]
approximates the computation of the second derivative along the 1-direction. 

\section{Convolution layers}
The key things to remember are:
\begin{itemize}
\item Each convolution layer consists of several discrete convolutions, each with its own  kernel.
\item The weights of the kernel, which determine its action (smoothing, first derivative, second derivative etc.), are learnable parameters and are determined when training the network. Thus the way to think about the learning process is that the network learns the operations (convolutions) that are appropriate for its task. The task can be a classification problem, for example. 
\end{itemize}

\begin{figure}[htbp!]
\begin{center}
\subfigure[Action of a convolution layer]{\includegraphics[width=0.45\textwidth]{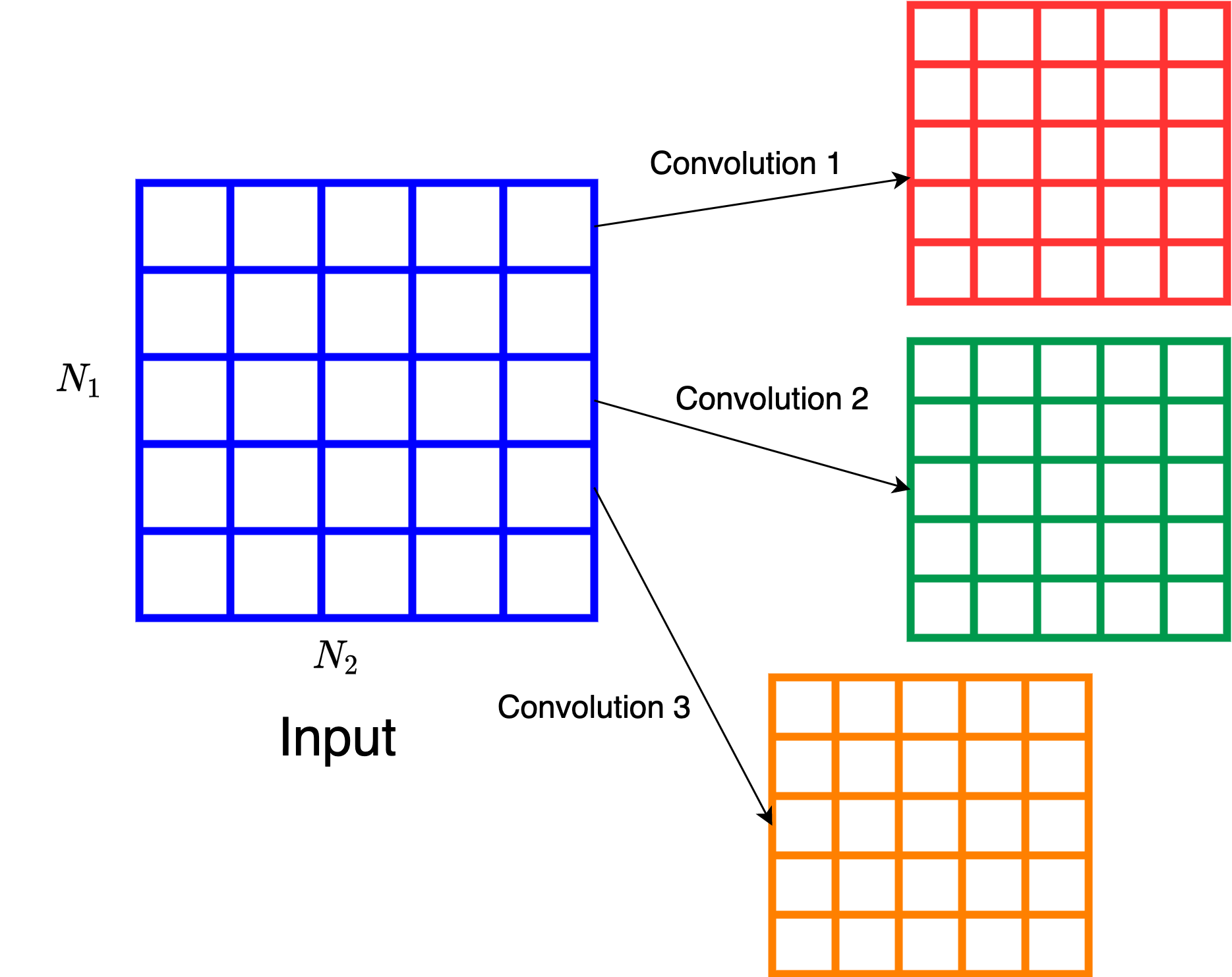}}
\hfill
\subfigure[Convolution without padding]{\includegraphics[width=0.45\textwidth]{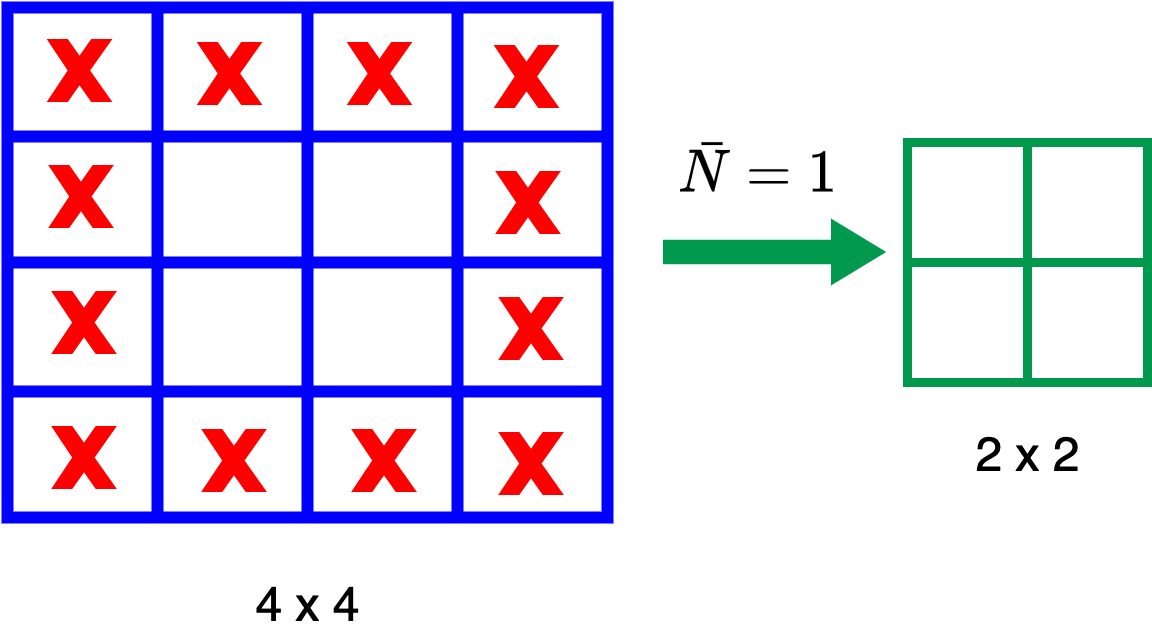}}
\subfigure[Convolution with zero-padding]{\includegraphics[width=0.45\textwidth]{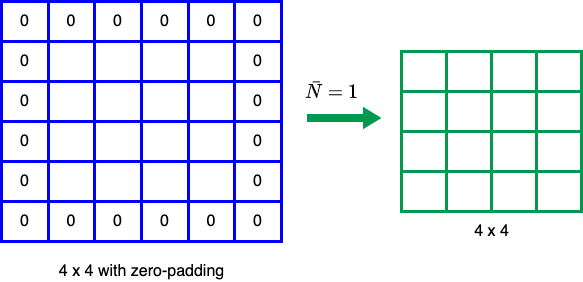}}
\hfill
\subfigure[Convolution with zero-padding and stride 2]{\includegraphics[width=0.45\textwidth]{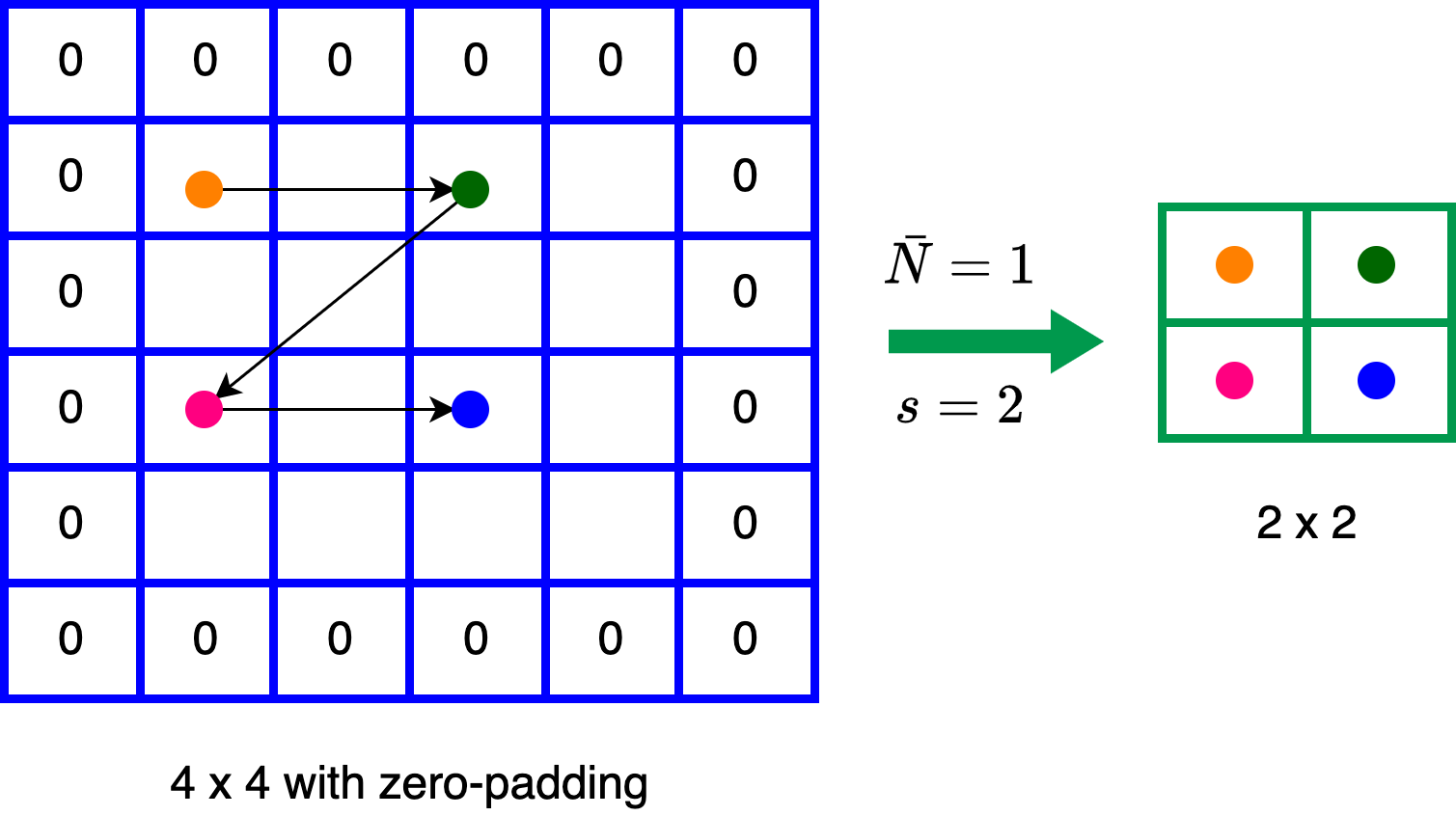}}
\caption{Action of a convolution layer/kernel.}
\label{fig:conv_layer}
\end{center}
\end{figure}

Let us assume we have an $N_1 \times N_2$ image as an input. Then, we will have multiple convolutions in a convolution layer, each of which will generate a different image, as shown in Figure \ref{fig:conv_layer}(a). The trainable parameters of this layer are the weights of each convolution kernel. Assuming the width of the kernels is $\bar{N}$ in each direction, and there are $P$ kernels, then the number of trainable weights will be $P \times (2\bar{N} + 1)^2$.

Next let us consider the size of the output image after applying a single kernel operation. Note that we will not be able to apply the kernel on the boundary pixels since there are no pixel-values available beyond the image boundary (see Figure \ref{fig:conv_layer}(b)). Thus, we will have to skip $\bar{N}$ pixels at each boundary when applying the kernel, leading to an output image of shape $(N_1 - \bar{N} + 1) \times (N_2 - \bar{N} + 1)$. 

One way to overcome this is by \textit{padding} the image with $\bar{N}$ pixels with zero value on each edge. Now we can apply the kernel on the boundary pixels and the output image will be the same size as the input image, as can be seen in Figure \ref{fig:conv_layer}(c).

Another feature of convolutions is known as the \textit{stride} which determines the number of pixels by which the kernel is shifted as we move over the image. In the examples above, the stride was 1 in both directions. In practice, we can choose a stride $>1$ which will further shrink the size of the output image. For instance, if stride was taken as $S$ in each direction (with zero-padding applied), then the output image size would reduce by a factor of $S$ in each direction  (see Figure \ref{fig:conv_layer}(d)).

\subsection{Average and Max Pooling}
\begin{figure}[htbp!]
\begin{center}
\subfigure[Stride 1]{\includegraphics[width=0.4\textwidth]{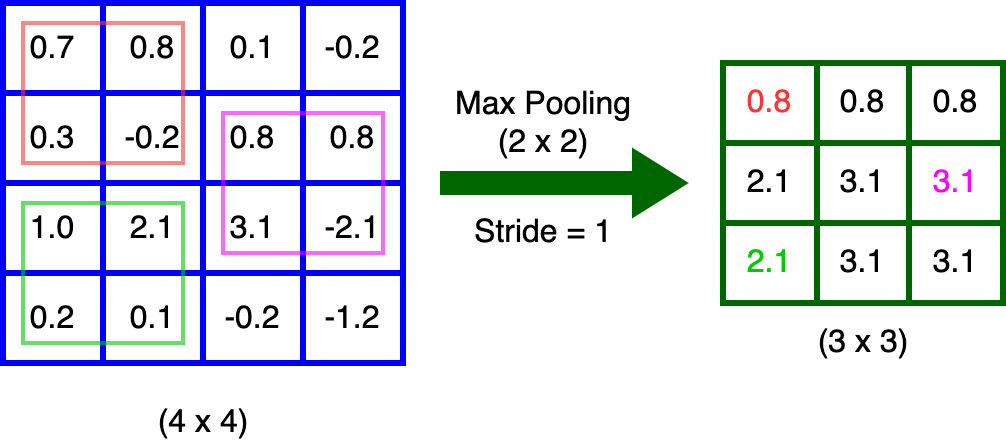}}
\subfigure[Stride 2]{\includegraphics[width=0.4\textwidth]{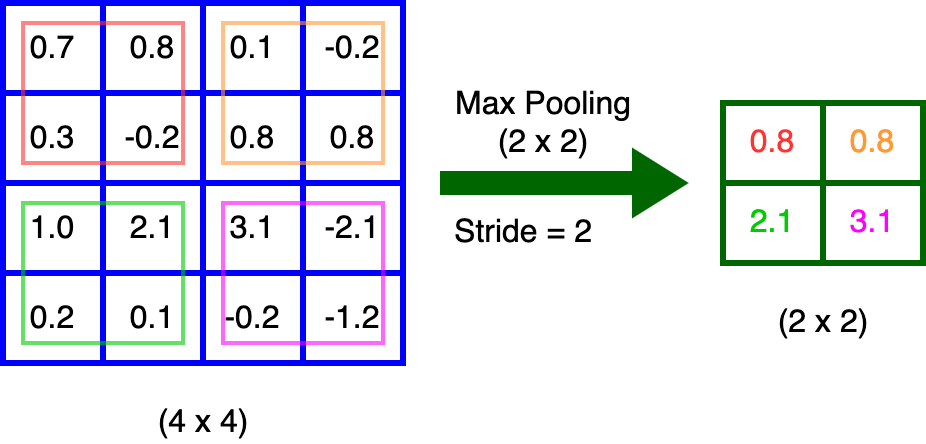}}
\caption{Max pooling applied to an image over patches of size $(2 \times 2)$.}
\label{fig:pooling}
\end{center}
\end{figure}

Pooling operations are generally used to reduce the size of an image, and allowing you to step through different scales of the image. If applied on an image of size $N \times N$ over patches of size $S \times S$, the new image will have dimensions $\frac{N}{S} \times \frac{N}{S}$, where $S$ is the stride of the pooling operation. 
This is shown in Figure \ref{fig:pooling} for $S=2$. 
Note it is typical to select the patch of pixels over which the max or average is computed to be $(S \times S)$, where $S$ is the stride. This is true for Figure \ref{fig:pooling} (b) but not for \ref{fig:pooling} (a).

\begin{figure}[htbp!]
\begin{center}
\subfigure[Original]{\includegraphics[width=0.3\textwidth]{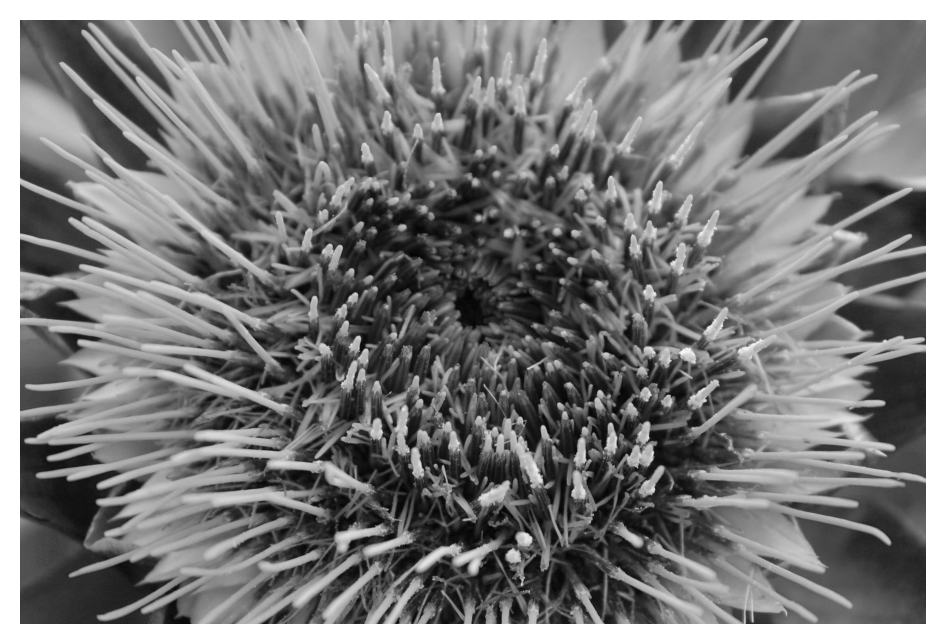}}
\subfigure[After 1 pooling op.]{\includegraphics[width=0.3\textwidth]{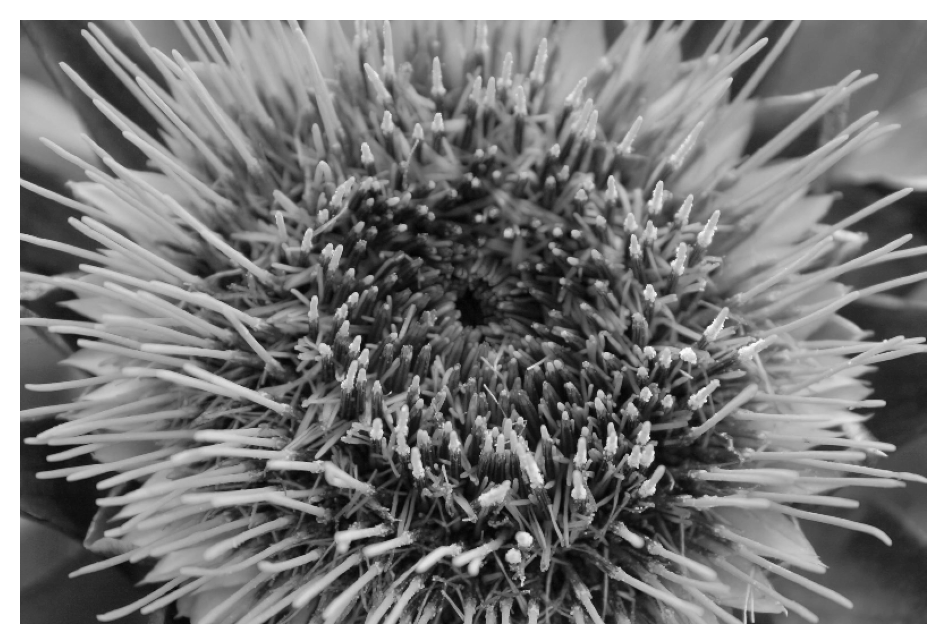}}
\subfigure[After 2 pooling op.]{\includegraphics[width=0.3\textwidth]{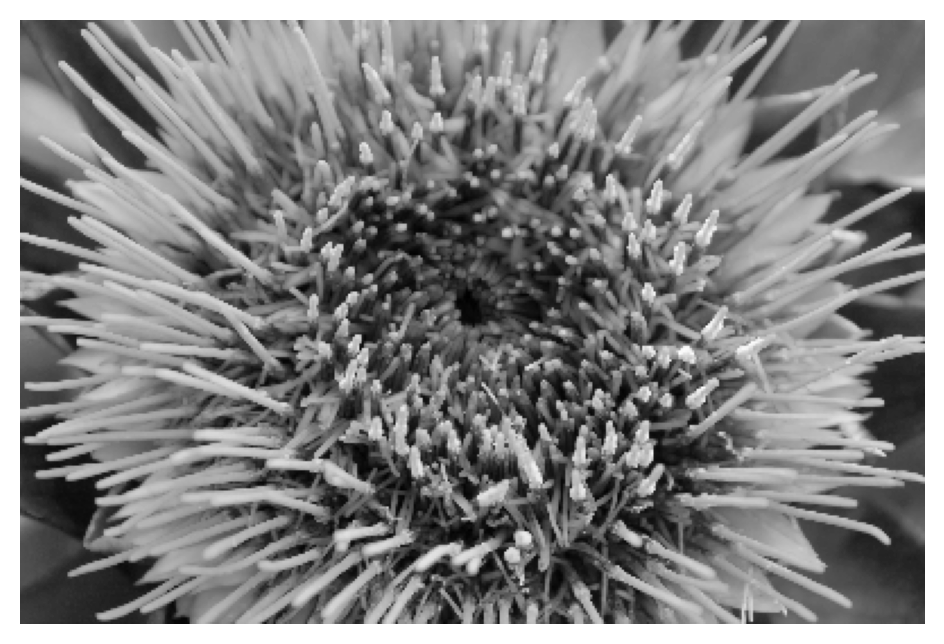}}
\subfigure[After 3 pooling op.]{\includegraphics[width=0.3\textwidth]{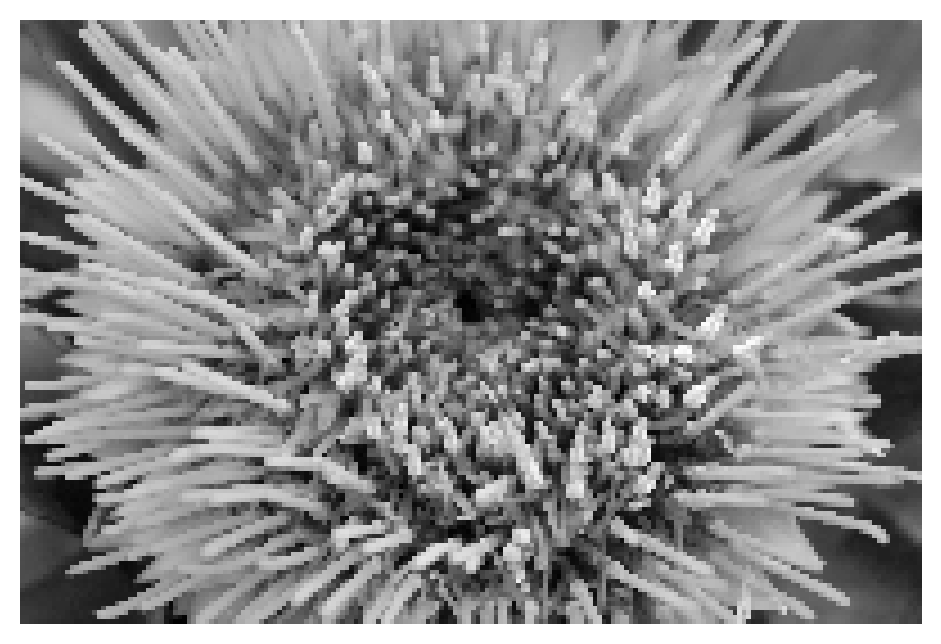}}
\subfigure[After 4 pooling op.]{\includegraphics[width=0.3\textwidth]{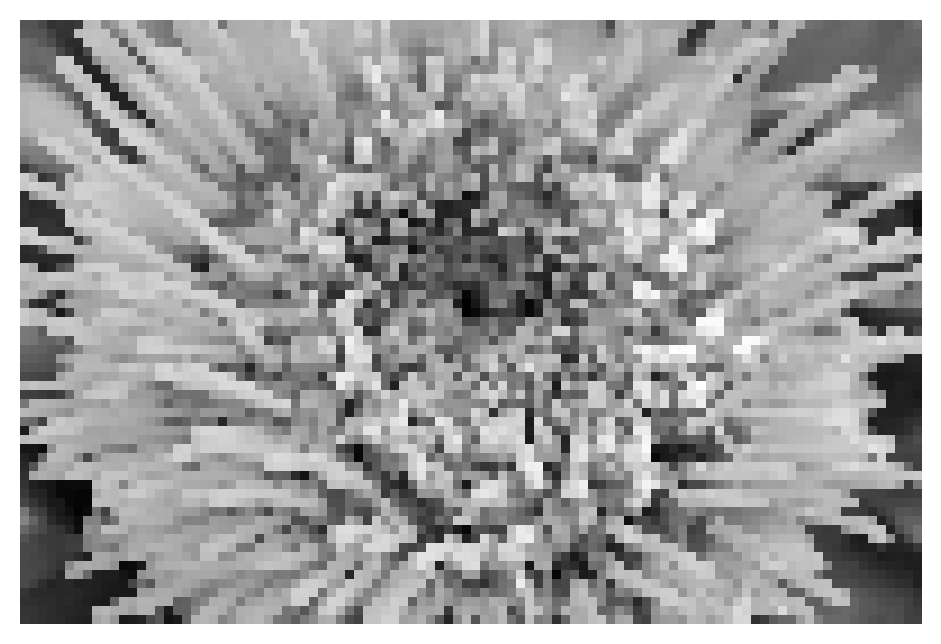}}
\subfigure[After 5 pooling op.]{\includegraphics[width=0.3\textwidth]{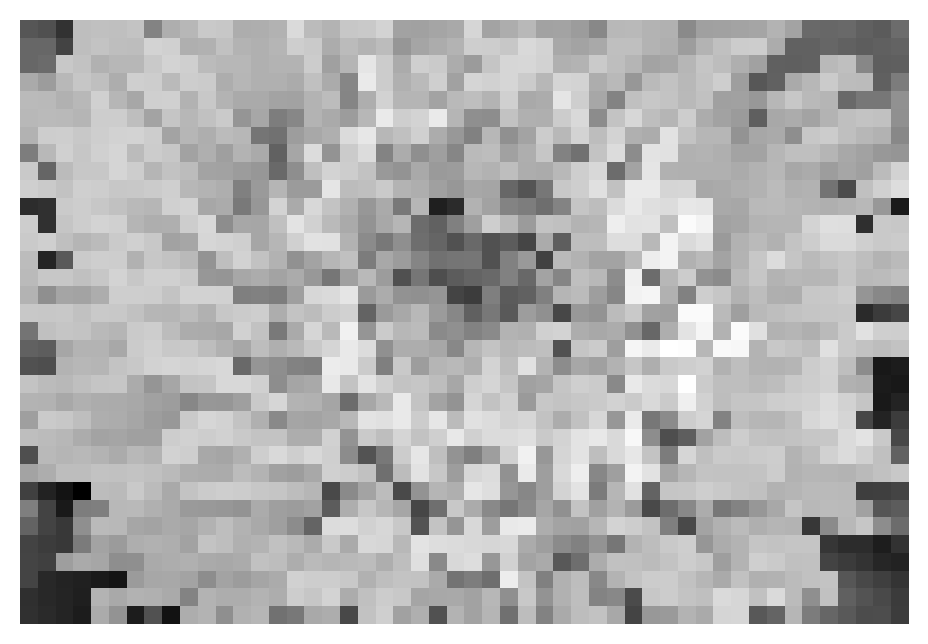}}
\caption{Max pooling applied repeatedly to an image over patches of size $(2 \times 2)$ with stride 2.}
\label{fig:pooling_scales}
\end{center}
\end{figure}

Also, we show in Figure \ref{fig:pooling_scales} how pooling allows us to move through various scales of the image, where the image gets coarser as more pooling operations are applied. Note that pooling operations do not have any trainable parameters. The pooling operation has strong analog in similar operators that are used when designing multigrid preconditioners for solving linear systems of algebraic equations.

\subsection{Convolution for inputs with multiple channels}
Assume that the input to a convolution layer is of size $N_1 \times N_2 \times C$, where $C$ is the number of channels in the input image. Then a convolution layer apply $P$ convolutions on this input and give an output of size $M_1 \times M_2 \times P$. Note that both the spatial resolution as well as the number of channels of the output image might be different from the input image. Furthermore, if a single convolution in the layer uses a kernel of width $k = 2 \bar{N} +1$, then the kernel will be of the shape $k \times k \times C$, i.e., the kernel will have $k \times k$ weights for each of the $C$ input channels of the input image. Each convolution will act on the input to give an output image of shape $M_1 \times M_2 \times 1$. The output of each convolution are stacked together to give the final output of the convolution layer. This can be written as, 
\[
\bar{U}[i,j,k] = \sum_{m=-\bar{N}}^{\bar{N}} \sum_{n=-\bar{N}}^{\bar{N}} \sum_{c=1}^{C} g_k[m,n,c] U[i+m,j+n,c], \quad  1 \leq i \leq M_1, \ 1 \leq j \leq M_2, \ 1 \leq k \leq P,
\]
where $g_k$ is the kernel of the $k$-th convolution in the layer. Note that the total number of trainable parameters will be $(2 \bar{N} + 1) \times (2 \bar{N} + 1) \times P \times C$. This is the type of convolutional layer most frequently encountered in a convolutional neural network, which is described in the following section. 

\section{Convolution Neural Network (CNN)}
Now let's put all the elements together to form the full network. Consider an image classification problem. Then the CNN will be given by $\y = \bm{\mathcal{F}}(\x; \btheta)$ where $\x \in \Ro^{N_1 \times N_2 \times N_3}$ is the input image with $N_3$ channels, while $\y \in \Ro^C$ is the probability vector whose $i$-th component of denotes the probability that the input image belongs the $i$-th class among a total of $C$ classes. The $\y$ are typically one-hot encoded. 

\begin{figure}[htbp!]
\begin{center}
\includegraphics[width=0.9\textwidth]{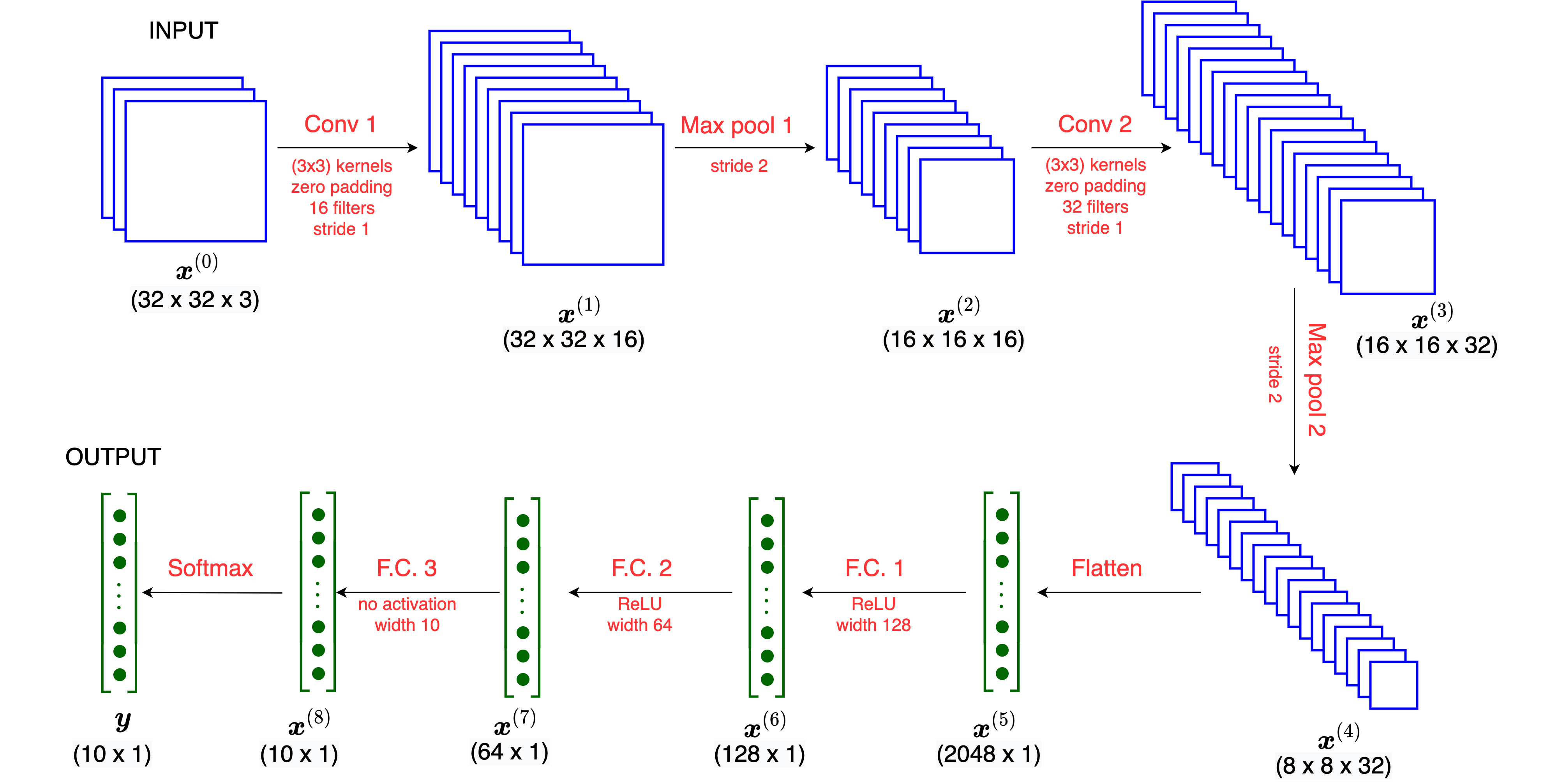}
\caption{Example of a CNN architecture for an image classification problem, for 10 classes.}
\label{fig:cnn}
\end{center}
\end{figure}

The possible architecture of this network is shown in Figure \ref{fig:cnn}. This consists of a number of convolution layers followed by pooling layers, which will reduce the spatial resolution of the input image while increasing the number of channels. The output of the final pooling layer is flattened to form a vector, which is then passed though a number of fully connected layers with some activation function (say ReLU). The final fully connected layer reduced the size of the vector to $C$ (which is taken to be 10 in the Figure), which is then passed through a softmax function to generate the predicted probability vector $\y$. Since we are solving a classification problem, the loss function is taken to be the cross-entropy function
\[
\Pi(\btheta) = - \sum_{i=1}^{N_\text{train}} \sum_{c=1}^C \left [y_c^{(i)} \log\left(\mathcal{F}_c(\x^{(i)}; \btheta) \right) \right].
\]
We train the CNN by trying to find $\btheta^* = \argmin{\btheta} \Pi(\btheta)$ with the final network being $\y = \bm{\mathcal{F}}(\x; \btheta^*)$.

We make some important remarks:
\begin{enumerate}
\item The  convolution operation is also a linear operation on the input, as is the case for a fully connected layer. The only difference is that in a fully-connected layer, the weights connect every pixel in the output to every pixel of the input, while in a convolution layer the weights connect one pixel of the output only to a patch of pixels in the input. Furthermore, the same weights are applied on each patch of the input.
\item In the CNN shown in Figure \ref{fig:cnn}, the convolution layers can be interpreted as encoding information about the input image, while the fully connected layers can be interpreted as using this information to solve the classification problem. This is why in the literature, convolution layers are said to perform \textit{feature selection}. Further the part of the network leading up to the ``flattened'' vector is sometimes referred to as the {\em encoder}. 

\item Sometime, activation is also applied to the output of the convolution layer along with a bias
\[
x^{(l+1)}[i,j,k] = \sigma(\sum_{m,n,c} g_k[m,n,c] x^{(l)}[i+m,j+n,c] + b_k)
\]
where a single bias $b_k$ is used for a given output channel $k$.
\item In the example above we considered an image classification problem. That is, the network was a transform from an image to a class label. We can think of other similar cases. For example, when the network is a transform from an image to a real number. This might have several useful applications in computational physics. Consider the case where you want to create an enstrophy calculator. That is a network that will take as input images of the velocity components of a fluid defined on a grid, and produce as output the integral of the square of the vorticity (called the enstrophy) over the entire domain. Another example would be a network that takes as input the components of the displacement of an elastic solid and produces as output the total strain energy stored within the solid. 

\item The architecture we have considered allows us to transform images to vectors, which is useful in problems involving image classification, image analysis, etc. However, there is another architecture that does the opposite, i.e., maps vectors into images. This is useful in applications involving image synthesis. Finally, by putting these two architecture together, we can transform an image to a vector and back to another image. Such image-to-image transformations are useful in applications such as image semantic segmentation. These ideas are described in the following Sections. 

\item It is worth taking a moment to analyze how  convolution layers act on images and why they are so useful. When dealing with images input in the context of 
deep learning, a first naive approach could be to flatten the image and feed it to a regular fully connected MLP. However, this would lead to different problems.
In fact, for regular images, the size of the flatten input would be extremely large. In that case, we would have 2 possibilities when defining the architecture of the network:
\begin{enumerate}
    \item We can size the first layers of the network to have a width comparable to the (large) dimension of the input. 
    \item We can have a sharp decrease in the width of the second hidden layer.
\end{enumerate}
Either of the strategies would lead to issues. In the first case, the size of the network would be too large. Hence, there would be too many trainable parameters which would require an unrealistic amount of data to train the network. In the second case, the compression of information happening between the  
first two layers would be too aggressive, making it very hard for the network to learn how to extract the right features across different images. Moreover, important spatial relationship among pixels (like edges, shapes, etc.) are lost by flattening an image. Ideally we would like to leverage these relations as much as possible, since they carry important spatial information.
Convolution layers can solve both issues. They allow the input to be a 2D image, while drastically decreasing the number of learnable parameters needed for the feature extraction task. In fact, kernels introduce a limited number of parameters compared to a classic fully connected layer. Since the same kernel is applied at different pixel locations in an image, .i.e. \textit{parameter sharing}, they utilize the computational resources in an efficient and smart manner.

\end{enumerate}

\section{Transpose convolution layers}
We have seen how convolution and pooling layers can be used to scale down images. We now consider layers that do the opposite, i.e, scale up images. To understand what this operation would look like, let us look at a few examples

\begin{figure}[htbp!]
\begin{center}
\subfigure[kernel size $(3 \times 3)$, checker-boarding]{\includegraphics[width=0.9\textwidth]{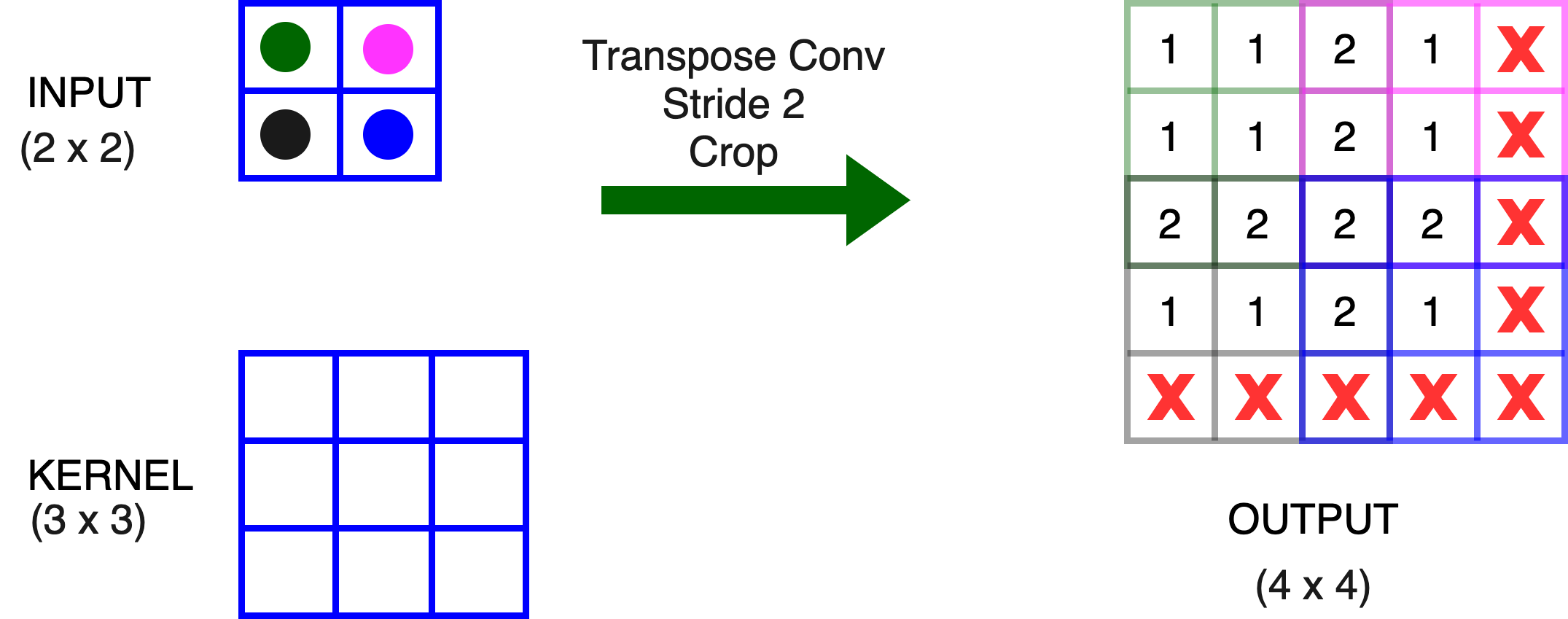}}
\subfigure[kernel size $(2 \times 2)$, no checker-boarding]{\includegraphics[width=0.9\textwidth]{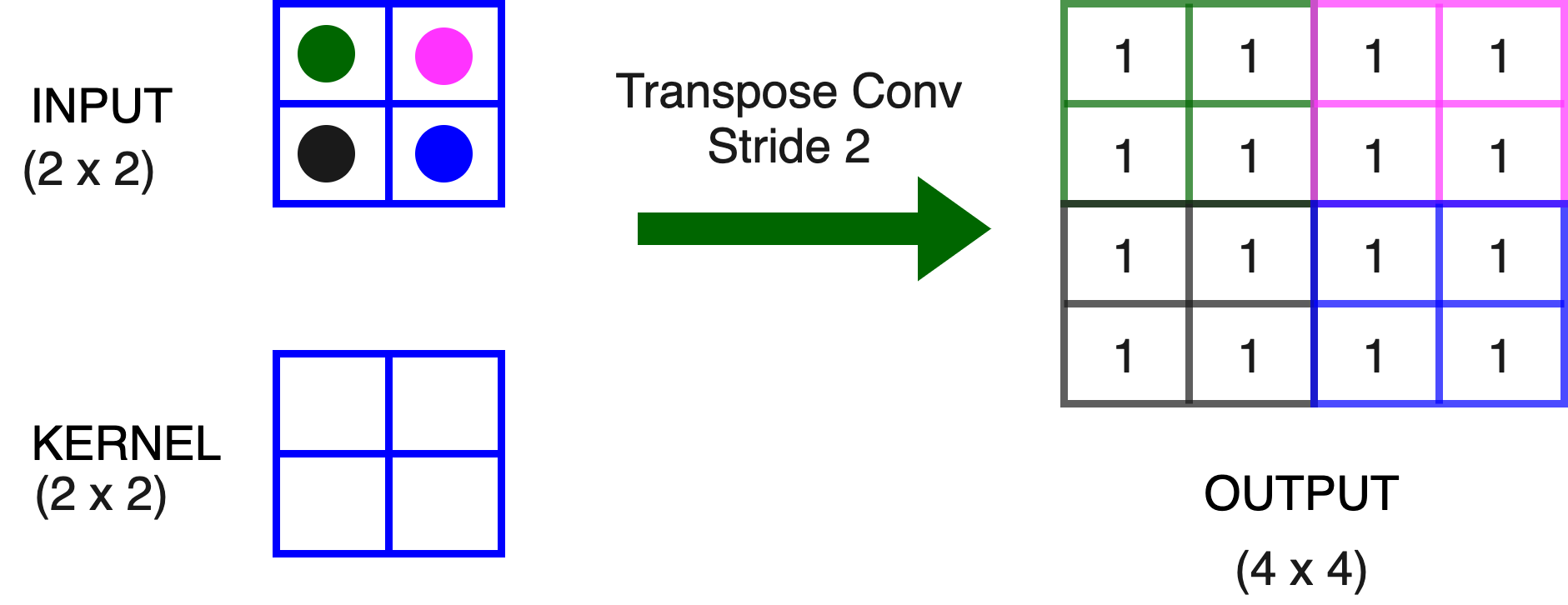}}
\caption{Example of a transpose convolution operation. The cells marked with red X's in the output are cropped out. The numbers denote the number of patches that contributed to the value in the output pixel.}
\label{fig:trans_conv}
\end{center}
\end{figure}

\begin{enumerate}
\item  Consider a 1D image of size 4
\[
\text{Input} = \begin{bmatrix} u_1, & u_2, & u_3, & u_4\end{bmatrix}
\]
Consider a kernel of size $3 \times 1$
\[
\bm{k} = \begin{bmatrix} x, & y, & z\end{bmatrix}
\]
Consider a convolution layer with the kernel $\bm{k}$, stride 1 and zero-padding layer of size 1. Then, the output of the layer acting on the input is
\[
\text{Output} = \begin{bmatrix} yu_1 + zu_2, & xu_1 + yu_2 + zu_3, & xu_2 + yu_3 + zu_4, & xu_3 + yu_4\end{bmatrix}
\]
The steps involved in convolution operator are: pad, dot-product, stride. Note that using padding and stride 1 have ensured the output has the same size as the input.

\item Consider another convolution with the same kernel $\bm{k}$, zero-padding layer 1 but stride 2. Then, the output of the layer acting on the same input as earlier is
\[
\text{Output} = \begin{bmatrix} yu_1 + zu_2, & xu_2 + yu_3 + zu_4 \end{bmatrix}
\]
Note that the size of the output has reduced by a factor of 2. In other words, increasing the stride has allowed us to downsample the input. The question we want to ask is whether we can perform an upsampling in a similar way? This can indeed be done by transposing every operation in a convolution layer.
\begin{itemize}
\item Instead of using a dot-product (inner-product), we will use an outer-product.
\item Instead of skipping pixels in the input (stride) we will skip pixels in the output.
\item Instead of padding, we will need to crop the output.
\end{itemize}

\item Let us now see an example of a transpose convolution layer. Consider a 1D input image of size $2 \times 1$
\[
\text{Input} = \begin{bmatrix} u_1, & u_2\end{bmatrix}
\]
and a kernel of size $3 \times 1$
\[
\bm{k} = \begin{bmatrix} x, & y, & z\end{bmatrix}.
\]
if we perform the outer-product of the input with $\bm{k}$, we will get
\[
\text{Outer product} = \begin{bmatrix} u_1x & u_1y & u_1z \\ u_2x & u_2y & u_2z \end{bmatrix}.
\]
If we use a stride of 2, we will need to shift the rows of the outer-product by 2 
\[
\text{Striding} =  \begin{bmatrix} u_1x & u_1y & u_1z & 0 & 0 \\ 0 & 0 & u_2x & u_2y & u_2z \end{bmatrix}.
\]
After striding is performed we need to add the entries in each column and crop the vector to get the output
\[
\text{Output} = \text{Crop}( \begin{bmatrix} u_1x, & u_1y, & u_1z + u_2 x, &  u_2y, & u_2z \end{bmatrix}) = \begin{bmatrix} u_1x, & u_1y, & u_1z + u_2 x, &  u_2y \end{bmatrix}
\]
where we have cropped out the last few elements (by convention) to get an output which has 2 times the size of the input.
\item We consider transpose convolution in 2D applied on a 2D image of size $(2 \times 2)$. The kernel is taken to be of shape $(3 \times 3)$ with stride 2 and padding (cropping). The action of this transpose convolution is shown in Figure \ref{fig:trans_conv}(a), where we first obtain an image of size $(5 \times 5)$ which is then cropped to give an output of size $(4 \times 4)$. Note that the output pixels get an unequal contribution from the various patches (indicated by numbers in the Figure), which leads to checker-boarding, which is undesirable. Checker-boarding refers to pixel-to-pixel variations in the values of the output image. 

\item One way to avoid checker-boarding, is by ensuring that the filter size is an integer multiple of the stride. Let us repeat the previous example but with a $(2 \times 2)$ kernel. The operation is illustrated in Figure \ref{fig:trans_conv}(b)In this case, we do not need to pad (crop) and each output pixel has an equal contribution. 
\end{enumerate}

We make some remarks:
\begin{enumerate}
\item Transpose convolution layers are also called fractionally-strided layers, because for every one step in the input, we take greater than one step in the output. This is the opposite of what happens in a convolution layer. In a convolution layer, we take step greater than one in the input image, for step of uint one in the output image. 
\item Transpose convolutions are a tool of choice for upscaling \textit{through learnable parameters}.
\item Upscaling is typically done with a reduction in the number of channels,  which is once again the opposite of what is done in convolution layers.
\end{enumerate}

\section{Image-to-image transformations}
Image-to-image to transformations can be seen analogous to function-to-function transformations. These types of networks are typically used in computer vision, super-resolution, style transfer, and also in computational physics where we (say) map the source (RHS) field to the solution of the PDE. 

\begin{figure}[htbp!]
\begin{center}
\includegraphics[width=0.9\textwidth]{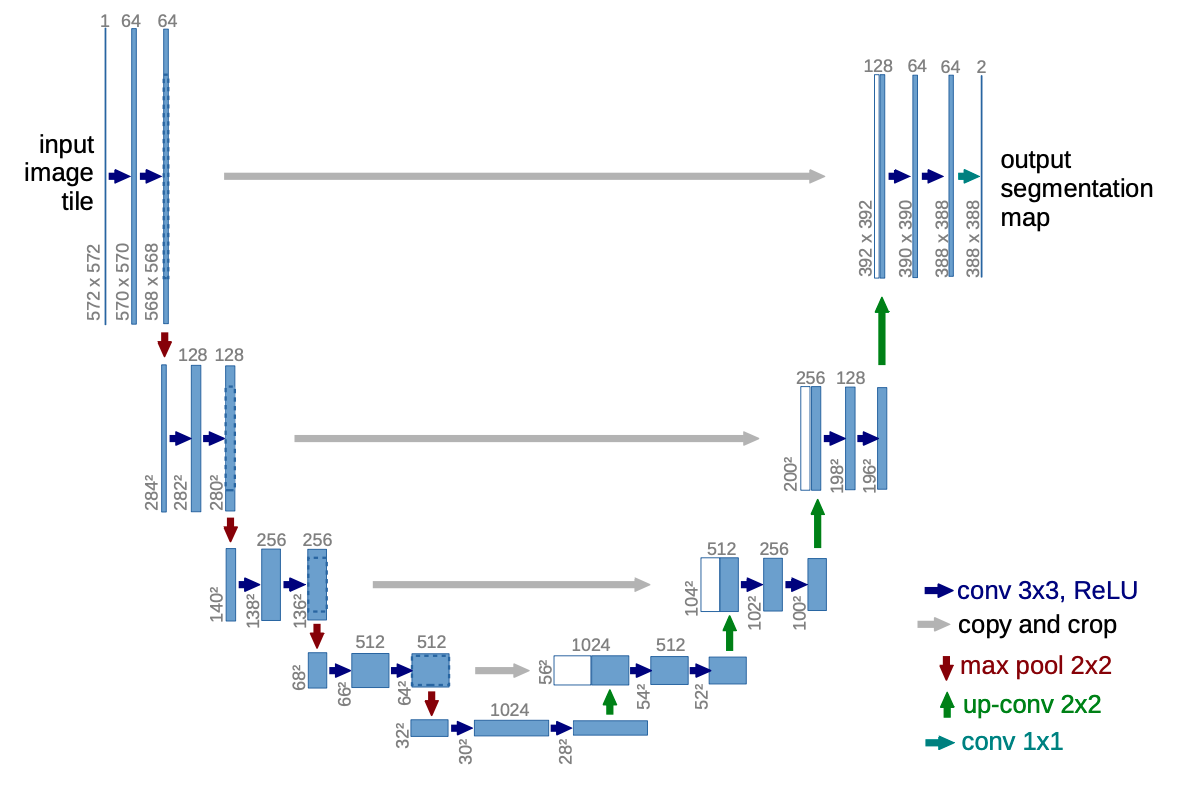}
\caption{Example of a U-Net taken from \cite{ronneberger2015unet}.}
\label{fig:unet}
\end{center}
\end{figure}

We will discuss a particular type of network for such transformations, which is known as U-Nets \cite{ronneberger2015unet}. In a U-Net (see Figure \ref{fig:unet}), there a down is a downward branch which takes an input image and downscales the images using a number of convolution layers and pooling operations. As we go down this branch, the number of channels typically increase. After we reach the coarsest level, we have an upward branch that scales up the image and reduced the number of channels using transpose convolution type operations, to finally give the output image. In addition to these branches, the U-Net also makes use of skip connections that combines information at a particular scale in the downward branch to the information in the upward branch at the same scale. These connections are similar to what are used in ResNets. In the upscaling branch of the U-net, if you consider the activation at one point, you will see they come from two different sources. One of these is the from the same spatial scale in the down-scaling branch of the U-net, and the other is from the coarser scales of the upscaling branch of the U-net. 

\begin{remark}
The U-net architecture shares many common features with the V-cycle that is typically used in multigrid preconditioners. 
\end{remark}

\begin{remark}
We can also think of a the U-net as an encode-decoder network with the additional feature of including skip connections. 

\end{remark}

%% file: DeepONets.tex

\chapter{Operator Networks}

\section{The problem with PINNs}

Recall that a typical MLP $\y = \bm{\mathcal{F}}(\x;\btheta)$ is a function that takes as input $\x \in \Ro^d$ and gives an output $\y \in \Ro^d$ with trainable weights $\btheta$. Also, as we discussed in Chapter \ref{chap:PINNs}, a PINN is a network of the form  $\bm{u}(\x) = \bm{\mathcal{F}}(\x;\btheta)$ taking as input the independent variable $\x$ of the underlying PDE and giving the solution $\bm{u}(\x)$ (of the PDE) as output. The network is trained by minimizing the weighted sum of the PDE and boundary residual. However, this is just one instance of the solution of the PDE for some given boundary condition and source term. For instance, if we consider the PDE
\begin{equation}\label{eqn:model_PDE}
\begin{aligned}
\nabla \cdot (\kappa \nabla u) &= f(x), \quad \x \in \Omega = [0,1] \times [0,1]\\
  u(x) & = g(x), \quad \x \in \partial \Omega
\end{aligned}
\end{equation}
and train a PINN $\mathcal{F}(\x;\btheta)$ to minimize the loss function
\[
\Pi(\btheta) = \frac{1}{N_v} \sum_{i=1}^{N_v} | \nabla \cdot (\kappa \nabla \mathcal{F}(\x_i;\btheta)) - f(\x_i) |^2 + \frac{\lambda_b}{N_b} \sum_{i=1}^{N_b} | \mathcal{F}(\x_i;\btheta) - g(\x_i) |^2
\]
Then, if $\btheta^* = \argmin{\btheta} \Pi(\btheta)$, the PINN solving \eqref{eqn:model_PDE} will be $u(\x) =\mathcal{F}(\x;\btheta^*)$. However, if we change $f$ or $g$ in \eqref{eqn:model_PDE}, we have no reason to believe that the same trained network would work. In fact, we would need to retrain the network (with perhaps the same architecture) for the new $f$ and $g$. This can be quite cumbersome to do, and we would ideally like to avoid it. In this chapter, we will see ways by which we can overcome this issue.

\section{Parametrized PDEs}
\begin{figure}[htbp]
\begin{center}
\includegraphics[width=0.6\textwidth]{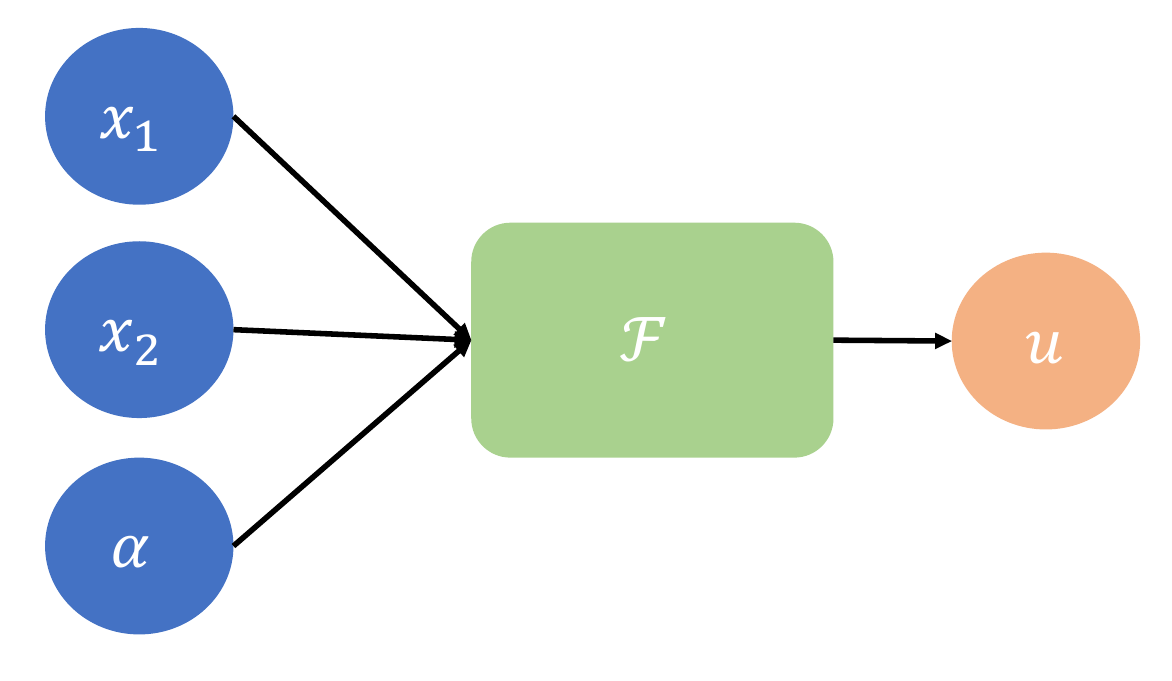}
\caption{Schematic of a PINN with a parameterized input.}
\label{fig:pinn_param}
\end{center}
\end{figure}

Assume the the source term $f$ in \eqref{eqn:model_PDE} is given as a parametric function $f(\x;\alpha)$. For instance, we could have
\[
f(x_1,x_2;\alpha) = 4 \alpha x_1 (1-x_1) x_2 (1-x_2) 
\]
Then we could train a PINN that accommodates for the parametrization by considering a network that takes as input both $\x$ and $\alpha$, i.e., $\mathcal{F}(\x, \alpha;\btheta)$. This is shown in Figure \ref{fig:pinn_param} This network can be trained by minimizing the loss function
\[
\Pi(\btheta) = \frac{1}{N_a} \sum_{j=1}^{N_a}\left[\frac{1}{N_v} \sum_{i=1}^{N_v} | \nabla \cdot (\kappa \nabla \mathcal{F}(\x_i,\alpha_j;\btheta)) - f(\x_i,\alpha_j) |^2 + \frac{\lambda_b}{N_b} \sum_{i=1}^{N_b} | \mathcal{F}(\x_i,\alpha_j;\btheta) - g(\x_i) |^2 \right]
\]
Note that we have to also consider collocation points for the parameter $\alpha$ while constructing the loss function. If $\btheta^* = \argmin{\btheta} \Pi(\btheta)$, then the solution to the parameterized PDE would be $u(\x,\alpha) = \mathcal{F}(\x, \alpha;\btheta^*) $. Further, for any new value of $\alpha = \hat{\alpha}$ we could find the solution by evaluating $\mathcal{F}(\x,\hat{\alpha};\btheta^*)$. We could use the same approach if there was a way of parameterizing the functions $\kappa(\x)$ and $g(\x)$. 

However, what if we wanted the solution for an arbitrary, non-parametric $f$?  In order to do this, we need to find a way to approximate \textbf{operators that map functions to functions.}

\section{Operators}
Consider a class of functions $\bm{a}(\y) \in A$ such that $\bm{a}: \Omega_Y\ \rightarrow \Ro^D$. The functions in this class might have certain properties, such as $\bm{a} \in C(\Omega_Y)$ or $\bm{a} \in L^2(\Omega_Y)$. Also consider the operator $\mathcal{N}: A \mapsto C(\Omega_X)$, with $\bm{u}(\x) = \mathcal{N}(\bm{a})(\x)$ for $\x \in \Omega_X$. Let us see some examples of operators $\mathcal{N}$.

\begin{enumerate}
\item Consider the PDE
\begin{equation}
\begin{aligned}
\nabla \cdot (\kappa \nabla u) &= f, \quad \x \in \Omega\\
  u & = g, \quad \x \in \partial \Omega 
\end{aligned}
\end{equation}

For this PDE, $\Omega_X = \Omega_Y = \Omega$ and the operator $\mathcal{N}$ maps the RHS $f$ to the solution (temperature) $u$ of the PDE. That is, $u = \mathcal{N}(f)(\x)$. The input and the output to the operator are related by the equation above where it is assumed that $\kappa$ and $g$ are given and fixed. 

\item Consider the PDE
\begin{equation}
\begin{aligned}
\nabla \cdot (\kappa \nabla u) &= f, \quad \x \in \Omega\\
  u & = g, \quad \x \in \partial \Omega
\end{aligned}
\end{equation}
which is the same as the previous PDE but we are assuming that the conductivity field $\kappa$ might change for the model, instead of the RHS. Then, $\Omega_X = \Omega_Y = \Omega$ and the operator $\mathcal{N}$ maps the conductivity $\kappa$ to the solution $u$ of the PDE. That is, $u = \mathcal{N}(\kappa)(\x)$. The input and the output to the operator are related by the equation above where it is assumed that $f$ and $g$ are given and fixed.

\item Once again, consider the same PDE but with conductivity and the boundary condition being allowed to change
\begin{equation}
\begin{aligned}
\nabla \cdot (\kappa \nabla u) &= f(x), \quad \x \in \Omega\\
  u(x) & = g(x), \quad \x \in \partial \Omega
\end{aligned}
\end{equation}
Then, the operator $\mathcal{N}$ maps the boundary condition $g$ and the conductivity $\kappa$ to the solution $u$ of the PDE. 
That is, $u = \mathcal{N}(\kappa,g)(\x)$. In this case the input to the operator are two functions $(g,\kappa)$ and the output is a single function. Therefore $\Omega_X = \Omega$, while $\Omega_Y =  \Omega \times \partial \Omega$. The input and the output are related through the solution to the PDE above where it is assumed that $f$ is given and fixed.

\item Now consider the equations of linear isotropic elasticity posed on a three-dimensional domain $\Omega \subset \mathbb{R}^3$,
\begin{equation}
\begin{aligned}
\nabla \cdot \big(\lambda (\nabla \cdot \bm{u}) \bm{I} + 2 \mu \nabla^S(\bm{u}) \big) &= \bm{f}(\x), \quad \x \in \Omega\\
  \bm{u}(\x) & = \bm{g}(\x), \quad \x \in \partial \Omega.
\end{aligned}
\end{equation}
Consider the operator defined by $\bm{u}(\x) = \mathcal{N} (\bm{f})(\x)$. Here the input function, $\bm{f}: \Omega \rightarrow \Ro^3$, and the output function $\bm{u}: \Omega \rightarrow \Ro^3$. The two are related by the equations above where $\lambda, \mu, \bm{g}$ are given and fixed. 

\item Now, consider a different PDE. In particular, the advection-diffusion-reaction equation,
\begin{equation}
\begin{aligned}
\df{u}{t} + \bm{a}\cdot \nabla u - \kappa \nabla^2 u + u(1-u) &= f, \quad (\x,t) \in \Omega \times (0,T]\\
  u(\x,t) & = g(\x,t), \quad (\x,t) \in \partial \Omega \times (0,T]\\
  u(\x,0) &= u_0(\x) , \quad \x \in \Omega.
\end{aligned}
\end{equation}
We want to find the operator $\mathcal{N}$ maps the initial condition $u_0$ to the solution $u$ at the final time $T$, i.e., $u(\x,T) = \mathcal{N}(u_0)(\x)$. In this case $\Omega_X = \Omega_Y = \Omega$. Further the input and the output functions are related to each other via the solution of the PDE above with $\bm{a}, \kappa, f, g $ given and fixed. 

\end{enumerate}

\begin{remark}

It is often useful to determine whether an operator is linear or non-linear. This is because if it is linear it can be well approximated by another linear operator. In the cases considered above the operators in examples 1 and 4 were linear whereas those in examples 2,3, and 5 were nonlinear. 

\end{remark}

We are interested in networks that approximate the operator $\mathcal{N}$. We will see how we can do this in the next section. These types of networks are often referred to as Operator Networks. They are two popular versions of these networks. One is referred to as a Deep Operator Network, or a DeepONet, and the other is referred to as a Fourier Neural Operator. We describe the DeepONet in the next section.

\section{Deep Operator Network (DeepONet) Architecture}

\begin{figure}[htbp]
\begin{center}
\includegraphics[width=0.6\textwidth]{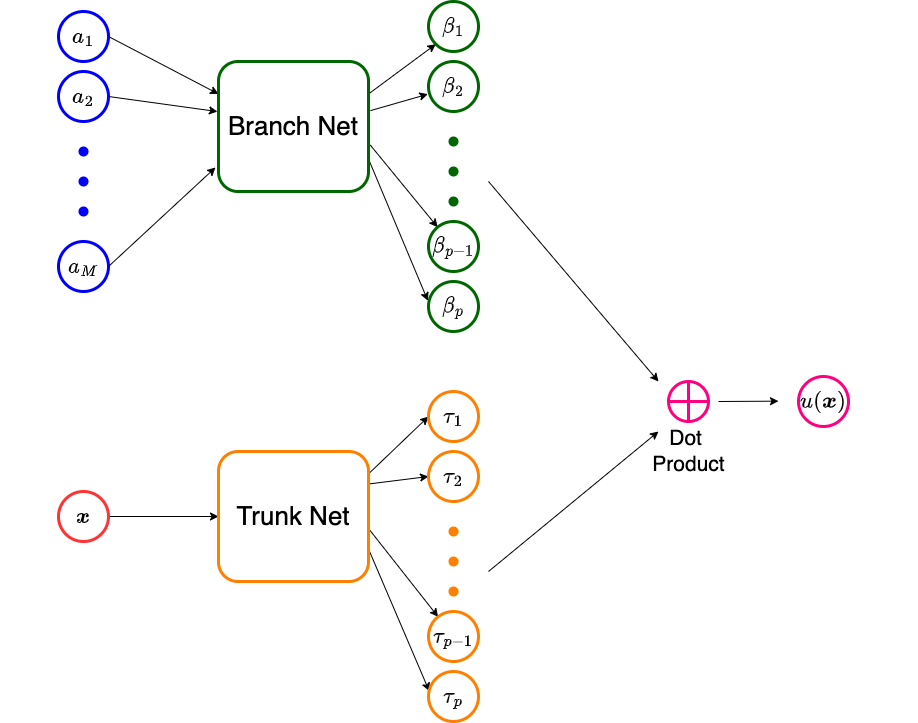}
\caption{Schematic of a DeepONet}
\label{fig:deeponet}
\end{center}
\end{figure}

Operator networks were first proposed by Chen and Chen \cite{chen95}, where they considered only shallow networks with a single hidden layer. This idea was rediscovered and extended to deep architectures more recently in \cite{deeponet} and were called DeepONets. A standard DeepONet comprises two neural networks. We describe below its construction to approximate an operator $\mathcal{N} : A \rightarrow U$, where $A$ is a set of functions of the form $a: \Omega_Y \subset \Ro^d \rightarrow \Ro$ while $U$ consists of functions of the form $u: \Omega_X \subset \Ro^D \rightarrow \Ro$. Furthermore, we assume that point-wise evaluations of both class of functions is possible. The architecture for the DeepONet for this operator is illustrated in Figure \ref{fig:deeponet}. It is explained below:
\begin{itemize}
\item Fix $M$ distinct sensor points $\y^{(1)},..., \y^{(M)}$ in $\Omega_Y$.
\item Sample a function $a \in A$ at these sensor points to get the vector $\bm{a} =  [a(\y^{(1)}), ... , \ a(\y^{(M)})]^\top \in \Ro^M$.

\item Supply $\bm{a}$ as the input to a sub-network, called the \textit{branch net} $\mathcal{B}(.;\bm{\theta}_B):\Ro^M \rightarrow \Ro^p$, whose output would be the vector $\bm{\beta} = [\beta_1(\bm{a}), ..., \ \beta_p(\bm{a})]^\top\in \Ro^p$. Here $\bm{\theta}_B$ are the trainable parameters of the branch net. The dimension of the output of the branch is relatively small, say $p \approx 100$.


\item Supply $\x$ as an input to a second sub-network, called the \text{trunk net} $\mathcal{T}(.;\bm{\theta}_T): \Ro^D \rightarrow \Ro^p$, whose output would be the vector $\bm{\tau} = [\tau_1(\x), ..., \ \tau_p(\x)]^\top \in \Ro^p$. Here $\bm{\theta}_T$ are the trainable parameters of the trunk net.

\item Take a dot product of the outputs of the branch and trunk nets to get the final output of the DeepONet $\widetilde{\mathcal{N}}(.,.;\bm{\btheta}):\Ro^D \times \Ro^M \rightarrow \Ro$ which will approximate the value of $u(\x)$
\begin{equation}
u(\x) \approx \widetilde{\mathcal{N}}(\x,\bm{a};\bm{\btheta}) = \sum_{k=1}^p \beta_k(\bm{a}) \tau_k(\x). \label{eq:deeponet}
\end{equation}
where the trainable parameters of the DeepONet will be the combined parameters of the branch and trunk nets, i.e., $\bm{\btheta} = [\btheta_T, \btheta_M]$.
\end{itemize}

In the above construction, once the DeepONet is trained (we will discuss the training in the following section), it will approximate the underlying operator $\mathcal{N}$, and allow us to approximate the value of any $\mathcal{N}(a)(\x)$ for any $a \in A$ and any $\x \in \Omega_X$. Note that in the construction of the DeepONet, the $M$ sensor points need to be pre-defined and cannot change during the training and evaluation phases.


We can make the following observations regarding the DeepONet architecture:
\begin{enumerate}
    \item The expression in (\ref{eq:deeponet}) has the form of representing the solution as the sum of a series of coefficients and functions. The coefficients are determined by the branch network, while the functions are determined by the trunk network. In that sense the DeepONet construction is similar to that of what is used in the spectral method or the finite element method. There is a critical difference though. In these methods, the basis functions are pre-determined and selected by the user. However, in the DeepONet these functions are determined by the trunk network and their final form depends on the data used to train the DeepONet. 
    \item Architecture of the branch sub-network: When points for sampling the input function are chosen randomly, the appropriate architecture for the branch network comprises fully connected layers. Further recognizing that the dimension of the input to this network can be rather large $N_1 \approx 10^4$, while the output is typically small $p \approx 10^2$, this network can be thought of as an encoder. 
    
    When points for sampling the input function are chosen on a uniform grid, the appropriate architecture for the branch network comprises convolutional layer layers. In that case this network maps an image of large dimension ($N_1 \approx 10^4$) to a latent vector of small dimension, $p \approx 10^2$. Thus it is best represented by a convolutional neural network.  
    
    \item Broadly speaking, there are two ways of improving the experssivity of the DeepONet. These involve increase the number of network parameters in the branch and trunk sub-networks, and increasing the dimension $p$ of the latent vectors formed by these sub-networks. 
    
    
\end{enumerate}
\section{Training DeepONets}
Training a DeepONet is typically supervised, and requires pairwise data. The following are the main steps involved:
\begin{enumerate}
\item Select $N_1$ representative function $a^{(i)}$, $1 \leq i \leq N_1$ from the set $A$. Evaluate the values of these functions at the $M$ sensor points, i.e., $a^{(i)}_j = a^{(i)}(\y^{(j)})$ for $1 \leq j \leq M$. This gives us the vectors $\bm{a}^{(i)} = [a^{(i)}(\y^{(1)}),...,a^{(i)}(\y^{(M)})]^\top \in \Ro^M$ for each $1 \leq i \leq N_1$.
\item For each $a^{(i)}$, determine (numerically or analytically) the corresponding functions $u^{(i)}$ given by the operator $\mathcal{N}$. 
\item Sample the function $u^{(i)}$ at $N_2$ points in $\Omega_X$, i.e., $u^{(i)}(\x^{(k)})$ for $1 \leq k \leq N_2$. 
\item Construct the training set 
\[
\mathcal{S} = \left \{\Big(\bm{a}^{(i)}, \x^{(k)}, u^{(i)}(\x^{(k)})\Big) : 1 \leq i \leq N_1, \ 1 \leq k \leq N_2 \right\}
\]
which will have $N_1 \times N_2$ samples.
\item Define the loss function
\[
\Pi(\btheta) = \frac{1}{N_1 N_2} \sum_{i=1}^{N_1} \sum_{k=1}^{N_2} | \widetilde{N}(\x^{(k)}, \bm{a}^{(i)};\btheta) - u^{(i)}(\x^{(k)})|^2.
\]
\item Training the DeepONet corresponds to finding $\btheta^* = \argmin{\btheta} \Pi(\btheta)$.
\item Once trained, then given any new $a \in A$ samples at the $M$ sensor points (which gives the vector $\bm{a} \in \Ro^M$), and a new point $\x \in \Omega_X$, we can evaluate the corresponding prediction $u^*(\x) = \widetilde{N}(\x, \bm{a};\btheta^*)$. 
\end{enumerate}

\begin{remark}
We need not choose the same $N_2$ points across all $i$ in the training set. In fact, these can be chosen randomly leading to a more diverse dataset.
\end{remark}

\begin{figure}[htbp]
\begin{center}
\includegraphics[width=0.7\textwidth]{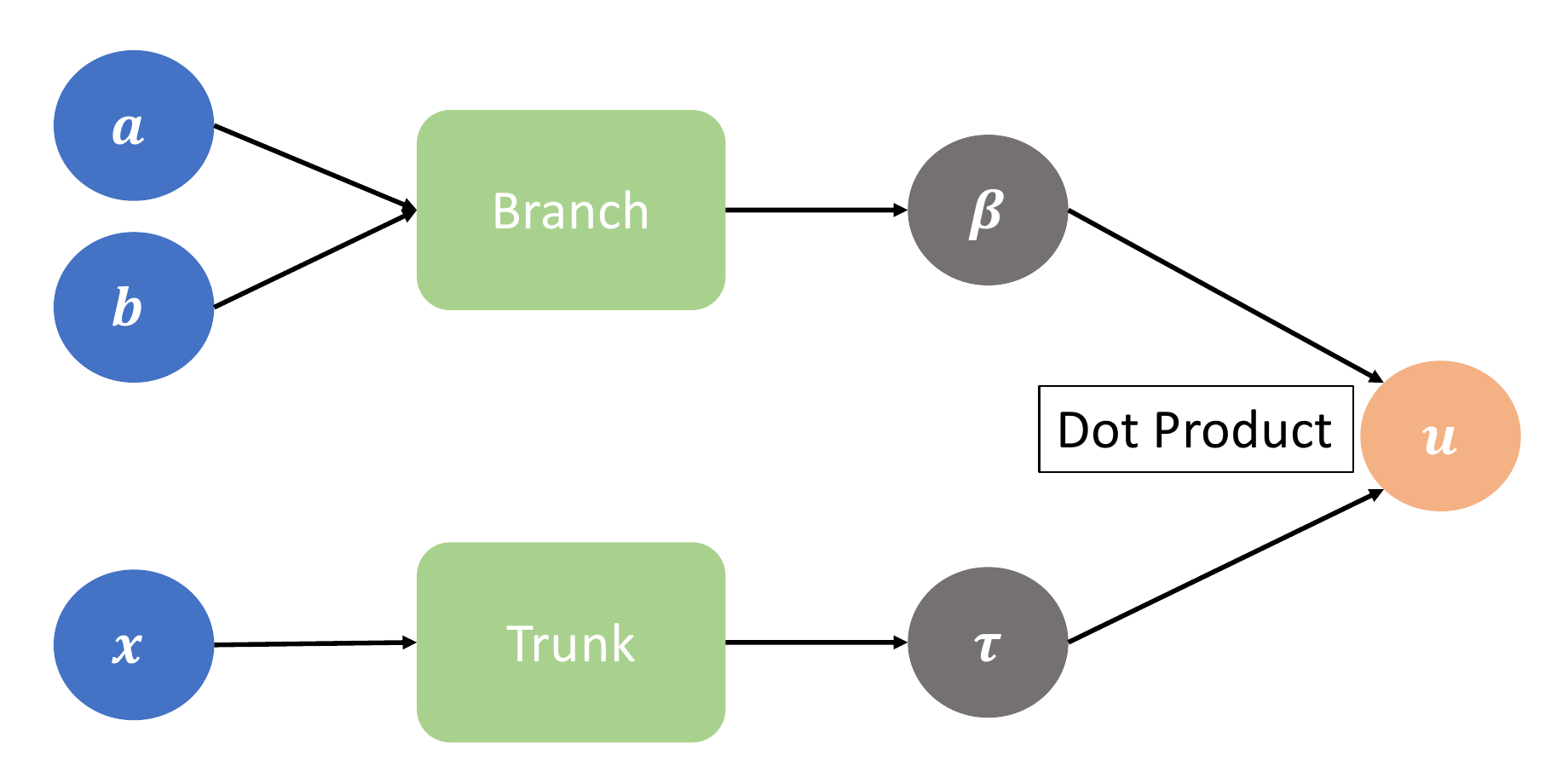}
\caption{Schematic of a DeepONet with two input functions.}
\label{fig:deeponet_multi_input}
\end{center}
\end{figure}

\begin{remark}
The DeepONet can be easily extended to the case where the input comprises multiple functions. In this case, the trunk network remains the same, however the branch network now has multiple vectors as input. The case corresponding to two input functions, $a(\y)$ and $b(\y)$, which when sampled yield the vectors, $\bm{a}$ and $\bm{b}$, is shown in Figure \ref{fig:deeponet_multi_input}.
\end{remark}

\begin{remark}
The DeepONet can be easily extended to the case where the output comprises multiple functions (say $D$ such functions). In this case, the output of the branch and trunk network leads to $D$ vectors each with dimension $p$. The solution is then obtained by taking the dot product of each one of these vectors.  The case corresponding to two output functions $u_1(\x)$ and $u_2(\x)$ is shown in Figure \ref{fig:deeponet_multi_input}. 
\end{remark}

\begin{figure}[htbp]
\begin{center}
\includegraphics[width=0.7\textwidth]{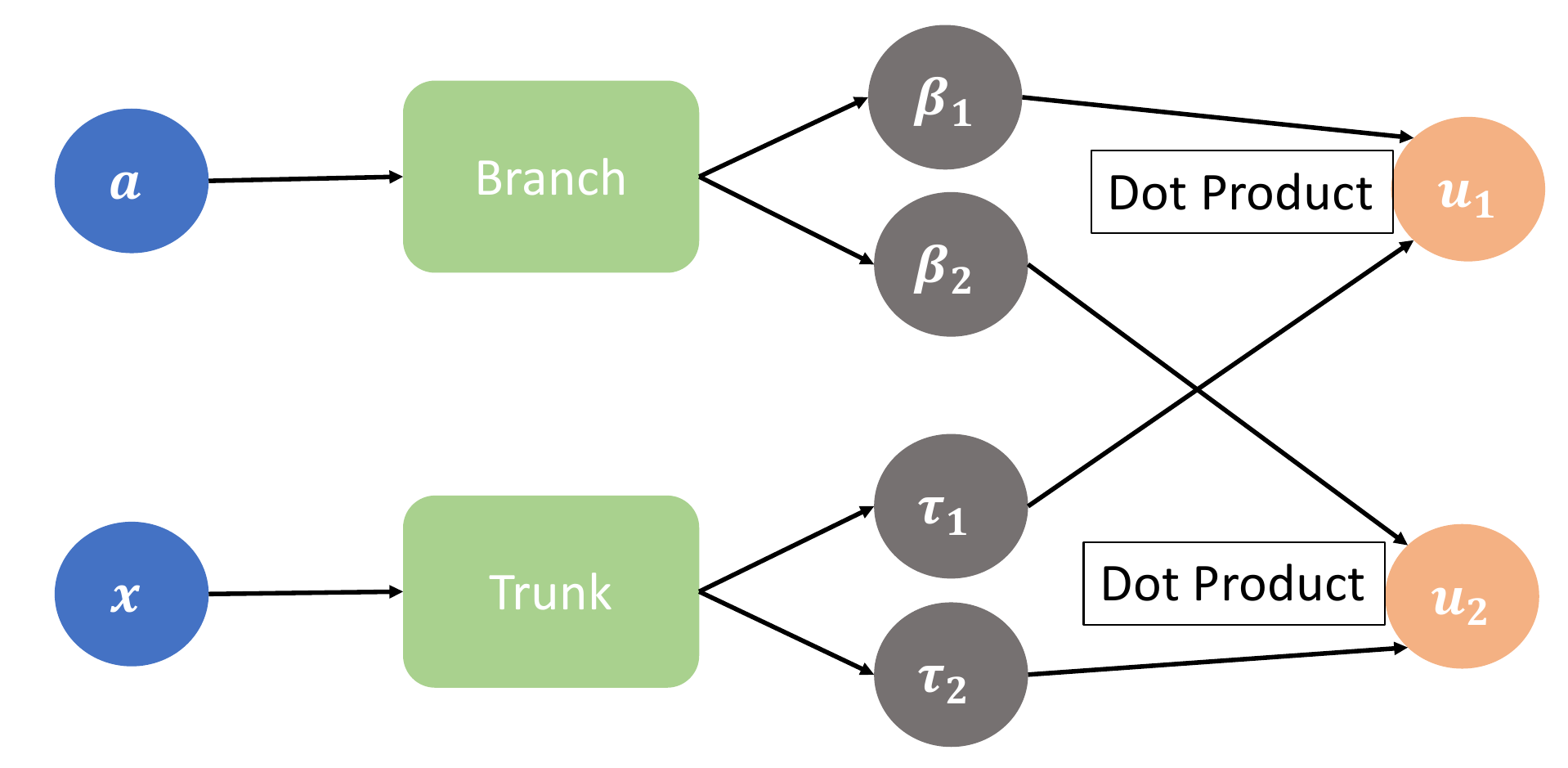}
\caption{Schematic of a DeepONet with two output functions.}
\label{fig:deeponet_multi_ouput}
\end{center}
\end{figure}

\section{Error Analysis for DeepONets}
There is a universal approximation theorem for a shallow version of DeepONets by Chen and Chen \cite{chen95}
\begin{theorem}
Suppose $\Omega_X$ and $\Omega_Y$ are compact sets in $\Ro^D$ (or more generally a Banach space) and $\Ro^d$, respectively. Let $\mathcal{N}$ be a nonlinear, continuous operator mapping $V \subset C(\Omega_Y)$ into $C(\Omega_X)$. Then given $\epsilon > 0$, there exists a DeepONet with $M$ sensors and a single hidden layer of width $P$ in the branch and trunk nets such that
\[
\max_{\substack{\x \in \Omega_X\\ a \in A}} |\widetilde{\mathcal{N}}(\x, \bm{a}; \btheta) - \mathcal{N}(a)(\x)| < \epsilon
\]
for a large enough $P$ and $M$.
\end{theorem}

This result has been extended to a deeper version of the network in \cite{deeponet}, and generalized further by removing the compactness assumptions on the spaces \cite{sid_deeponet}.

Recently in \cite{patel2022variationally}, the authors have developed an estimate of the error in a DeepONet that clearly pinpoints the different sources of error in a DeepONet. This estimate states, the error measured in the $L^2(\Omega)$ norm is bounded as
\begin{equation}\label{eqn:varmion_err}
\max_{a \in A}\|\widetilde{\mathcal{N}}(\x, \bm{a}; \btheta) - \mathcal{N}(a)(\x)\|_{L^2(\Omega)} \leq C \left(\epsilon_h + \sqrt{\epsilon_t} + \epsilon_s  +  M^{-\alpha_1} + (N_2)^{- \alpha_2}\right)
\end{equation}
where $\epsilon_h$ is the error made by the numerical solver used to generate the approximate target solutions $u^{(i)}$ in the training set (as compared to the exact solutions), $\epsilon_t$ is the final training error/loss attained, while $\epsilon_s$ bounds the distance of any $a \in A$ from the set of functions $\{a^{(i)}\}_{i=1}^{N_1}$ used to construct the construct the training set, i.e., an estimate of how well the training samples covers the input space $A$. Further, since the input function is evaluated at $M$ finite sensor nodes, while the output is evaluated at $N_2$ output nodes, this will lead to an additional discretization (or quadrature) error which is given by the last two terms in \eqref{eqn:varmion_err}. Note that this is similar to the error estimates obtained for PINNs in \eqref{eqn:err_bd4}.

\section{Physics-Informed DeepONets}
Recall that DeepONets approximates $u(\x) = \mathcal{N}(a)(\x) \approx \widetilde{N}(\x, \bm{a};\btheta)$. Assume that the pair $a$ and $u$ satisfy a PDE. For example,
\begin{equation}
\begin{aligned}
\nabla \cdot (\kappa u) &= f \ \text{in } \Omega\\
u &=g \ \text{on } \partial \Omega
\end{aligned}
\end{equation}
where $\kappa$ and $g$ are prescribed. To construct the operator $\mathcal{N}$ that maps $f$ to $u$, we need to solve the PDE externally. However, in addition to this, we can also use a PINN-type loss function and add that to the total loss. This is the idea of Physic-Informed DeepONets proposed in \cite{pi_deeponet}. So for the above model PDE, the loss additional physics-based loss would would be, 
\begin{equation}
\begin{aligned}
\Pi_p (\btheta) &= \frac{1}{\bar{N}_1} \frac{1}{\bar{N}_2} \sum_{i=1}^{\bar{N}_1} \sum_{k=1}^{\bar{N}_2} \left| \nabla_{\x} \cdot \left(\kappa \nabla_{\x} \widetilde{\mathcal{N}}(\x^{(k)}, \bm{f}^{(i)}; \btheta)\right) - f^{(i)}(\x^{(k)})\right|^2. 
\end{aligned}
\end{equation}
This is in addition to the standard data-driven loss function which, for this example is given by
\begin{equation}
\Pi_d (\btheta) = \frac{1}{N_1 N_2} \sum_{i=1}^{N_1} \sum_{k=1}^{N_2} | \widetilde{N}(\x^{(k)}, \bm{f}^{(i)};\btheta) - u^{(i)}(\x^{(k)})|^2.
\end{equation} 
The total loss function is a weighted sum of these two terms:
\begin{equation}
\Pi(\btheta) =  \Pi_d (\btheta) + \lambda \Pi_p (\btheta),
\end{equation} 
where $\lambda$ is a hyper-parameter. 
A few comments are in order:
\begin{enumerate}
    \item The output sensor points used in the physics-based loss function are usually distinct from the output sensor points used in the data-driven loss term. The former represent the locations at which we wish to minimize the residual of the PDE, while the latter represent the points at which the solution is available to us through external means. The total number of the output sensor points in the physics-based portion of the loss function is denoted by $\bar{N}_2$, whereas in the data-driven loss function it is denoted by $N_2$. 
    \item The set of input functions used to construct the physics-based loss function is usually distinct from the set of input functions used to construct the data-driven loss function. The former set represents the functions for which we wish to minimize the residual of the PDE, while the latter set represents the collection of input functions for which the solution is available to us through external means. The total number of functions in the set used to construct the the physics-based portion of the loss function is denoted by $\bar{N}_1$, whereas the total number of functions in the set used to construct the data-driven portion of the loss function is denoted by $N-1$.      
\end{enumerate}

As earlier, we train the network by finding $\btheta^* = \argmin{\btheta} \Pi(\btheta)$ and approximate the solution for a new $a$ by $u^*(\x) = \widetilde{N}(\x, \bm{a};\btheta^*)$. The advantages of adding the extra physics-based loss are:
\begin{enumerate}
    \item It reduces the demand on the amount of data in the data-driven loss term. What is means is that we don't have to generate as many solutions of the PDE for training the DeepONet. 
    \item It makes the network more robust in that it becomes more likely to produce accurate solutions for the type of input functions not included in the training set for the data-driven loss term. 
\end{enumerate}

\section{Fourier Neural Operators - Architecture}

We now introduce and discuss Fourier Nerual Operators (FNOs) \cite{li2020fourier}. We discuss their architecture in this section, and then discuss other aspects in the following section. 

Our approach in developing the architecture for a FNO will be to begin with the architecture of a typical feed-forward MLP that maps a scalar to another scalar, and systematically extend it so that the extended version maps a scalar valued function to another scalar valued function.

\begin{figure}[htbp]
\begin{center}
\includegraphics[width=1.0\textwidth]{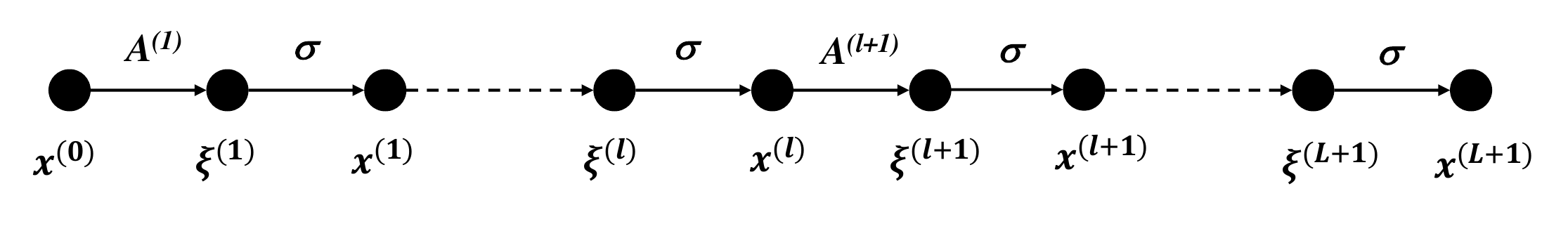}
\caption{Computational graph for a feed-forward MLP.}
\label{fig:comp_graph_mlp_for}
\end{center}
\end{figure}

In Figure \ref{fig:comp_graph_mlp_for} we have plotted the computational graph of an MLP. We are focused only on the forward part (not the back-propagation) part of this network. For simplcity, we assume that the input $\x^{(0)} = x$ is a scalar and the output $\x^{(L)} = y$ is also a scalar. Further all the other hidden variables (with the exception of $\bm{\xi}^{(L+1)}$) are vectors with $H$ components. That is, the width of each layer is $H$.

\begin{figure}[htbp]
\begin{center}
\includegraphics[width=1.0\textwidth]{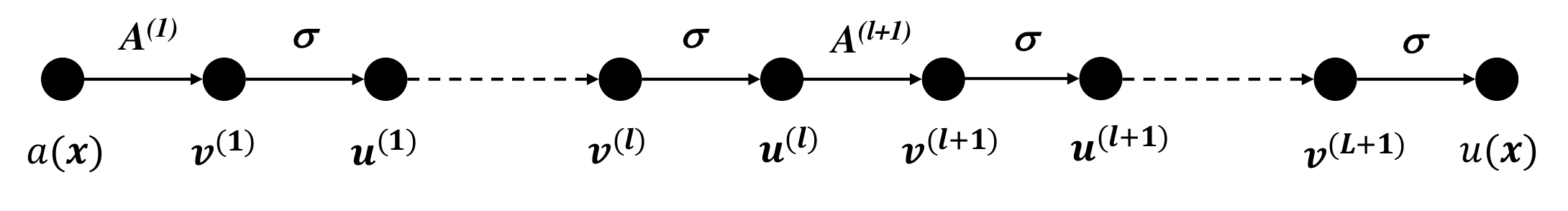}
\caption{Computational graph for a feed-forward Fourier Neural Operator (FNO) network.}
\label{fig:comp_graph_fno}
\end{center}
\end{figure}

The first step in this process will be to replace the input and the output with functions. The input will now be the function $a(\x): \Omega \mapsto \Ro^1$. Similarly the output is the function $u(\x): \Omega \mapsto \Ro^1$. This leads us to the computational graph shown in Figure \ref{fig:comp_graph_fno}.

The next step is consider the variables in the hidden layers. In the MLP, these were all vectors with $H$ components. In the FNO, these will be functions with $H$ components. That is, 
\begin{equation}
\bm{v}^{(1)}, \cdots, \bm{v}^{(L)}, \bm{u}^{(1)}, \cdots, \bm{u}^{(L)}: \Omega \mapsto \Ro^H. 
\end{equation}
As shown in Figure \ref{fig:comp_graph_fno}, $\bm{v}^{(n)}$ and $\bm{u}^{(n)}$ are the counterparts of $\bm{\xi}^{(n)}$ and $\x^{(n)}$, respectively. Further since $\bm{\xi}^{(L+1)}$ was a scalar, we will set $\bm{v}^{(L+1)}$ to be a scalar valued function. 

We are now done with extending the input, the output and the variables in the hidden layers from vectors to functions. Next, we need to extend the operators that transform vectors to vectors within an MLP to those that transform functions to functions within an FNO.

We begin with the operator $A^{(1)}$, which in an MLP is an affine map from a vector with one component to a vector with $H$ components. Its straightforward extension to functions is,
\begin{equation}
\bm{v}^{(1)} (\x) = A^{(1)}(a)(\x),
\end{equation}
where 
\begin{equation}
v^{(1)}_i (\x) = W_{i}^{(1)}a(\x) + b_{i}^{(1)}, \qquad i = 1,\cdots, H.
\end{equation}
Here $W_{i}^{(1)}$ and $b_{i}^{(1)}$ are the weights and biases associated with this layer. 

Similarly, in an MLP the operator $A^{(L+1)}$ is an affine map from a vector with $H$ components to a vector with $1$ component. It's straightforward extension to functions is,
\begin{equation}
v^{(L+1)} (\x) = A^{(L+1)}(\bm{u}^{(L)})(\x),
\end{equation}
where 
\begin{equation}
v^{(L+1)} (\x) = W_{i}^{(L+1)}u_i^{(L)}(\x) + b^{(L+1)}, \qquad i = 1,\cdots, H.
\end{equation}
Here $W_{i}^{(L+1)}$ and $b^{(L+1)}$ are the weights and the bias associated with this layer. 

Next we describe the action of the activation on input functions. It is a simple extension of the activation function applied to the point-wise values of the input function. That is,
\begin{equation}
\bm{u}^{(n)} (\x) = \sigma (\bm{v}^{(n)})(\x),
\end{equation}
where 
\begin{equation}
u^{(n)}_i (\x) = \sigma (v_i^{(n)}(\x)), \qquad i = 1,\cdots, H.
\end{equation}

Finally it remains to extend the operators $A^{(n)}, n = 2,\cdots, L$ to functions. These are defined as, 
\begin{equation}
\bm{v}^{(n+1)} (\x) = A^{(n+1)}(\bm{u}^{(n)})(\x),
\end{equation}
where 
\begin{eqnarray}
v^{(n+1)}_i  (\x) &=& W_{ij}^{(n+1)}u_j^{(n)}(\x) + b_i^{(n+1)} \\
& & + \int_{\Omega} \kappa_{ij}^{(n+1)}(\y - \x) u_j^{(n)}(\y) d\y, \qquad i = 1,\cdots, H.
\end{eqnarray}
In the equation above the summation over the dummy index $j$ (from 1 to $H$) is implied. 
The new term that appears in this equation is a convolution. It is motivated by the observation that a large class of linear operators can be represented as convolutions. An example is the so-called Green's operator which maps the right hand side (also called the forcing function) of a linear PDE to its solution. The functions $\kappa_{ij}^{(n+1)}(\bm{z})$ are called the kernels of the convolution. We note that there are $H^2$ of these functions in each layer. 

It is instructive to examine a specific case of a convolution. Let us consider $\Omega = [0,L_1] \times [0,L_2]$, where we denote the two coordinates by either $x_1$ and $x_2$, or $y_1$ or $y_2$. In this case we may write the convolution as,
\begin{eqnarray}
v_i  (x_1,x_2) &=&  \int_{0}^{L_1} \int_{0}^{L_2} \kappa_{ij}(y_1-x_1,y_2-x_2) u_j(y_1,y_2) \; dy_2 dy_1, \qquad i = 1,\cdots, H.
\label{eq:conv2dcont}
\end{eqnarray}
In the equation above, we have dropped the superscripts since they are not relevant to the discussion.

\begin{remark}
We may interpret the FNO as a sequence of an affine transform and convolution followed by a point-wise nonlinear activation. This combination of linear and nonlinear (activation) operations allows us to approximate nonlinear operator using this architecture. 
\end{remark}

\begin{remark}
It is instructive to list all the trainable entities in a FNO. First we list all the trainable parameters:
\begin{equation}
W_{i}^{(1)}, W_{ij}^{(2)}, \cdots, W_{ij}^{(L)}, W_{i}^{(L+1)}; b_{i}^{(1)}, b_{i}^{(2)}, \cdots, b_{i}^{(L)}, b^{(L+1)}. 
\end{equation}
Thereafter, all the trainable kernel functions 
\begin{equation}
\kappa_{ij}^{(n)}(\bm{z}), \qquad n = 2, \cdots, L. 
\end{equation}

\end{remark}

The neural operator introduced in this section acts directly on functions and transforms them into functions. However, when implementing this operator on a computer the functions have to be represented discretely. This is described in the following section. 

\section{Discretization of the Fourier Neural Operator}
The functions that appear in the neural operator described in the previous section are:
\begin{equation}
    a, \bm{v}^{(1)}, \bm{u}^{(1)}, \cdots, \bm{v}^{(L)}, \bm{u}^{(L)}, v^{(L+1)}, u. 
\end{equation}
Each of these functions is defined on the domain $\Omega$. We discretize this domain with $N$ uniformly distributed points, and represent each function using its values on these points. 

As an example, in two dimensions, with $\Omega = [0,L_1] \times [0,L_2]$, we represent the function $a(x_1,x_2)$ as,
\begin{equation}
    a[m,n] = a(x_{1m}, x_{2n}),\quad m = 1\cdots, N_1,\, n = 1\cdots, N_2.
\end{equation}
where 
\begin{eqnarray}
    x_{1m} &=& (m-1) \times \frac{L_1}{N_1-1} \\
    x_{1n} &=& (n-1) \times \frac{L_2}{N_2-1}.
\end{eqnarray}
The same representation will be used for all other functions. 

We now have to consider the discrete version of all operations on these functions as well. This is described below for the special case of $\Omega = [0,L_1] \times [0,L_2]$. 

We begin with the operator $A^{(1)}$. The discretized version is
\begin{equation}
\bm{v}^{(1)}[m,n] = A^{(1)}(a)[m,n],
\end{equation}
where 
\begin{equation}
v^{(1)}_i [m,n] = W_{i}^{(1)}a[m,n] + b_{i}^{(1)}, \qquad i = 1,\cdots, H.
\end{equation}
Similarly, the discretized version of the operator $A^{(L+1)}$ is,
\begin{equation}
v^{(L+1)} [m,n] = A^{(L+1)}(\bm{u}^{(L)})[m,n],
\end{equation}
where 
\begin{equation}
v^{(L+1)} [m,n] = W_{i}^{(L+1)}u_i^{(L)}[m,n] + b^{(L+1)}, \qquad i = 1,\cdots, H.
\end{equation}
Next we describe the action of the activation function on discretized input functions. It is given by 
\begin{equation}
\bm{u}^{(n)} [m,n] = \sigma (\bm{v}^{(n)})[m,n],
\end{equation}
where 
\begin{equation}
u^{(n)}_i [m,n] = \sigma (v_i^{(n)}[m,n]), \qquad i = 1,\cdots, H.
\end{equation}
Finally it remains to develop the discrete version of the operators $A^{(n)}, n = 2,\cdots, L$. These are defined as, 
\begin{equation}
\bm{v}^{(p+1)} [m,n] = A^{(p+1)}(\bm{u}^{(p)})[m,n],
\end{equation}
where 
\begin{eqnarray}
v^{(p+1)}_i [m,n] &=& W_{ij}^{(p+1)} u_j^{(p)}[m,n] + b_i^{(p+1)} \\
& & + \sum_{r = 1}^{N_1} \sum_{s = 1}^{N_2}  \kappa_{ij}^{(p+1)}[r-m,s-n] u_j^{(p)}[r,s] h_1 h_2, \qquad i = 1,\cdots, H,
\end{eqnarray}
where $h_1 = \frac{L_1}{N_1-1}$ and $h_2 = \frac{L_2}{N_2-1}$. Note that the integral in the convolution is now replaced by a sum over all the grid points. Computing this integral for each value of $i$ and $m,n$ involves $O(N_1 N_2 H)$ flops. And since this needs to be done for $H$ different values of $i$, $N_1$ values of $M$, and $N_2$ values of $j$, the total cost of discretizing the convolution operation is $O(N_1^2 N_2^2 H) = O(N^2 H^2)$, where $N = N_1 \times N_2$. The factor of $N^2$ in this cost is not acceptable and makes the implementation of this algorithm impractical. In the following section we describe how the use of Fourier Transforms (forward and inverse) overcomes this bottleneck and leads to a practical algorithm. This is also the reason that this algorithm is referred to as a ``Fourier Neural Operator."

\section{The Use of Fourier Transforms}
Consider a periodic function $u(x_2,x_2)$ define on $\Omega \equiv  [0,L_1] \times [0,L_2]$. If this function is sufficiently smooth it may be approximated by a truncated Fourier series,
\begin{equation}
    u(x_1,x_2) \approx \sum_{m = -N_1/2}^{N_1/2} \sum_{n = -N_2/2}^{N_2/2} \hat{u}[m,n] e^{2 \pi {\rm i}(\frac{mx_1}{L_1} + \frac{n x_2}{L_2}) }.  \label{eq:ift}
\end{equation}
Here $N_1$ and $N_2$ are even integers, the coefficients $\hat{u}[m,n]$ are the Fourier coefficients and ${\rm i} = \sqrt{-1}$. We note that while the function $u$ is real-valued the coefficients are complex-valued. However, since $u$ is real-valued, they obey the rule $\hat{u}[-m,-n] = \hat{u}^*[m,n]$, where $(.)^*$ denotes the complex-conjugate of a complex number. The approximation can be made more accurate by increasing $N_1$ and $N_2$, and as these numbers tend to infinity, we recover the equality. The relation above is often referred to as the inverse Fourier transform, since it maps the Fourier coefficients to the function in the physical space. 

The forward Fourier transform (which maps the function in the physical space to the Fourier coefficients) can be obtained from the relation above by
\begin{enumerate}
    \item Multiplying both sides by $e^{- 2 \pi {\rm i}(\frac{ r x_1}{L_1} + \frac{s x_2}{L_2})}$, where $r$ and $s$ are integers. 
    \item Integrating both sides over $\Omega$. 
    \item Recognizing that the integral $\int_\Omega e^{2 \pi {\rm i}(\frac{ (m-r) x_1}{L_1} + \frac{ (n-s) x_2}{L_2})} dx_1 dx_2$ is non-zero only when $m = r$ and $n = s$, and in that case it evaluates to $L_1 L_2$. 
\end{enumerate}
These steps yield the final relation:
\begin{equation}
    \hat{u}[r,s] = \frac{1}{L_1 L_2 } \int_0^{L_1}  \int_0^{L_2} u(x_1,x_2) 
 e^{- 2 \pi {\rm i}(\frac{ r x_1}{L_1} + \frac{ s x_2}{L_2})} dx_1 dx_2. \label{eq:ft}
\end{equation}

We now describe how Fourier transforms can be used to evaluate the convolution efficiently. To do this we consider the special case of 2D convolution in (\ref{eq:conv2dcont}). We begin with substituting $u_j(y_1,y_2) = \sum_{m = -N_1/2}^{N_1/2} \sum_{n = -N_2/2}^{N_2/2} \hat{u}_j [m,n] e^{2 \pi {\rm i}(\frac{m y_1}{L_1} + \frac{n y_2}{L_2}) }$ in this equation to get,
\begin{eqnarray}
v_i  (x_1,x_2) &=&  \int_{0}^{L_1} \int_{0}^{L_2} \kappa_{ij}(y_1-x_1,y_2-x_2) \sum_{m,n}  \hat{u}_j [m,n]  e^{2 \pi {\rm i}(\frac{m y_1}{L_1} + \frac{n y_2}{L_2}) }  \; dy_2 dy_1 \nonumber \\
&=& \sum_{m,n}  \hat{u}_j [m,n]   \int_{0}^{L_1} \int_{0}^{L_2} \kappa_{ij}(y_1-x_1,y_2-x_2)  e^{2 \pi {\rm i}(\frac{m y_1}{L_1} + \frac{n y_2}{L_2}) }  \; dy_2 dy_1 \nonumber \\
&=& \sum_{m,n}  \hat{u}_j [m,n]   \int_{-x_1}^{L_1-x_1} \int_{-x_2}^{L_2-x_2} \kappa_{ij}(z_1,z_2)  e^{2 \pi {\rm i}(\frac{m (z_1 + x_1)}{L_1} + \frac{n (z_2 + x_2) }{L_2}) }  \; dz_2 dz_1 \nonumber \\
&=& \sum_{m,n}  \hat{u}_j [m,n] e^{2 \pi {\rm i}(\frac{m  x_1}{L_1} + \frac{n x_2}{L_2}) }   \int_{0}^{L_1} \int_{0}^{L_2} \kappa_{ij}(z_1,z_2)  e^{2 \pi {\rm i}(\frac{m z_1}{L_1} + \frac{n z_2}{L_2}) } \; dz_2 dz_1 \nonumber \\
&=& L_1 L_2 \sum_{m,n}  \hat{u}_j [m,n] \hat{\kappa}_{ij} [-m,-n] e^{2 \pi {\rm i}(\frac{m  x_1}{L_1} + \frac{n x_2}{L_2}) }. 
\label{eq:conv2dft}
\end{eqnarray}
In the development above, in going from the first to the second line we have taken the summation outside the integral and recognized that the coefficients $\hat{u}_j [m,n] $ do not depend on $y_1$ and $y_2$. In going from the second to the third line we have introduced the variables $z_1 = y_1 - x_1$ and $z_2 = y_2 - x_2$. In going from the third to the fourth line we have made use of the fact that the functions $\kappa_{ij}(z_1,z_2)$ are periodic. Finally in going from the fourth to the fifth line we have made use of the definition of the Fourier Transform (\ref{eq:ft}). This final relation tells us that the convolution can be computed by:
\begin{enumerate}
    \item Computing the Fourier Transform of $u_j$.
    \item Computing the Fourier Transform of $\kappa_{ij}$.
    \item Computing the product of the coefficients of these two transforms. 
    \item Computing the inverse Fourier Transform of the product. 
\end{enumerate}

Next, we account for the fact that we will only work with the discrete forms of the functions $u_j$ and $\kappa_{ij}$. This means that we evaluate the inverse Fourier transform (\ref{eq:ift}) at a finite set of grid points. Further, it means that we have to approximate the integral in the Fourier transform (\ref{eq:ft}). This alternate form is given by 
\begin{equation}
    \hat{u}[r,s] = \frac{h_1 h_2}{L_1 L_2 } \sum_{m = 1}^{N_1} \sum_{n = 1}^{N_2}  u[m,n] 
 e^{- 2 \pi {\rm i}(\frac{ r x_{1m}}{L_1} + \frac{ s x_{2n}}{L_2})} . \label{eq:dft}
\end{equation}
Here $h_1 = \frac{L_1}{N_1}$ and $h_2 = \frac{L_2}{N_2}$, $x_{1m} = (m-1)h_1$ and $x_{2n} = (n-1)h_2$. 

The final observation is that the evaluating the sums in (\ref{eq:ift}) and (\ref{eq:dft}) require $O(N^2)$ operations. This would make the evaluation of the convolution via the Fourier method impractical except for when $N$ is very small. However, the use of Fast Fourier Transform (FFT) reduces this cost to $O(N \log N)$. Thus the cost of implementing the convolution reduces to $O(N \log N H^2)$. This makes the implementation of Fourier Neural Operators practical.

%% file: probML_GANS.tex

\chapter{Probabilistic Deep Learning}

So far, we have considered regression and classification problems, where for a given input $\x$ we need to compute a \textit{single} output $\y$. However, this may not be enough for many problems of interest. In fact, there may be \textit{many} $\y$'s for a given $\x$. For example,
\begin{enumerate}
\item $\y$ and $\x$ might be measured with some \textit{random noise}.
\item $\y$ and $\x$ might be \textit{inherently stochastic}. For instance, $\y$ could be the pressure measured in a turbulent flow at some point $\x$ in space.
\item \textit{inverse problems} can have multiple solutions. For instance, the forward/direct problems would be determining the temperature field given the head conductivity, while the inverse problem could be determining the conductivity field given the (possibly noisy) temperature field.
\end{enumerate}

Thus, we need to formulate a probabilistic framework to use deep learning algorithms to solve such problems. Recall, that our deterministic model was given by $\y = \bm{\mathcal{F}}(\x;\btheta)$. In the probabilistic setup, $\y$, $\x$ and $\btheta$ are treated as \textit{random variables}.

Before we can work with random variables we need to understand some key elements of the theory of probability that are necessary in defining random variables.

\section{Key elements of Probability Theory}
A \textbf{random experiment} is described by a procedure and a set of one or more observations/measurements. For example,
\begin{enumerate}
\item Observe a switch and determine whether it is \textit{on} or \textit{off}.
\item Toss a coin 3 times and note the sequence of heads $H$ or tails $T$.
\item Toss a coin 3 times and count the number of times $H$ appears. Note that this is the same experiment as earlier but the measurement is different.
\item Spin a spinner, and measure the final angle in radians.
\end{enumerate}

The \textbf{outcome} is the results of the experiment that cannot be broken down into smaller parts. The \textbf{sample space}, denoted by $\mathbb{S}$, is the set of all possible outcomes of an experiment. For each of the four random experiments observed above, we have
\begin{enumerate}
\item $\mathbb{S} = \{ \text{on}, \ \text{off} \}$.
\item $\mathbb{S} = \{ TTT, \ TTH, \  THT, \ HTT, \ THH, \ HTH, \ HHT, \ HHH \}$.
\item $\mathbb{S} = \{ 0, \ 1, \ 2, \ 3 \}$.
\item $\mathbb{S} = \{ \psi \ : \ \psi \in (0,2\pi] \}$.
\end{enumerate}
Note that there is a fundamental difference between the first 3 experiments, where $\mathbb{S}$ is discrete and countable, and the last experiment where the $\mathbb{S}$ is uncountable. 

An \textbf{event} is a collection of outcomes, i.e., a subset of $\mathbb{S}$. Typically the outcomes in an event satisfy a condition. Let's see some examples for the above experiments
\begin{enumerate}
\item $A = \{ \text{on} \}$ or  $A = \{ \text{on}, \ \text{off} \} = \mathbb{S}$ .
\item $A$ are all outcomes with at least 2 $H$, i.e., $A = \{THH, \ HTH, \ HHT, \ HHH \}$.
\item $A$ are all outcomes with at least 2 $H$, i.e., $A= \{  2, \ 3 \}$. If we define $B$ to be all outcomes with 4 $H$, then no outcome would satisfy this condition, i.e., $B=\emptyset$ the null set. 
\item $A$ are all outcomes with $\psi > \pi/4$, i.e, $A = \{ \psi \ : \ \psi \in (\pi/4,2\pi] \}$.
\end{enumerate}

An \textbf{event class} $\mathscr{E}$ is a collection of all event (sets) over which probabilities can be defined. When $\mathbb{S}$ is countable, $\mathscr{E}$ is all subsets of $\mathbb{S}$. When $\mathbb{S}$ is not countable, $\mathscr{E}$ is the \textit{Borel} field (or Borel algebra), which is the collection of all open and closed sets in $\mathbb{S}$. 

The \textbf{probability law} is a rule that assigns a probability to all sets in an event class $\mathscr{E}$. 
We list the \textbf{axioms of probability}, which are the requirements of a probability law.

Consider a sample space $\mathbb{S}$ for an experiment and the corresponding event class $\mathscr{E}$. Let  $P: \mathscr{E} \mapsto [0,1]$ satisfy
\begin{enumerate}
\item $P[A] \geq 0$ for all $A \in  \mathscr{E}$.
\item $P[\mathbb{S}] = 1$.
\item If $A_1,A_2,...$ are events such that $A_i \cap A_j = \emptyset$ for all $i \neq j$, i.e., the events are \textit{mutually exclusive}, then 
\[
P[\bigcup\limits_{i=1}^{\infty} A_i] = \sum_{i=1}^\infty P[A_i]. 
\]
\end{enumerate}
Any assignment $P$ that satisfies the above conditions is said to be a \textit{valid probability law}. Note that probability is like mass. It is non-negative (axiom 1), conserved (total mass is always 1, axiom 2), and for distinct points the total mass is obtained by adding individual masses (axiom 3).

If $\mathbb{S}$ is countable, then it is sufficient to define a probability law for all elements of $\mathbb{S}$, i.e., for all elementary outcomes, while making sure that the probabilities are non-negative and add up to 1 (the first two axioms).
Let us try to assign probability laws for the first three examples which have a countable $\mathbb{S}$ using these criteria. 
\begin{enumerate}
\item For some $p \in [0,1]$, define $P[\text{on}] = p$;  $P[\text{off}] = 1-p$.
\item For a fair die with no memory, $P[a_i] = 1/8$, where $a_i \in \mathbb{S}, i = 1, \cdots, 8$. 
\item For a fair die with no memory, $P[0] = 1/8, \ P[1] = 3/8, \ P[2] = 3/8, \ P[3] = 1/8$. 
\end{enumerate}

\begin{remark}
As an exercise, verify that the axioms are satisfied for each of these cases.    
\end{remark}

For a continuous sample space, it is sufficient to define a probability law for all open and closed intervals, while ensuring axioms 1 and 2. Let us consider the fourth example which has an uncountable $\mathbb{S}$. If the spinner is completely unbiased, then the probability is uniformly distributed. Then for $b \geq a$, we say that $P[(a,b]] = (b-a)/(2 \pi)$. Note that the probability of singleton sets in a continuous sample space is zero (for any distribution). 

\begin{remark}
From this point on, whenever we talk about the sample space $\mathbb{S}$, we will implicitly assume that we are referring to the triplet $(\mathbb{S}, \mathscr{E}, P)$. This triplet is also known as a \text{"measure space"}.
\end{remark}

\section{Random Variables} 

A \textbf{random variable} $X$ is a function defined from $\mathbb{S}$ to the real line with the property that the set $A_b = \{ \xi \in \mathbb{S} :  X(\xi) \leq b \}$ belongs to $\mathscr{E}$ for all $b \in \Ro$. Note that in the measure theoretic language, we are requiring $X$ to be a \textit{measurable function}. Also note that according to the definition of $A_b$, we are enforcing the requirement that we should be able to evaluate $P[A_b]$ for all $b \in \Ro$.

Let us define random variables (RVs) for the above examples:
\begin{enumerate}
\item For $\mathbb{S} = \{ \text{on}, \text{off} \}$ with $P[\text{on}] = p$,  $P[\text{off}] = 1-p$, define the RV
\begin{equation}\label{eqn:rv_switch}
X(\xi) = \begin{cases}
0 & \quad \text{if } \xi = \text{off}\\
1 & \quad \text{if } \xi = \text{on}.
\end{cases}
\end{equation}
This is also known as a \textit{Bernoulli Random Variable}. 

\item For $\mathbb{S} = \{ TTT, \ TTH, \  THT, \ HTT, \ THH, \ HTH, \ HHT, \ HHH \}$ with $P[a_i] = 1/8$ for all $a_i \in \mathbb{S}$, define
\begin{equation}\label{eqn:rv_heads}
X(\xi) = \text{Number of heads in } \xi.
\end{equation}
Note that this is the random event that was described in Experiment 3. 

\item This random event is already a random variable. 

\item For the spinner experiment with $\mathbb{S} = \{ \psi \ : \ \psi \in (0,2\pi] \}$ with $P[(a,b]] = (b-a)/(2 \pi)$, define
\begin{equation}\label{eqn:rv_spinner}
X(\psi) = \frac{\psi}{2 \pi}.
\end{equation}
\end{enumerate}
If $X$ is defined on a discrete sample space, it is called a \textit{discrete random variable}, while if it is defined on a continuous sample space, it is called a \textit{continuous random variable}.

As described above, a random variable inherits its probabilistic interpretation from the measure space used to define it. In the following sections we define the probabilistic interpretation of a random variable. 

\subsection{Cumulative distribution function}
\begin{figure}[htbp]
\begin{center}
\subfigure[Bernoulli]{\includegraphics[width=0.3\textwidth]{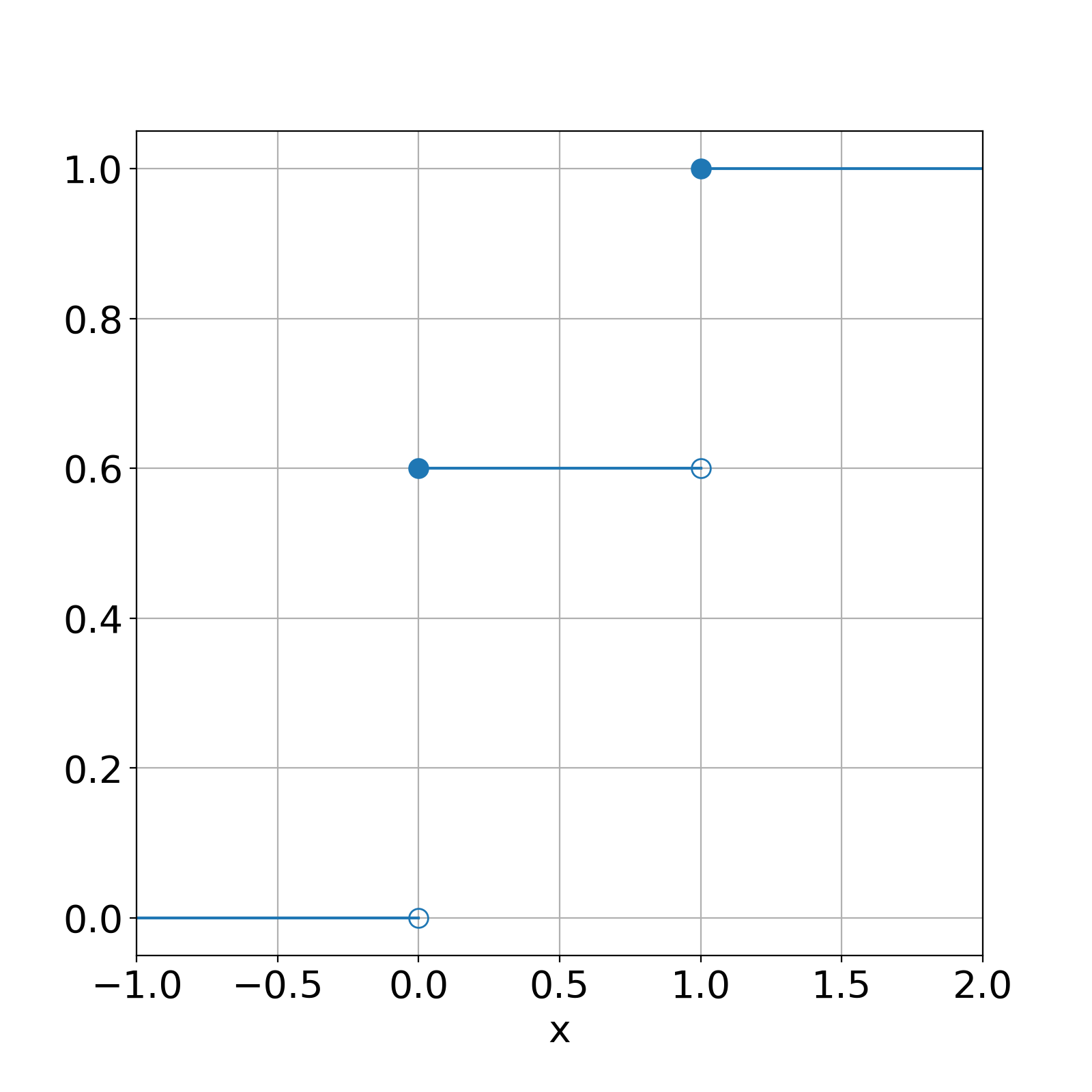}}
\subfigure[3 coin tosses]{\includegraphics[width=0.3\textwidth]{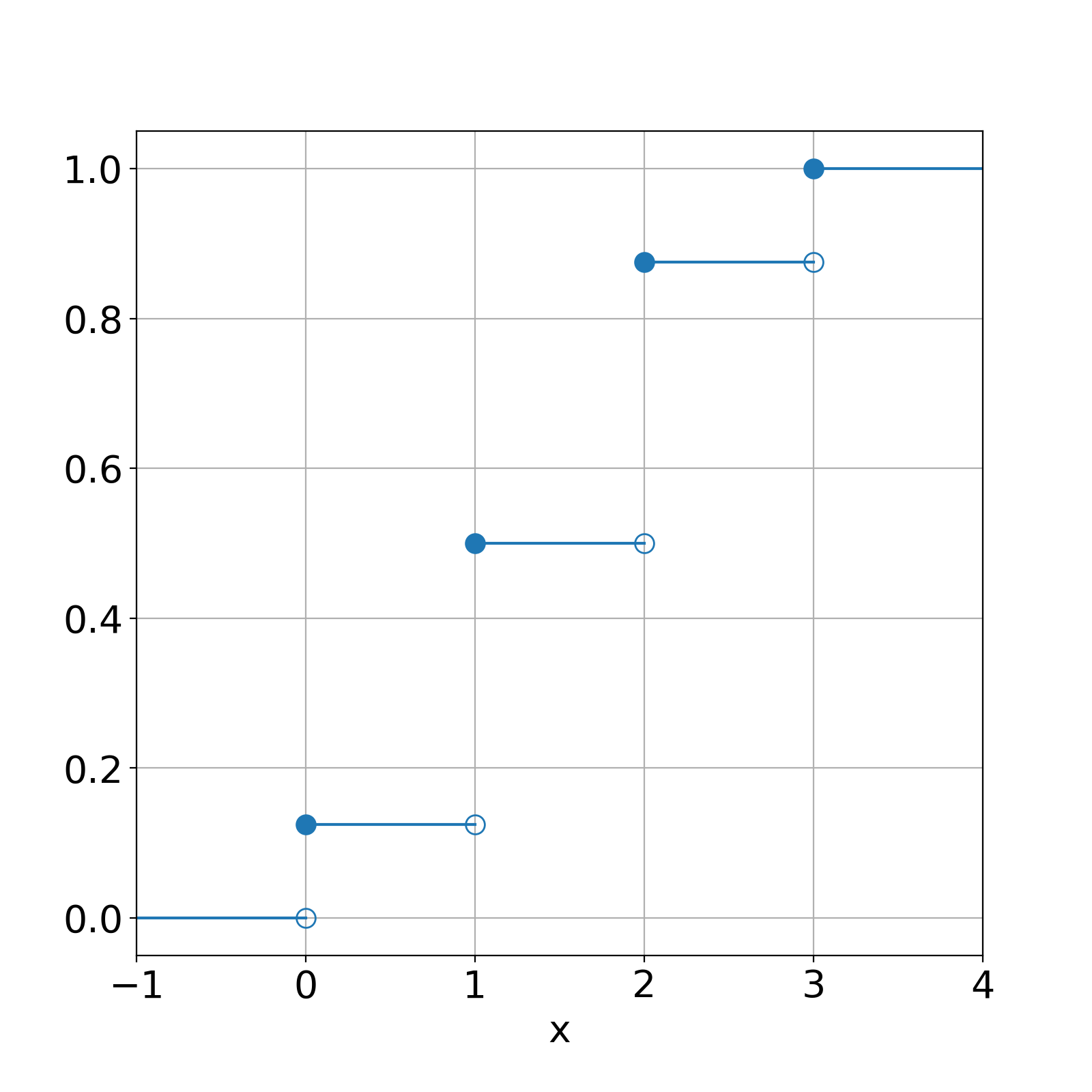}}
\subfigure[Spinner]{\includegraphics[width=0.3\textwidth]{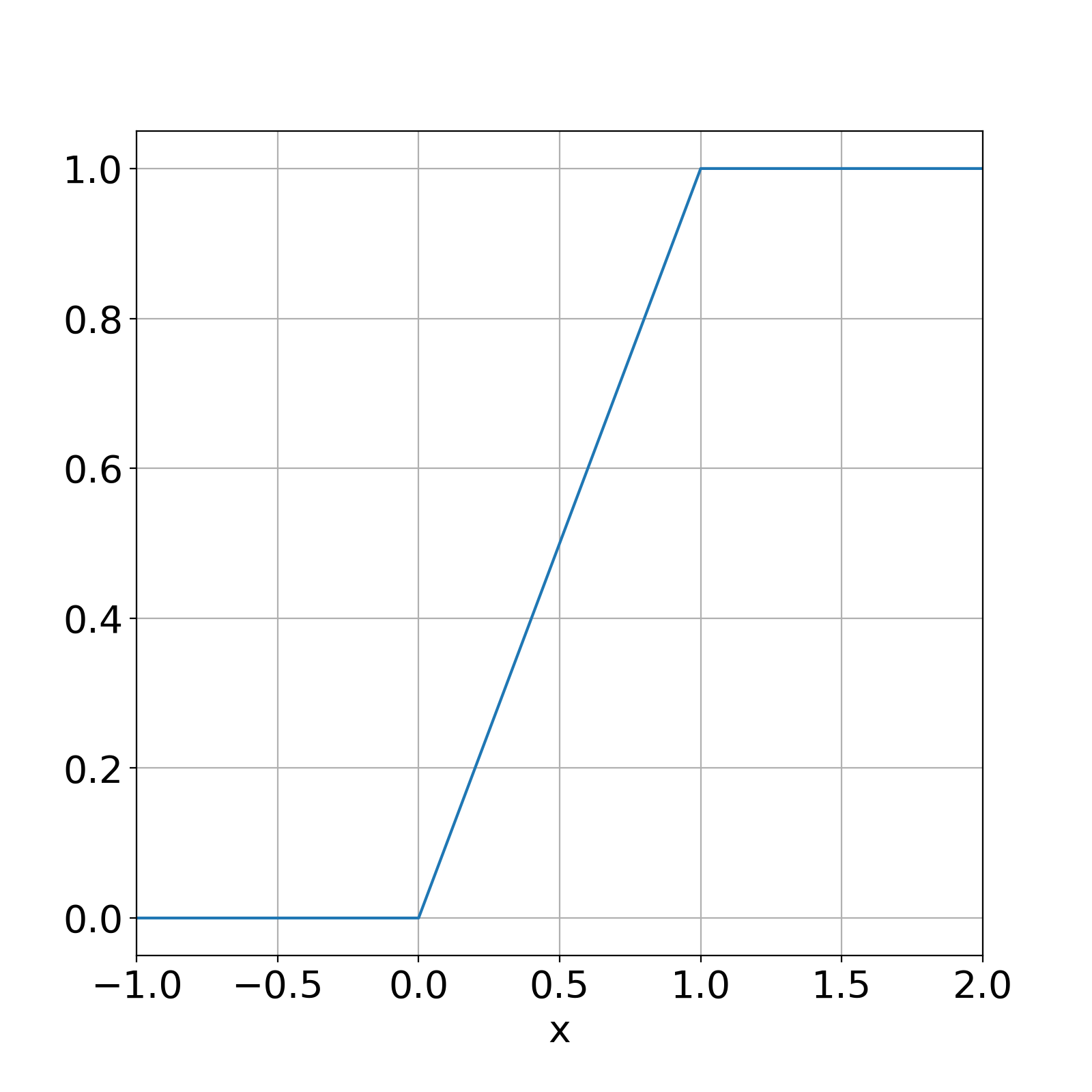}}
\caption{Examples of cumulative distribution functions}
\label{fig:cdf}
\end{center}
\end{figure}

The \textbf{cumulative distribution function} (cdf) of a random variable $X$ is given by
\[
F_X(x) = P[\xi : X(\xi) \leq x]  
\]
which defines a probability on $\Ro$ of $X$ taking values in the interval $(-\infty,x]$. Let us define the cdf for the above examples:
\begin{enumerate}
\item For the Bernoulli RV defined by \eqref{eqn:rv_switch}
\begin{itemize}
\item if $x < 0$, then $F_X(x) = P[\emptyset] = 0$
\item if $0 \leq x < 1$, then $F_X(x) = P[\{\text{off}\}] = 1-p$
\item if $x \geq 1$, then $F_X(x) = P[\{\text{on},\text{off}\}] = 1$
\end{itemize}
The full cdf is shown in Figure \ref{fig:cdf}(a). 

\item For the RV defined by \eqref{eqn:rv_heads}
\begin{itemize}
\item if $x < 0$, then $F_X(x) = P[\emptyset] = 0$
\item if $0 \leq x < 1$, then $F_X(x) = P[\# of H = 0] = P[\{TTT\}] = 1/8$
\item if $1 \leq x < 2$, then $F_X(x) = P[\# of H = 0,1]  = 1/8 + 3/8 = 4/8$
\item if $3 \leq x < 3$, then $F_X(x) = P[\# of H = 0,1,2] = 1/8 + 3/8 + 3/8 = 7/8$
\item if $x \geq 3$, the $F_X(x) = P[\# of H = 0,1,2,3] =  1$
\end{itemize}
The full cdf is shown in Figure \ref{fig:cdf}(b). 

\item This random variable is the same as Example 2. 

\item For the spinner experiment with the RV defined by \eqref{eqn:rv_spinner}
\[
F_X(x) = P[\psi : X(\psi) \leq x] = P[\{\psi: \psi \leq 2 \pi x\}]  
\]
\begin{itemize}
\item if $x < 0$, then $F_X(x) = P[\emptyset] = 0$
\item if $0 \leq x < 1$, then $F_X(x) = P[\{\psi \in (0,2 \pi x]\}] = \frac{2 \pi x}{2 \pi} = x$
\item if $x \geq 1$, then $F_X(x) = P[\{\psi \leq 2 \pi\}] = 1$
\end{itemize}
The full cdf is shown in Figure \ref{fig:cdf}(c).

\end{enumerate}

Let us discuss some properties of $F_X$:
\begin{enumerate}
\item $0 \leq F_X(x) \leq 1$.
\item $\lim_{x \rightarrow \infty} F_X(x) = 1$.
\item $\lim_{x \rightarrow -\infty} F_X(x) = 0$.
\item $F_X$ is monotonically increasing.
\item The cdf is always continuous from the right
\[
F_X(x) = \lim_{h \rightarrow 0+} F_x(x+h).
\]
Note that the $F_X$ for discrete RV (see Figure \ref{fig:cdf}) are discontinuous at finitely many $x$. In fact, the cdf for discrete RVs can be written as a finite sum of the form
\[
F_X(x) = \sum_{k=1}^K p_k H(x-x_k), \quad \sum_{k=1}^K p_k = 1,
\]
where $p_k$ is the probability mass and $H$ is the Heaviside function
\[
H(x) = \begin{cases} 1 & \quad \text{if } x >0\\0 & \quad \text{if } x \leq 0 \end{cases}.
\]
\end{enumerate}

\begin{remark}
Once we have the $F_X$ we can calculate the probability that $X$ will take values in "any" interval in $\Ro$, i.e., we can compute $P[a < X \leq b]$. Note that
\begin{align*}
F_X(b) = P[X \leq b] &= P[(X \leq a) \cup (a < X \leq b)]  \\
&= P[X \leq a] + P[a < X \leq b] \qquad \mbox{(mutually exclusive events)} \\
&= F_X(a) + P[a < X \leq b].
\end{align*}
Thus,
\[
P[a < X \leq b] = F_X(b) - F_X(a).
\]    
\end{remark}

\subsection{Probability density function}
We define the \textbf{probability density function} (pdf). For a continuous $F_X$, it is defined as
\begin{equation}\label{eqn:pdf}
f_X(x) = \dd{}{x} F_X(x)
\end{equation}
which enjoys the following properties inherited from the cdf:
\begin{enumerate}

\item $f_X(x) \geq 0, \ \forall x \in \Ro$, since $F_X$ is monotonically increasing.

\item $\lim_{x\rightarrow -\infty}f_X(x) = \lim_{x\rightarrow \infty}f_X(x)  =  0$.

\item Integrating \eqref{eqn:pdf} from $(-\infty,x]$ gives us
\[
\int_{-\infty}^x f_X(y) \ud y = F_X(x) - \lim_{x\rightarrow -\infty}F_X(x) = F_X(x).
\]

\item Also
\[
P[a < X \leq b] = F_X(b) - F_X(a) = \int_{-\infty}^b f_X(y) \ud y - \int_{-\infty}^a f_X(y) \ud y = \int_{a}^b f_X(y) \ud y.
\]
Thus, the integral of a pdf in an interval gives the "probability mass" which is the probability that the RV lies in that interval. This is the reason why the pdf  is called a "density".

\item Furthermore,
\[
\int_{-\infty}^\infty f_X(y) \ud y =\lim_{x\rightarrow \infty}F_X(x) - \lim_{x\rightarrow -\infty}F_X(x)= 1.
\]

\item For a very small $h > 0$, we have the interpretation
\[
P[a < X \leq a+h] = \int_{a}^{a+h} f_X(y) \ud y \approx h f_X(a) .  
\]
Note that as $h \rightarrow 0^+$, $P[a < X \leq a+h]  \rightarrow 0$. That is, for a continuous RV the probability of attaining a single value is zero.

\end{enumerate}

\subsection{Examples of Important RVs}

\begin{figure}[htbp]
\begin{center}
\subfigure[Uniform RV ($a=-1,b=1$)]{\includegraphics[width=0.3\textwidth]{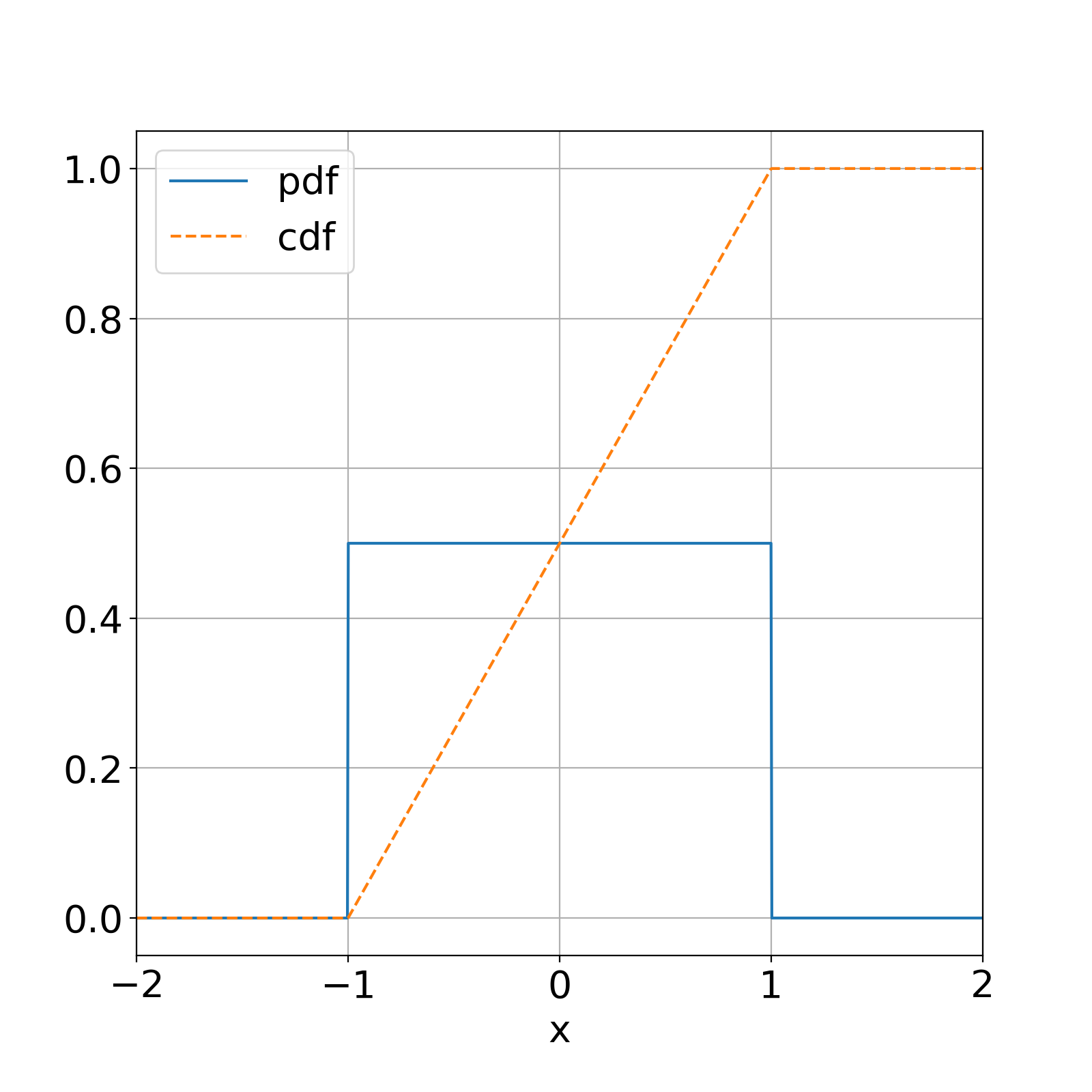}}
\subfigure[Exponential RV ($\lambda=0.8$)]{\includegraphics[width=0.3\textwidth]{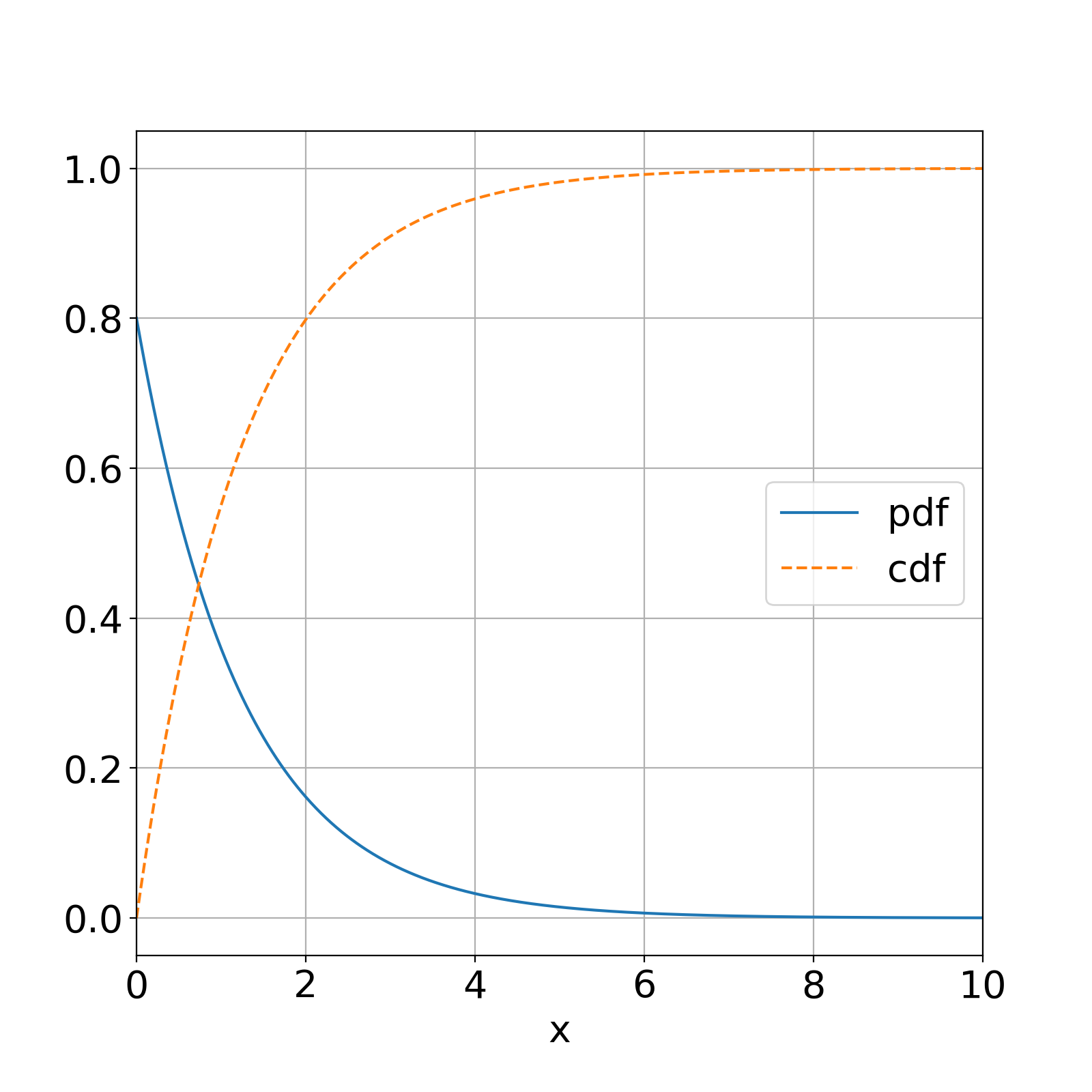}}
\subfigure[Gaussian RV]{\includegraphics[width=0.3\textwidth]{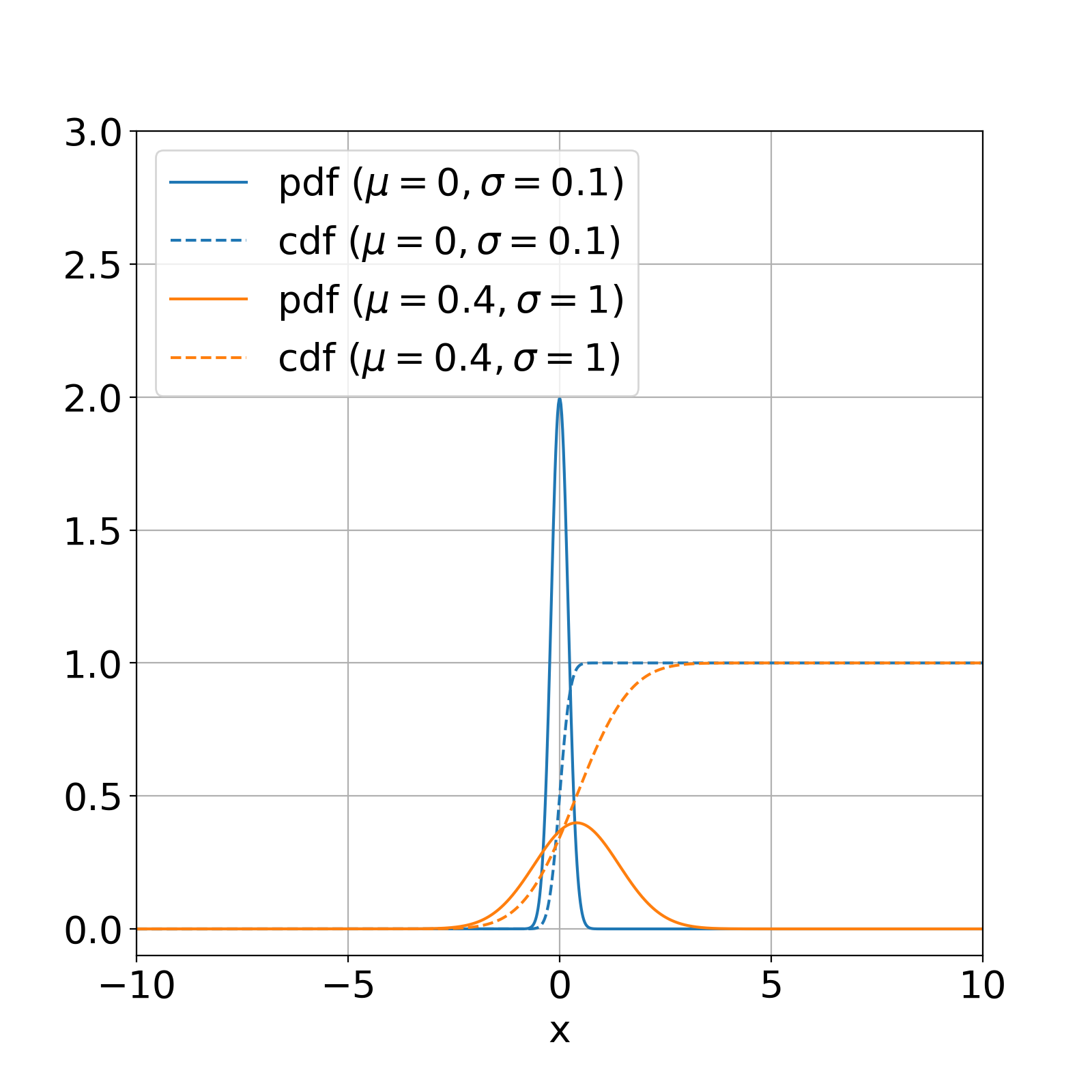}}
\caption{Continuous random variables}
\label{fig:cont_rv}
\end{center}
\end{figure}

Let us look at some important random variables and the associated cdf, pdf (also see Figure \ref{fig:cont_rv}):
\begin{enumerate}
\item \textbf{Uniform RV:} for some interval $(a,b]$, the pdf is given by 
\[
f_X(x) = \begin{cases} \frac{1}{b-a} & \quad \text{if } x \in(a,b] \\ 0 & \quad \text{other wise}\end{cases},
\]
while the cdf is given by
\[
F_X(x) = \begin{cases} 0 & \text{if } x < a\\ \frac{x-a}{b-a} & \quad \text{if } x \in(a,b] \\ 1 & \quad  \text{if } x > b\end{cases}.
\]

\item \textbf{Exponential RV:} used to model lifetime of devices/humans after a critical event. In this case, $X$ represents the time to failure and $P[X > x] = e^{-\lambda x}$ where $\lambda >0$ is a model parameter which denotes the rate of failure. Thus,
\[
F_X(x) = P[X \leq x] = 1 - P[X > x] = 1 - e^{-\lambda x},
\]
and
\[
f_X(x) = \dd{}{x}F_X(x) = \lambda e^{-\lambda x}.
\]

\item \textbf{Gaussian RV:} used to model natural things like height, weight, etc. In fact, through the Central Limit Theorem, this is also the distribution given by an aggregate of many RVs. The pdf is given by
\[
f_X(x) = \frac{1}{\sqrt{2 \pi} \sigma } e^{-\frac{1}{2}\left(\frac{x-\mu}{\sigma}\right)^2}
\]
which is parameterized by the \textit{mean} $\mu$ which denotes the center of this distributions, and the \textit{variance} $\sigma^2$ which denotes its spread. The corresponding cdf is given by
\[
F_X(x) = \frac{1}{2} \left[ 1 + \text{erf}\left( \frac{x-\mu}{\sigma \sqrt{2}}\right) \right], \quad \text{erf}(x) = \frac{2}{\sqrt{\pi}} \int_{0}^x e^{-t^2} \ud t.
\]
\end{enumerate}

In probabilistic Machine Learning one makes extensive use of uniform and Gaussian random variables.

\subsection{Expectation and variance of RVs}
Given a RV $X$ with pdf $f_X$, we can calculate its \textbf{expected value} or \textbf{expectation} or \textbf{mean} as
\[
\mu_X:=\mathbb{E}[X] = \int_{-\infty}^\infty x f_X(x) \ud x.
\]
The expectation has the following properties:
\begin{itemize}
\item Note that if a pdf is symmetric about $x=m$, then $\mathbb{E}[X] = m$. To see this, note that $(m-x)f_X(x)$ will be anti-symmetric about $m$. Thus
\[
0 = \int_{-\infty}^\infty (m-x) f_X(x) \ud x = m \int_{-\infty}^\infty f_X(x) \ud x - \int_{-\infty}^\infty x f_X(x) \ud x  \ \ \implies \ \ \int_{-\infty}^\infty x f_X(x) \ud x = m.
\]
Using this property, we can easily say the mean for a uniform RV is $(a+b)/2$, while for a Gaussian RV it is $\mu$.

\item $\mathbb{E}[c] = c$ for a constant $c$.

\item We can calculate the expected value of functions of RVs as
\[
\mathbb{E}[g(X)] =  \int_{-\infty}^\infty g(x) f_X(x) \ud x.
\]

\item The expectation is linear, i.e.,
\[
\mathbb{E}[g(X) + c h(X)] = \mathbb{E}[g(X)] + c \mathbb{E}[h(X)].
\]
\end{itemize}

The \textbf{variance} of a RV measures its variation about the mean. It is evaluated as
\[
\text{VAR}[X] = \int_{-\infty}^\infty (x-\mu_X)^2 f_X(x) \ud x.
\]
Furthermore, we denote the \textbf{standard deviation} as
\[
\sigma_X := \text{STD}[X] = \sqrt{\text{VAR}[X] }.
\]
For a uniform RV
\[
\text{VAR}[X] = \int_{a}^b\left(x - \frac{b+a}{2} \right)^2 \frac{1}{b-a} \ud x = \frac{(b-a)^2}{12}.
\]
For a Gaussian RV, we first use the property of the pdf to write
\[
 \int_{-\infty}^{\infty}  e^{-\frac{1}{2}\left(\frac{x-\mu}{\sigma}\right)^2} \ud x = \sqrt{2 \pi} \sigma.
\]
Taking a derivative with respect to $\sigma$ on both sides lead to
\[
\int_{-\infty}^{\infty}  e^{-\frac{1}{2}\left(\frac{x-\mu}{\sigma}\right)^2} (x-\mu)^2\sigma^{-3}\ud x = \sqrt{2 \pi}
\]
which after a bit of algebra gives us
\[
\text{VAR}[X] = \int_{-\infty}^{\infty}  (x-\mu)^2 \frac{1}{\sqrt{2 \pi} \sigma } e^{-\frac{1}{2}\left(\frac{x-\mu}{\sigma}\right)^2} \ud x = \sigma^2.
\]

\subsection{Pair of RVs}

In probabilistic ML we deal with multiple random variables. For example, the input, the output and the weights might all be RVs. Thus we need to extend concepts from a single RV to a vector of RVs. We do this in this section by first considering a pair of RVs. Most of the concepts defined for a pair of RVs carry forward to a vector of RVs.

A pair of RVs is a mapping from the measure space, with event class $\mathscr{E}$, of the form 
\[
\bm{X} : \mathscr{E} \rightarrow \Ro^2,
\]
where the mapping can be discrete or continuous. We will sometimes use the notation $\bm{X} = (X,Y)$. For example, we can spin the spinner twice and measure $\psi_1 \in (0, 2 \pi]$, $\psi_2 \in (0, 2 \pi]$. In this case, we can define the two RVs
\[
X(\psi_1) = \frac{\psi_1}{2 \pi}, \quad Y(\psi_2) = \frac{\psi_2}{2 \pi}.
\]

Events for $\bm{X}$ are sets in $\Ro^2$. To compute probability of events, we need to define the \textbf{joint cdf} $F_{XY} : \Ro^2 \rightarrow \Ro$, where
\[
F_{XY}(x,y) = P[X \leq x, Y \leq y] = P[\xi \in \mathbb{S}: X(\xi) \leq x, Y(\xi) \leq y].
\]
Analogous to single RVs
\begin{itemize}
\item Joint cdfs are non-increasing functions of $x$, $y$. In other words, for $x \geq x^\prime$ and $y \geq y^\prime$
\[
F_{XY}(x,y) \geq F_{XY}(x^\prime,y^\prime).
\]

\item $\lim_{x \rightarrow -\infty} F(x,y) = 0$, $\lim_{y \rightarrow -\infty} F(x,y) = 0$, $\lim_{x,y \rightarrow \infty} F(x,y) = 1$.

\item We can calculate 
\[
P[x_1 < X \leq x_2, y_1 < Y \leq y_2] = F_{XY}(x_2,y_2) + F_{XY}(x_1,y_1) - F_{XY}(x_1,y_2) - F_{XY}(x_2,y_1).
\]
\end{itemize}

For $X,Y$ jointly continuous, we can define the \textbf{joint pdf} as
\[
f_{XY}(x,y)  = \frac{\partial^2F_{XY}(x,y)}{\partial x \partial y}
\]
which enjoys the following properties
\begin{itemize}
\item $f_{XY}(x,y) = 0$ as $x \rightarrow \pm \infty$ or $y \rightarrow \pm \infty$.
\item $F_{XY}(x,y) = \int_{-\infty}^x \int_{-\infty}^y f_{XY}(r,s) \ud r \ud s$.
\item $\int_{-\infty}^\infty \int_{-\infty}^\infty f_{XY}(r,s) \ud r \ud s = 1$.
\item $P[x_1 < X \leq x_2, y_1 < Y \leq y_2]  = \int_{x_1}^{x_2}\int_{y_1}^{y_2} f_{XY}(x,y) \ud x \ud y$.
\item $P[\bm{X} \in B]  = \int\int_{B} f_{XY}(x,y) \ud x \ud y$.
\end{itemize}

Let us look at some important joint random variables :
\begin{enumerate}
\item \textbf{Joint uniform RV:} for some region $(a,b] \times (c,d]$, the joint pdf is given by 
\begin{equation}\label{eqn:joint_uniform}
f_{XY}(x,y) = \begin{cases} \frac{1}{(b-a)(d-c)} & \quad \text{if } (x,y) \in(a,b] \times (c,d] \\ 0 & \quad \text{otherwise}\end{cases}.
\end{equation}

\item \textbf{Joint Gaussian RV:} the joint pdf is given by
\[
f_{XY}(x,y) = \frac{1}{\sqrt{(2 \pi)^2 \text{det}(\bm{\Sigma})} } \exp\left[-\frac{1}{2}(\x - \bm{\mu})^\top \bm{\Sigma}^{-1} (\x-\bm{\mu})\right]
\]
where $\x = (x,y)$, $\bm{\mu} = (\mu_x,\mu_y)$ is the mean, and $\bm{\Sigma}$ is called the covariance matrix 
\[
\bm{\Sigma} = \begin{bmatrix} \sigma_x^2 & \rho \sigma_x \sigma_y\\  \rho \sigma_x \sigma_y & \sigma_y^2\end{bmatrix}.
\]
The covariance matrix is symmetric and positive definite. 
\end{enumerate}

We can define the \textbf{marginal PDF} of the RV $X$, which is the pdf of $X$ assuming $Y$ attains all possible values
\[
f_X(x) = \int_{-\infty}^\infty f_{XY}(x,y) \ud y.
\]
Similarly, the marginal of $Y$ is 
\[
f_Y(y) = \int_{-\infty}^\infty f_{XY}(x,y) \ud x.
\]
The RVs $X$ and $Y$ are said to be \textbf{independent} if $f_{XY}(x,y) = f_X(x)f_Y(y)$.  

\begin{question}
Show that the joint uniform RVs with joint pdf \eqref{eqn:joint_uniform} are independent.
\end{question}

Consider the function $g(\bm{X})$, which can be scalar-, vector-, or tensor-valued, then its expected value is given by
\[
\mathbb{E}[g(\bm{X})] = \int_{-\infty}^\infty \int_{-\infty}^\infty  g(\x) f_{XY}(x,y) \ud x \ud y
\]
as long as the integral is defined.  For instance:
\begin{itemize}
\item For $g(\bm{X}) = X$, we have $\mathbb{E}[g(\bm{X})]  = \int_{-\infty}^\infty \int_{-\infty}^\infty  x f_{XY}(x,y) \ud x \ud y$.
\item For $g(\bm{X}) = \bm{X}$, we have a vector valued expectation $\mathbb{E}[g(\bm{X})]= [\mathbb{E}[X], \mathbb{E}[Y]]$.
\item For $g(\bm{X}) = X+Y$, we have $\mathbb{E}[g(\bm{X})]  = \mathbb{E}[X] + \mathbb{E}[Y]$.
\end{itemize}

The \textbf{covariance} of $\bm{X}$ is given by 
\[
\text{COV}[\bm{X}] = \mathbb{E}[(\bm{X} - \mathbb{E}[\bm{X}])  \otimes (\bm{X} - \mathbb{E}[\bm{X}]) ]
\]
where
\begin{align*}
\text{COV}[\bm{X}]_{11} &= \mathbb{E}[(X - \mathbb{E}[X])^2] = \text{VAR}[X]\\
\text{COV}[\bm{X}]_{22} &= \mathbb{E}[(Y - \mathbb{E}[Y])^2] = \text{VAR}[Y]\\
\text{COV}[\bm{X}]_{12} &= \text{COV}[\bm{X}]_{21} = \mathbb{E}[(X - \mathbb{E}[X])(Y - \mathbb{E}[Y])].
\end{align*}
$X$ and $Y$ are said to be \textbf{uncorrelated} if $\text{COV}[\bm{X}]_{12} = 0$. Furthermore, $\text{COV}[\bm{X}]_{12} = 0$ for independent RVs. \textbf{Caution: $\text{COV}[\bm{X}]_{12} = 0$ does  not imply the RVs are independent!}

Finally, we are interested in looking at the pdf of $Y$ when we know $X=\hat{x}$. A good guess would be $f_{XY}(\hat{x},y)$. However, this need not be a pdf as it need not integrate to unity over $y$. This leads us to the \textbf{conditional pdf} of $Y$ when we know $X=\hat{x}$,
\[
f_{Y|X}(y|\hat{x}) = \frac{f_{XY}(\hat{x},y)}{\int_{-\infty}^\infty f_{XY}(\hat{x},y) \ud y} = \frac{f_{XY}(\hat{x},y)}{f_X(\hat{x})}.
\]
Similarly, we can write the conditional pdf of $X$ given $Y=\hat{y}$ as
\[
f_{X|Y}(x|\hat{y}) = \frac{f_{XY}(x,\hat{y})}{f_Y(\hat{y})}.
\]
We note that the extension of a regression problem to probabilistic framework leads us to determining the conditional distribution of the output given an instance of the input. This will be discussed in Section \ref{sec:sup_prob}.

\section{Unsupervised probabilistic deep learning algorithms}
We begin with a vector of RVs $\X$ with $N_X$ components with a pdf given by $f_X$. 
Let's assume that we are given a dataset of samples $\{\x_i\}$ sampled from the density $f_X$, which we denote by $\x_i \sim f_X$. For instance, these samples could correspond to RGB images of cars, with a resolution of $512 \times 512$. Note that this would mean that the samples would lie in a space with dimension $N_X = 512 \times 512 \times 3$, which is quite large! We can treat each pixel of the images as a RV, taking values given by the pixel intensities (across all 3 channels). Thus, these images can be seen as samples of a $N_X$-dimensional RV with some unknown density $f_X$. Also, because of the inherent structure of the objects (i.e. the cars) in these images, the various components of the multidimensional RV can be expected to be highly correlated, leading to a non-trivial form of $f_X$. This correlation also implies that it might be possible to reduce the dimension of $\X$ from $N_X$ to a smaller number and thus make the representation simpler. 

We are interested in using the given finite set of samples $\{\x_i\}$ to learning the density $f_X$ of the data, and generating new samples from the learned distribution. Such methods are known as \textit{generative algorithms}. Although a number of generative algorithms are available, we focus on a specific type of deep learning algorithm known as Generative Adversarial Networks, or GANs for short.

\subsection{GANs}
GANs were first proposed by Goodfellow et al. \cite{goodfellow2014generative} in 2014. Since then, many variants of GANs have been proposed which differ based on the network architecture and the objective function used to train the GAN. We begin by describing the abstract problem setup followed by the architecture and training procedure of a GAN. 

Consider the dataset $\mathcal{S} = \{ \x_i \in \Omega_X \subset \Ro^{N_X}: 1 \leq i \leq N_\text{train}\}$. We assume the samples are realizations of some RV $\X$ with density $f_X$, i.e., $\x_i \sim f_X$. We want to train a GAN to learn $f_X$ and generate new samples from it.

A GAN typically comprises two sub-networks, a generator and a discriminator (or critic). The generator is a network of the form
\begin{equation}
    \g(.;\btheta_g) : \Omega_Z \rightarrow \Omega_X, \quad \g : \z \mapsto \x
\end{equation}
where $\btheta_g$ are the trainable parameters and $\z \in \Omega_Z \subset \Ro^{N_Z}$ is the realization of a RV $\Z$ following a simple distribution, such as an uncorrelated multivariate Gaussian with density
\[
f_{Z}(\z) = \frac{1}{\sqrt{(2 \pi)^2 \text{det}(\bm{\Sigma})} } \exp\left[-\frac{1}{2}(\z - \bm{\mu})^\top \bm{\Sigma}^{-1} (\z-\bm{\mu})\right] \ \ \text{with} \ \ \bm{\mu} = \bm{0}, \quad \bm{\Sigma} = \bm{I}.
\]
Typically, $N_Z \ll N_X$ with $\Z$ known as the \textit{latent variable} of the GAN. The architecture of the generator will depend on the size/shape of $\x$. If $\x$ is a vector, then $\g$ can be an MLP with input dimension $N_Z$ and output dimension $N_X$. If $\x$ is an image, say of shape $H \times W \times 3$, then the generator architecture will have a few fully connected layers, followed by a reshape into a coarse image with many channels, which is pushed through a number of transpose convolution channels that gradually increase the spatial resolution and compress the number of channels to finally scale up to the shape $H \times W \times 3$. This is also known as a \textit{decoder} architecture, similar to the upward branch of a U-Net (see Figure \ref{fig:unet}.) In either case, for a fixed $\btheta_g$, the generator $\g$ transforms the RV $\Z$ to another RV,
$\X^g = \g(\Z;\btheta_g)$
with density $f^g_X$, which corresponds to the latent density $f_Z$ pushed-forward by $\g$. We want to choose $\btheta_g$ such that $f^g_X$ is close to the unknown target distribution $f_X$.

The critic network is of the form
\begin{equation}
    d(.;\btheta_d) : \Omega_X \rightarrow \Ro
\end{equation}
with the trainable parameters $\btheta_d$. Once again, the critic architecture will depend on the shape of $\x$. If $\x$ is a vector then $d$ can be an MLP with input dimension $N_X$ and output dimension $1$. If $\x$ is an image, then the critic architecture will have a few convolution layers, followed be a flattening layer and a number of fully connected layers. This is similar to the CNN architecture shown in Figure \ref{fig:cnn} but with a scalar output and without an output function. 

\begin{figure}[htbp]
\begin{center}
\includegraphics[width=1.0\textwidth]{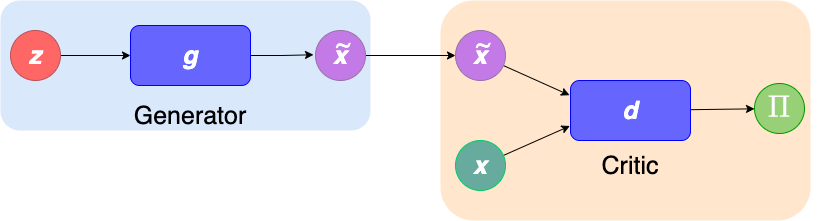}
\caption{Schematic of a GAN}
\label{fig:GAN}
\end{center}
\end{figure}

The schematic of the GAN along with the inputs and outputs of the sub-networks is shown in Figure \ref{fig:GAN}. The generator and critic play adversarial roles. The critic is trained to distinguish between true samples coming from $f_X$ and fake samples generated by $\g$ with the density $f_X^g$. The generator on the other hand is trained to fool the critic by trying to generate realistic samples, i.e., samples similar to those sampled from $f_X$. 

We define the objective function describing a \textit{Wasserstein GAN (WGAN)} \cite{arjovsky2017wasserstein_proc}, which has better robustness and convergence properties compared to the original GAN. The objective function is given by
\begin{equation}
    \Pi(\btheta_g,\btheta_d) = \underbrace{\frac{1}{N_\text{train}} \sum_{i=1}^{N_\text{train}} d(\x_i;\btheta_d)}_{\text{critic value on real samples}} - \underbrace{\frac{1}{N_\text{train}} \sum_{i=1}^{N_\text{train}} d(\g(\z_i;\btheta_g);\btheta_d)}_{\text{critic value on fake samples}}
\end{equation}
where $\x_i \in \mathcal{S}$ are samples from the true target distribution $f_X$, while $\z_i \sim f_Z$ are passed through $\g$ to generate the fake samples. To distinguish between true and fake samples, the critic attains large positive values when evaluated on real samples and large negative values on fake generated samples. Thus, critic is trained to maximize objective function. In other words, we want to solve the problem
\begin{equation}\label{eqn:max}
    \btheta_d^*(\btheta_g) = \argmax{\btheta_d}\Pi(\btheta_g,\btheta_d) \ \ \text{for any } \btheta_g.
\end{equation}
Note that the optimal parameters of the critic will depend on $\btheta_g$. Now to fool the critic, the generator $\g$ tries to minimize the objective function,
\begin{equation}\label{eqn:min}
    \btheta_g^* = \argmin{\btheta_g}\Pi(\btheta_g,\btheta^*_d).
\end{equation}
Thus, training the WGAN corresponds to solving a minmax optimization problem. We note that the critic and the generator are working in an adversarial manner. That is, while the former is trying to maximize the objective function, the latter is trying to minimize it. Hence the name generative \textit{adversarial} network. 

In practice, we need to add a stabilizing term to the critic loss. So the critic is trained to maximize 
\begin{equation}\label{eqn:critic_loss}
    \Pi_c(\btheta_g,\btheta_d) = \Pi(\btheta_g,\btheta_d) - \frac{\lambda}{\bar{N}} \sum_{i=1}^{\bar{N}} \left( \left\| \df{d}{\hat{\x}}(\hat{\x}_i;\btheta_d)\right\| - 1\right)^2
\end{equation}
where $\hat{\x}_i = \alpha \x_i + (1-\alpha) \g(\z_i;\btheta_g)$ and $\alpha$ is sampled from a uniform RV in $(0,1)$. The additional term in \eqref{eqn:critic_loss} is known as a \textit{gradient penalty} term and is used to constraining the (norm of) gradient of the critic $d$ with respect to its input to be close to 1, and thus be 1-Lipschitz function. For further details on this term, we direct the interested readers to \cite{gulrajani2017improved}.

The iterative Algorithm \ref{alg:gan} is used train $\g$ and $d$ simultaneously, which is also called \textit{alternating steepest descent}, where $\eta_d$ and $\eta_g$ are the learning rates for the critic and the generator, respectively. Note that we take $K > 1$ optimization steps for the critic followed by a single optimization step for the generator. This is because we want to solve the inner maximization problem first so that the critic is able to distinguish between real and fake samples. Although taking a very large $K$ would lead to a more accurate solve of the minmax problem, it would also make the training algorithm computationally intractable for moderately sized networks. Thus, $K$ is typically chosen between 4 to 6 in practice. 

\RestyleAlgo{ruled}
\begin{algorithm}
\caption{Algorithm to train a GAN}\label{alg:gan}
\KwIn{$\btheta^0_d, \btheta^0_g, K, N\_\text{epochs}, \eta_d,\eta_g$}
\For{$n=1,...,N\_\text{epochs}$}{
    $\hat{\btheta}_d \leftarrow \btheta^{(n-1)}_d$\\
    \For{$k=1,...,K$}{
         Maximization update: 
         \[\hat{\btheta}_d \leftarrow \hat{\btheta}_d + \eta_d \df{\Pi_c}{\btheta_d}(\btheta_g^{(n-1)},\hat{\btheta}_d)\]
    }
    $\btheta^{(n)}_d \leftarrow \hat{\btheta}_d$\\
    Minimization update: 
    \[\btheta^{n}_g \leftarrow \btheta^{(n-1)}_g - \eta_g \df{\Pi}{\btheta_g}(\btheta_g^{(n-1)},\btheta^{(n)}_d)\]
    }
\end{algorithm}

The minmax problem is a hard optimization problem to solve, and  convergence is usually reached after training for many epochs. Alternatively, the critic optimization steps can be done over mini-batches of the training data, with many mini-batches taken per epoch, leading to a similar number of optimization steps for a relatively small number of epochs. As the iterations go on, $d$ becomes better at detecting fake samples and $\g$ becomes better at creating samples that can fool the critic.

Under the assumption of infinite capacity ($N_{\btheta_g},N_{\btheta_d} \rightarrow \infty$), infinite data ($N_\text{train}\rightarrow \infty$) and a perfect optimizer, we can prove that the generated distribution $f_X^g$ \textbf{converges weakly} to the target distribution $f_X$ \cite{arjovsky2017wasserstein_proc}. This is equivalent to saying
\begin{equation}
\mathbb{E}_Z[\ell(\gen(\Z;\btheta_g^*))] \longrightarrow \mathbb{E}_X[\ell(\X)], 
\end{equation}
for every continuous, bounded function $\ell$ on $\Omega_X$, i.e., $\ell \in C_b(\Omega_X)$. Once the GAN is trained, we can use the optimized $\g$ to generate new samples from $f_X^g \approx f_X$ by first sampling $\z \sim f_Z$, and then passing it through the generator to get the sample $\x = \g(\z;\btheta^*_g)$. Furthermore, due to the weak convergence described above, the statistics (mean, variance, etc) of the generated samples will convergence to the true statistics associated with $f_X$.

\begin{remark}
We make a few important remarks here:
\begin{enumerate}
    \item Once the GAN is trained, we typically only retain the generator and don't need the critic. The primary role of  training the critic is to obtain a suitable $\g$ that can generate realistic samples.
    \item The reason the term "Wasserstein" appears in the name WGAN is because one can show that solving the minmax problem is equivalent to minimizing the Wasserstein-1 distance between $f_X^g$ and $f_X$ \cite{arjovsky2017wasserstein_proc,villani2008optimal}. The Wasserstein-1 distance is a popular metric used to measure discrepancies between two probability measures. 
    \item Since the dimension $N_Z$ of the latent variable is typically much smaller than the dimension $N_X$ of samples in $\Omega_X$, the trained generator also provides a low dimensional representation of high-dimensional data, which can be very useful in several downstream tasks \cite{patel2021,patel2022}.  
\end{enumerate}
\end{remark}

\section{Supervised probabilistic deep learning algorithms}
\label{sec:sup_prob}

Recall the deterministic problem where given the labelled/pairwise dataset 
\begin{equation}\label{eqn:dataset_prob}
    \mathcal{S} = \{(\x_i,\y_i): \x_i \in \Omega_X \subset \Ro^{N_X}, \ \y \in \Omega_Y \subset \Ro^{N_Y}\}_{i=1}^{N_\text{train}}
\end{equation}
we want to find $\y$ for a new $\x$ not appearing in $\mathcal{S}$. We have seen in the previous chapters how neural networks can be used to solve such a regression (or classification) problem.

Now let us consider the probabilistic version of this problem. We assume that $\x$ and $\y$ are modelled using RVs $\X$ and $\Y$, respectively. Further, let the paired samples in \eqref{eqn:dataset_prob} be drawn from the unknown joint distribution $f_{XY}$. Then given a realization $\X = \hat{\x}$, we wish to use $\mathcal{S}$ to determine the conditional distribution $f_{Y|X}(\y|\hat{\x})$ and generate samples from it.

There are several popular approaches to solve this probabilistic problem, such as Bayesian neural networks, variational inference, dropouts or deep Boltzman machines. But we will focus on an extension of GANs which also addresses these type of problems. 

\subsection{Conditional GANs}
Conditional GANs were first  proposed in \cite{mirza2014} to learn conditional distributions. We will discuss a special variant of these models known as conditional Wasserstein GANS (cWGANs) which were developed in \cite{adler2018deep}, and used to solve a number of physics-based (inverse) problems in \cite{ray2022}.  

\begin{figure}[htbp]
\begin{center}
\includegraphics[width=1.0\textwidth]{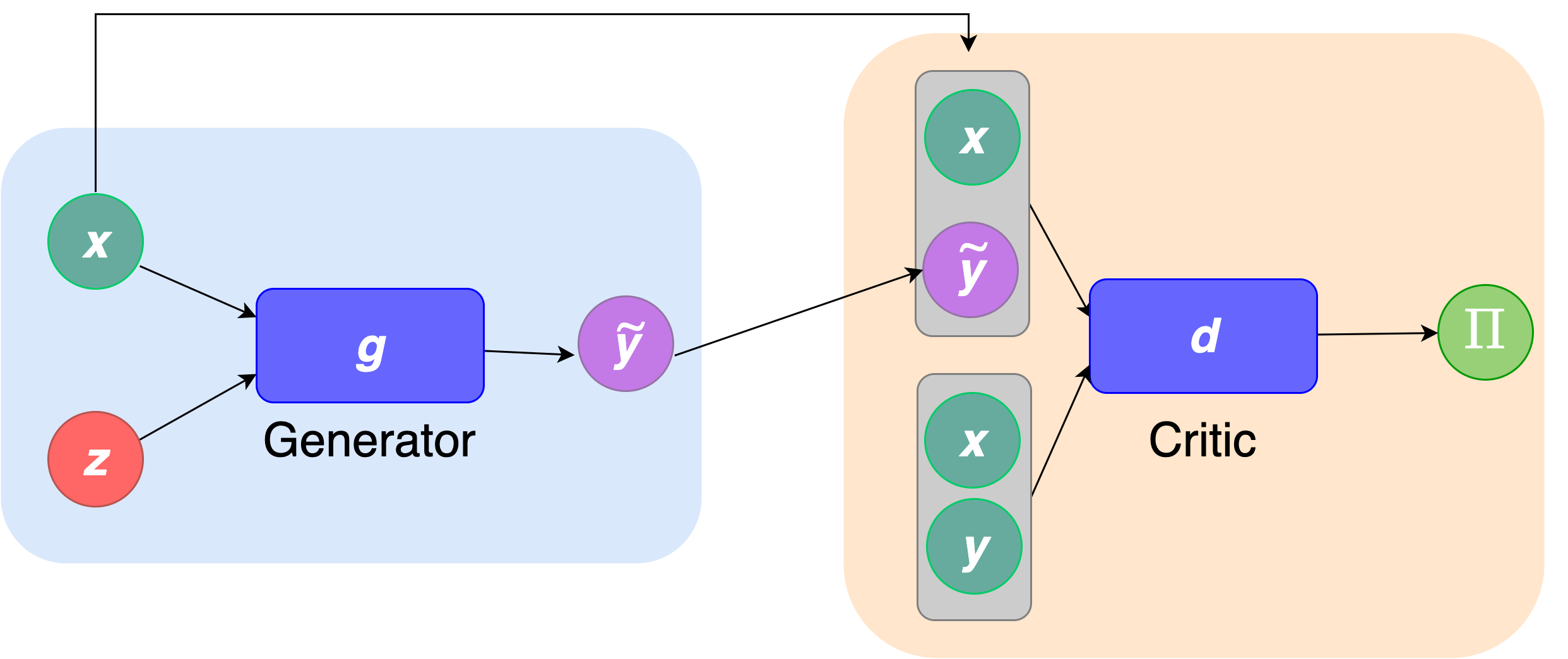}
\caption{Schematic of a conditional GAN}
\label{fig:cgan}
\end{center}
\end{figure}

The schematic of a conditional GAN is depicted in Figure \ref{fig:cgan}. The generator is a network of the form
\begin{equation}
    \g(.;\btheta_g) : \Omega_Z \times \Omega_X \rightarrow \Omega_Y, \quad \g : (\z,\x) \mapsto \y
\end{equation}
where $\z \sim f_Z$ is the latent variable. Note that unlike a GAN, the generator in a conditional GAN also takes as input $\x$. For a given value of $\X=\hat{\x}$,  sampling $\z \sim f_Z$ will generate many samples of $\y$ from some induced conditional distribution $f^g_{Y|X}(\y|\hat{\x})$. The goal is to prescribe the parameters $\btheta_g$ such that $f^g_{Y|X}(\y|\hat{\x})$ approximates the true conditional $f_{Y|X}(\y|\hat{\x})$ for (almost) every value of $\hat{\x}$.

The critic is a network of the form
\begin{equation}
    d(.;\btheta_d) : \Omega_X \times \Omega_Y \rightarrow \Ro
\end{equation}
which is trained to distinguish between paired samples $(\x,\y)$ generated from the true joint distribution $f_{XY}$ and the fake pairs $(\x,\hat{\y})$ where ${\hat{\y}}$ is generated by $\g$ given (real) $\x$.

The objective function for a cWGAN is given by
\begin{equation}
    \Pi(\btheta_g,\btheta_d) = \underbrace{\frac{1}{N_\text{train}} \sum_{i=1}^{N_\text{train}} d(\x_i,\y_i;\btheta_d)}_{\text{critic value on real pairs}} - \underbrace{\frac{1}{N_\text{train}} \sum_{i=1}^{N_\text{train}} d(\x_i,\g(\z_i,\x_i;\btheta_g);\btheta_d)}_{\text{critic value on fake pairs}}.
\end{equation}
As earlier, the critic is trained to maximize the objective function (given by \eqref{eqn:max}) while the generator is trained to minimize it (given by \eqref{eqn:min}). Further, a stabilizating gradient penalty term needs to be included when optimizing the critic (see \cite{ray2022}). The generator and critic are trained using the alternating steepest descent algorithm described for GANs.

Under the assumption of infinite capacity ($N_{\btheta_g},N_{\btheta_d} \rightarrow \infty$), infinite data ($N_\text{train}\rightarrow \infty$) and a perfect optimizer, we can prove \cite{adler2018deep} that the generated conditional distribution $f_{Y|X}^g(\y|\hat{\x})$ converges in a weak sense to the target condition distribution $f_{Y|X}(\y|\hat{\x})$ (on average) for a given $\X=\hat{\x}$.